\documentclass[journal]{IEEEtran}

\usepackage[english]{babel}

\usepackage{ifpdf}

\usepackage{cite} 
\usepackage{url}
\usepackage{hyperref}

\usepackage{./latex_resources/tikz-network}

\usepackage{pgf, tikz, pgfplots}
\usetikzlibrary{shapes, arrows, automata, plotmarks}
\usetikzlibrary{calc, hobby, decorations}

\usepackage{color}
\usepackage{verbatim}
\usepackage{xcolor}
\usetikzlibrary{positioning,decorations.pathreplacing,shapes}
\usepackage{microtype}
\usetikzlibrary{math}

\usepackage{amsmath}
\usepackage{amsfonts, amssymb, amsthm}
\usepackage{mathrsfs}

\usepackage{algorithm,algpseudocode}
	\algnewcommand{\LeftComment}[1]{\Statex \(\triangleright\) #1}

\newcommand{\makeboxlabel}[1]{#1\hfill}

\usepackage{enumerate}
\usepackage{multirow}
\usepackage{rotating}
\usepackage{subcaption}
\usepackage{needspace}

\input{./latex_resources/mySymbol.sty}
\input{./latex_resources/pennColors.sty}

\newtheorem{theorem}{\hspace{0pt}\bf Theorem}

\newtheorem{remark}{\hspace{0pt}\bf Remark}

\newtheorem{definition}{\hspace{0pt}\bf Definition}

\begin{document}

\title{Sampling and Uniqueness Sets in Graphon Signal Processing}

\author{$^\clubsuit$Alejandro Parada-Mayorga and $^\diamondsuit$Alejandro Ribeiro
\thanks{$^\clubsuit$ Dept. of Electrical Engineering, University of Colorado at Denver, USA e-mail: alejandro.paradamayorga@ucdenver.edu.  $^\diamondsuit$ Dept. of Electrical and Systems Eng., Univ.of Pennsylvania.}}

\markboth{Signal Processing}
{Shell \MakeLowercase{\textit{et. al.}}: Bare Demo of IEEEtran.cls for Journals}

\maketitle



\begin{abstract}
In this work, we study the properties of sampling sets on families of large graphs by leveraging the theory of graphons and graph limits. We extend to graphon signals the notion of removable and uniqueness sets, which was developed originally for the analysis of signals on graphs. We state the formal definition of a $\Lambda-$removable set and conditions under which a bandlimited graphon signal can be represented uniquely when its samples are obtained from the complement of a $\Lambda-$removable set in the graphon. By leveraging such results we show that graphon representations of graph signals can be used as a common framework to compare sampling sets between graphs with different numbers of nodes and node labelings. Additionally, given a sequence of graphs that converges to a graphon, we show that the sequences of sampling sets whose graphon representation is identical in $[0,1]$ are convergent as well. We exploit the convergence results to provide an algorithm that obtains approximately close to optimal sampling sets in large graphs where traditional methods are intractable. Performing a set of numerical experiments, we evaluate the quality of these sampling sets. Our results open the door for the efficient computation of optimal sampling sets in large graphs relying on existing methods that can be applied in small graphs. 
\end{abstract}


\IEEEpeerreviewmaketitle




\section{Introduction}


The problem of sampling signals on graphs is extensively studied in graph signal processing (GSP)~\cite{puysampling,anis_conf,segarra_samlocagreg,chensamplingongraphs,tremblayAB17,pesensonams1,pesenson2009,pesenson2019,fuhrpesenson,pesensonschrodinger,pesenson2010,ortega_proxies,ortega2022introduction,tsitsverobarbarossa,6638704,bookgspchapsampling,7055883,8047995,7581102,JAYAWANT2022108436,alejopm_phdthesis,alejopm_cographs_c,alejopm_BN_j}. In this problem, multiple methods and heuristics aim to find, in an efficient way, a subset of nodes that completely determines a family of bandlimited signals and the subset of nodes that reduce the reconstruction error when the signal is polluted with noise~\cite{bookgspchapsampling,ortega2022introduction}. At the center of this problem are the concepts of \textit{uniqueness set} and \textit{$\Lambda$-removable set} introduced by Pensenson~\cite{pesensonams1,pesenson2009,pesenson2019,fuhrpesenson,pesensonschrodinger,pesenson2010,ortega_proxies}. These concepts allow the characterization of the goodness of a given sampling set. In particular, if the sampling problem under consideration is noise-free, two bandlimited signals that coincide on a uniqueness set also coincide on all the nodes of the graph, with a bandwidth that is reciprocal to a constant that characterizes the sampling set as a removable set~\cite{pesensonams1,pesenson2009,pesenson2019,fuhrpesenson,pesensonschrodinger,pesenson2010,ortega_proxies}. Highly efficient methods are readily available for the sampling of graphs of different sizes~\cite{puysampling,anis_conf,chensamplingongraphs,tremblayAB17,ortega_proxies,ortega2022introduction,tsitsverobarbarossa,6638704,bookgspchapsampling,7055883,8047995,7581102,alejopm_phdthesis,JAYAWANT2022108436}, however, once the number of nodes goes above several thousand, such methods become inapplicable even when the large graphs share common structural properties with small graphs, where optimal sampling methods can be applied.


In this paper, we study the properties of sampling sets on collections of graphs with common structural properties. To this end we rely on the theory of graphons and graph limits developed in~\cite{lovasz2012large}. Using graphon representations of graphs, we show that the uniqueness sets of bandlimited signals can be interrelated, and the closeness of such relationship is determined by the distance in the $L_{2}$ norm between the graphon shift operators associated with the graphs. We exploit the concepts of $\Lambda$-removable sets and uniqueness sets in graphs -- originally proposed by Pensenson in~\cite{pesensonams1,pesenson2009,pesenson2019,fuhrpesenson,pesensonschrodinger,pesenson2010,ortega_proxies} -- to provide the following contribution:


\smallskip
\begin{list}
      {}
      {\setlength{\labelwidth}{22pt}
       \setlength{\labelsep}{0pt}
       \setlength{\itemsep}{0pt}
       \setlength{\leftmargin}{22pt}
       \setlength{\rightmargin}{0pt}
       \setlength{\itemindent}{0pt} 
       \let\makelabel=\makeboxlabel
       }

\item[{\bf (C1)}] We extend the notion of removable sets and uniqueness sets to bandlimited graphon signals.
\end{list}
\smallskip


\noindent The implication of~\textbf{(C1)} is the fundamental cornerstone tool used to quantitively characterize the properties of subsets in $[0,1]$ for the description and representation of bandlimited graphon signals. Then, by building on~\textbf{(C1)} and taking advantage of the graphon space as a general framework for graphs and graph signal representation, we provide the following contribution:


\smallskip
\begin{list}
      {}
      {\setlength{\labelwidth}{22pt}
       \setlength{\labelsep}{0pt}
       \setlength{\itemsep}{0pt}
       \setlength{\leftmargin}{22pt}
       \setlength{\rightmargin}{0pt}
       \setlength{\itemindent}{0pt} 
       \let\makelabel=\makeboxlabel
       }
\item[{\bf (C2)}] We prove that uniqueness sets from arbitrary graphs can be compared on the graphon space and such comparison is measured by the differences between the graphon shift operators defined on the induced graphon representations.
\end{list}
\smallskip


\noindent More specifically we derive inequalities that relate the $\Lambda$-removable constants of the complement of the uniqueness sets of the two graphs. Such inequalities take place when the subsets of nodes in the graphs are identical when represented on the graphon space. By exploiting the quantitative relationships derived from \textbf{(C2)} we lay out the way for the following contribution:


\smallskip
\begin{list}
      {}
      {\setlength{\labelwidth}{22pt}
       \setlength{\labelsep}{0pt}
       \setlength{\itemsep}{0pt}
       \setlength{\leftmargin}{22pt}
       \setlength{\rightmargin}{0pt}
       \setlength{\itemindent}{0pt} 
       \let\makelabel=\makeboxlabel
       }
\item[{\bf (C3)}]  We show that the uniqueness sets are inherited in collections of graphs whose structural properties are similar and whose similarity is measured by the difference between graphon shift operators.
\end{list}
\smallskip


\noindent For {\bf (C3)} we establish a concrete inequality that interrelates the uniqueness sets taking into account the effects attributed to the number of nodes of each graph. As a consequence of the result stated in {\bf (C3)}, we observe a dynamic behavior for the admissibility region of the uniqueness sets in terms of the number of nodes in each graph. This is, a change in the number of nodes in one graph deforms the region of possible sampling sets in the other graph, assuming such sets are identical when represented in the graphon space. This particular implication is important as it also establishes concrete degrees of freedom on how a sampling set can change when a graph is affected by edge droppings. In this latter case, the characteristics of the region of possible sampling sets in the new graph -- with modified edges -- are a consequence of the differences in the graphon shift operators. Considering the inequalities derived in {\bf (C3)} on sequences of convergent graphs we provide the following contribution:


\smallskip
\begin{list}
      {}
      {\setlength{\labelwidth}{22pt}
       \setlength{\labelsep}{0pt}
       \setlength{\itemsep}{0pt}
       \setlength{\leftmargin}{22pt}
       \setlength{\rightmargin}{0pt}
       \setlength{\itemindent}{0pt} 
       \let\makelabel=\makeboxlabel
       }
\item[{\bf (C4)}] We prove that sequences of uniqueness sets on a convergent sequence of graphs to a graphon -- in the $L_{2}$ norm -- converge to a limit uniqueness set.
\end{list}
\smallskip


\noindent Contribution \textbf{(C4)} has important implications because it highlights the fact that uniqueness sets are structurally preserved on a convergent sequence of graphs. This last point opened the door for the following contribution:


\smallskip
\begin{list}
      {}
      {\setlength{\labelwidth}{22pt}
       \setlength{\labelsep}{0pt}
       \setlength{\itemsep}{0pt}
       \setlength{\leftmargin}{22pt}
       \setlength{\rightmargin}{0pt}
       \setlength{\itemindent}{0pt} 
       \let\makelabel=\makeboxlabel
       }
\item[{\bf (C5)}] We introduce a generic algorithm that derives \textit{approximately optimal} sampling sets in large graphs that belong to a sequence of graphs that converges to a graphon. We achieve this by exploiting the optimal sampling sets in the small graphs of the sequence and reusing them in larger graphs.
\end{list}
\smallskip


\noindent Contribution {\bf (C5)} has direct implications in practical scenarios where large graphs are involved. In particular, if the graph is prohibitively large no existing sampling technique can be used. However, in the light of {\bf (C5)}, if the structural properties of the graph are shared with another small graph, one can use the existing sampling techniques to find the optimal sampling set in the small graph and then reuse it to approximate the optimal sampling set in the large graph. Notice that such scenarios appear in common applications where networks increase in size in somewhat structured ways, such as power electricity networks and sensor networks~\cite{owerko1,owerko2,owerko3}. Contribution \textbf{(C5)} also has a fundamental connection with the results derived in~\cite{graphon_pooling_j} for the operation of pooling in large graphs, where the equipartitions in $[0,1]$ can be used to reduce the size of a graph while preserving structural properties of the original graphs.

We test the Algorithm's performance in~\textbf{(C5)} considering multiple scenarios with several types of graphs, graphons, and with several instances for the number of nodes. Our experiments allow us to compare the performance of the approximately optimal sampling sets against the optimal ones and random patterns. We show that the approximately optimal sampling sets offer significantly better performance than the one obtained using random sampling. The comparison against random sampling follows the fact that random sampling is the only alternative available for graphs of extremely large sizes~\cite{JAYAWANT2022108436}.


This paper is organized as follows. In Section~\ref{sec_gsp} we discuss the basics of graph signal processing, describing in detail terminology and the sampling problem of bandlimited signals on graphs. We also introduce the notion of uniqueness sets and removable sets on graphs. In Section~\ref{sec_gphon_sp} we discuss graphon signal processing and we introduce some definitions necessary for the following sections. In Section~\ref{sec_samp_on_graphons} we extend the notions of uniqueness sets and $\Lambda$-removable sets for bandlimited graphon signals. We establish a quantitative relationship between uniqueness sets of graphs that are interrelated by a graphon representation. We prove that in a convergent sequence of graphs, sequences of uniqueness sets converge to a universal uniqueness set defined on the graphon limit. Additionally, we present an algorithm for the computation of approximately optimal sampling sets taking into account the results derived on the convergence of uniqueness sets. In Section~\ref{sec_num_sim} we present a set of numerical experiments to evaluate the performance of the algorithm introduced in previous sections and to validate our results and insights. Finally, in Section~\ref{sec_conclusions} we present a discussion and some conclusions.




\section{Graph Signal Processing}
\label{sec_gsp}

Let $G=(V(G),E(G),w_{G})$ be an undirected graph with a set of vertices $V(G)$, a set of edges $E(G)$, and a weight function $w_{G}: E(G)\rightarrow\mathbb{R}^{+}$. A signal on $G$ is defined as the map $\bbx: V(G)\rightarrow\mathbb{R}$. Then, for each node $i\in V(G)$ there is a scalar, $\mathbf{x}(i)\in\mbR$, associated with it. If $\vert V(G)\vert$ indicates the number of vertices in $V(G)$, the indexed collection of the values $\{ \mathbf{x}(i) \}_{i=1}^{\vert V(G)\vert}$ is identified with the vector $\mathbf{x}\in\mbR^{\vert V(G)\vert}$~\cite{ortega2022introduction,gdsp_moura}. We will use the symbol $(G,\bbx)$ to denote a graph signal on $G$. This emphasizes that each component of $\bbx$ is associated with a node in $V(G)$.

Convolutional processing of signals on graphs is carried out by using polynomials whose independent variable is a matrix representation of the graph. Such a matrix is known as the \textit{shift operator} and it is represented by $\bbS_{G}$. Typical choices of this operator include the adjacency matrix, the Laplacian, and the normalized Laplacian among others~\cite{ortega2022introduction,alejopm_phdthesis,gdsp_moura}. For our discussion, the shift operator is the adjacency matrix, $\bbA$, whose entries are given by $\mathbf{A}(i,j) = w_{G}(\{i,j\})$ with $\{i,j\}\in E(G)$. Then, with $\bbS_{G} = \bbA$, we can write convolutional filters as
\begin{equation}
\bbH (\bbS_G)
    =
      \sum_{k=0}^{K}h_{k}\bbS^{k}_G
      ,
\end{equation}
where the action of $\bbH (\bbS_G)$ on the graph signal $(G, \bbx)$ produces the signal $(G,\bby)$ with
\begin{equation}
\bby =
\bbH (\bbS_G)\bbx
    =
      \sum_{k=0}^{K}h_{k}\bbS^{k}_G
      \bbx
      .
\end{equation}

If $\bbS_G$ is diagonalizable, we can write $\mathbf{S}_G =\mathbf{U}\mathbf{\Lambda}\mathbf{U}^{\mathsf{T}}$, where $\bbU$ is an orthogonal matrix and $\mathbf{\Lambda}$ is a diagonal matrix containing the eigenvalues of $\bbS_G$, $\lambda_{i}(\bbS_{G})$. By leveraging this decomposition we can calculate the \textit{graph Fourier transform (GFT)} of the graph signal $(G, \bbx)$ as $(\widehat{G},\widehat{\bbx})$ where
\begin{equation}
\widehat{\bbx}
=
\bbU^{\mathsf{T}}\bbx
,
\end{equation}
and where the symbol $\widehat{G}$ indicates that the domain of $\widehat{\bbx}$ is given by the eigenvalues $\lambda_{i}(\bbS_{G})$. The $i$th component of the vector $\widehat{\bbx}\in\mbR^{N}$ is associated with the eigenvalue $\lambda_{i}(\bbS_G)$. In this paper, the ordering of the $\lambda_{i}( \bbS_G )$'s is given according to $\vert\lambda_{1}(\bbS_G )\vert\geq\vert\lambda_{2}(\bbS_G)\vert \geq\cdots\geq\vert\lambda_{\vert V(G)\vert}(\bbS_G)\vert$. This is, while the labeling/ordering of the nodes in $V(G)$ can be arbitrary, the ordering of the frequency components is dictated by the absolute value of the $\lambda_{i}(\bbS_G)$. We adopt this convention in order to facilitate the connection between spectral representations of graphs and their induced graphons which will be discussed in Section~\ref{sec_gphon_sp}.

A graph signal $(G,\bbx)$ is said to be bandlimited with bandwidth $\omega$ if $\widehat{\bbx}(i) = 0$ for all $\vert \lambda_i (\bbS_G)\vert<\omega$. The set of bandlimited signals with bandwidth $\omega$ is denoted by $\mathcal{PW}_{\omega}(G)$. This is, $\mathcal{PW}_{\omega}(G)$ is spanned by those columns in $\bbU$ associated with the eigenvalues that satisfy $\vert \lambda_i (\bbS_G) \vert \geq \omega$.


\subsection{Sampling  signals on Graphs}

Given a space of signals $\mathcal{M}$ on a graph $G=(V(G),E(G),w_{G})$ the problem of sampling can be defined in multiple scenarios~\cite{ortega2022introduction,alejopm_phdthesis,alejopm_BN_j}, among which we highlight the following:
\smallskip
\begin{list}
      {}
      {\setlength{\labelwidth}{26pt}
       \setlength{\labelsep}{-3pt}
       \setlength{\itemsep}{10pt}
       \setlength{\leftmargin}{26pt}
       \setlength{\rightmargin}{0pt}
       \setlength{\itemindent}{0pt} 
       \let\makelabel=\makeboxlabel
       }
\item [(SC1)]~Find the subset of nodes $\ccalS\subset V(G)$ of minimum size such that any signal $\bbx\in\mathcal{M}$ is uniquely determined from its values in $\ccalS$.
\item[(SC2)]~Find the subset of nodes $\ccalS\subset V(G)$ of minimum size such that any signal $\bbx\in\mathcal{M}$ can be reconstructed from their values in $\ccalS$ with the lowest error, assuming the sampled values are polluted by noise. In this scenario, when the noise goes to zero the subsets obtained are identical to those in (SC1).
\item[(SC3)]~Given a fixed number $m<\vert V(G)\vert$, find the subset of nodes $\ccalS\subset V(G)$ with $\vert \ccalS\vert = m$ such that any signal $\bbx\in\mathcal{M}$ can be reconstructed with the lowest reconstruction error assuming the samples are polluted by noise.  
\end{list}
\medskip
In scenario (SC1) the notion of the quality of a possible reconstruction of the signal is not specified since the samples of the signal are assumed to be noiseless. In this case, the reconstruction is expected to be the same for any sampling that satisfies the uniqueness guarantees of the reconstruction~\cite{ortega2022introduction,alejopm_phdthesis,alejopm_BN_j,alejopm_BN_c1,alejopm_BN_c2,alejopm_cographs_c}. The scenario (SC2) provides a new set of challenges since with the addition of noise there will be a difference when reconstructing the signals among the sets where the signal have a unique representation, i.e. there will be a \textit{quality} property associated to each sampling set. Finally, scenario (SC3) is connected with scenario (SC2) but with a restriction on the number of samples that can be captured.

As pointed out in~\cite{alejopm_phdthesis, ortega_proxies} the three scenarios are interrelated. The solution sets in (SC2) are a subset of the possibly multiple solution sets in (SC1). Additionally, the solution set in (SC3) can be obtained as a modified set from the solution sets in (SC1) and (SC2).

The space of signals $\mathcal{M}$ considered for the sampling problem is traditionally defined by a subset of frequencies given by the graph Fourier transform. The typical choice is $\ccalM = \mathcal{PW}_{\omega}(G)$ for some $\omega\in\mbR$, which will be the same considered in this paper.

The properties of a given sampling set concerning any sampling scheme can be characterized with the notions of \textit{$\Lambda$-removable set} introduced in~\cite{pesensonams1,pesenson2009,pesenson2019,fuhrpesenson,pesensonschrodinger,pesenson2010}. Due to its importance, we state this notion formally in the following definition.


\begin{definition}\label{def_removable_set_G}
A subset of nodes $\ccalS\subset V(G)$ on the graph $G$ is said to be a removable if there exists $\Lambda\in\mathbb{R}_{+}$ such that
\begin{equation}\label{eq:lambdasetsgraphs}
\left\Vert 
\mathbf{S}_G \bbx 
\right\Vert_2
                          <
                             \Lambda\Vert\bbx\Vert_2 ~\forall~\bbx\in L_{2}(\ccalS)
                             ,
\end{equation}
where $L_{2}(\ccalS)$ is the set of finite energy signals whose support is contained in $\ccalS$. The infimum among these constants is denoted by $\Lambda_\ccalS$.

\end{definition}


Definition~\ref{def_removable_set_G} although simple, provides a powerful characterization of subsets of $V(G)$. As shown in~\cite{pesensonams1,pesenson2009,pesenson2019,fuhrpesenson,pesensonschrodinger,pesenson2010}, using the notion of removable sets it is possible to determine when band-limited signals can be uniquely represented from their samples on a given set. In particular, if $\ccalS\subset V(G)$, any signal in $\mathcal{PW}_{\omega}(G)$ can be uniquely represented by its samples in $\ccalS$ as long as $\omega >\Lambda_{\ccalS^c}$~\cite{pesensonams1,pesenson2009,pesenson2019,fuhrpesenson,pesensonschrodinger,pesenson2010}. This can be rephrased as follows. If the graph signals $(G,\bbx)$ and $(G,\bby)$ coincide in $\ccalS$ -- i.e. $\bbx = \bby$ in $\ccalS$ --, $(G, \bbx), (G, \bby) \in\mathcal{PW}_{\omega}(W)$, and $\omega>\Lambda_{\ccalS^c}$, then $\bbx =\bby$ on $V(G)$. We will refer to those sets $\ccalS$ with $\omega>\Lambda_{\ccalS^c}$ as the \textit{uniqueness set} for signals in $\mathcal{PW}_{\omega}(G)$. Note that the characterization provided by removable sets does not require or depend on the reconstruction methods of the signals from their samples.

Anis et al. in~\cite{ortega_proxies} exploited the concept of $\Lambda$-removable sets to build one of the most efficient algorithms up to date, to find good uniqueness sets on small and medium size graphs, starting with the premise that noise is added to the samples. It is important to point out that despite the efficacy and efficiency of this algorithm, it is not suitable for increasingly large graphs.


\begin{remark}\normalfont
We emphasize that the definitions of removable and uniqueness sets are tied to the ordering of the eigenvalues of the shift operator, $\bbS_{G}$. The original definitions were stated considering an ascending ordering of eigenvalues. Such ordering alters the direction of the inequalities used, i.e. the direction of the inequality in~\eqref{eq:lambdasetsgraphs}, and whether one must consider $\omega >\Lambda_{\ccalS^c}$ or $\omega <\Lambda_{\ccalS^c}$ for the characterization of the uniqueness sets. Additionally, notice that the notion of removable set has also been named as a Poincare type inequality in Pensenson's work on sampling on manifolds and Banach spaces~\cite{pesenson_maniflPW,pesenson_sampvechilbert,feich_pesenson_samplhyperbolic,pesenriemman2017}. We choose the decreasing ordering to facilitate the connection between graph shift operators and graphon shift operators -- which will be discussed later. 
\end{remark}





\section{Graphons and Signal Processing on Graphons}
\label{sec_gphon_sp}


\begin{figure}
\centering
	\centering\includegraphics[angle=0,scale=0.5]{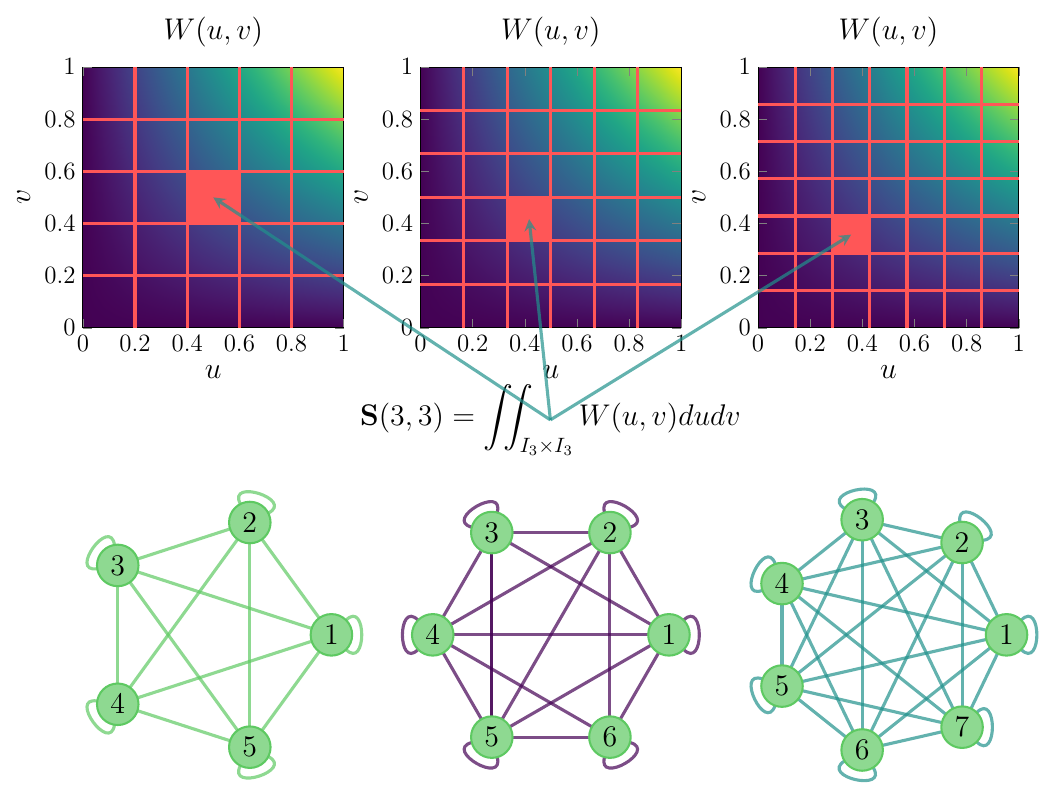}
\caption{
Three graphs are generated from a graphon, $W(u,v)$ utilizing the discretization method (GD1). A regular grid is built in $[0,1]^2$ defining an equipartition. The integral of  $W(u,v)$ over each subdomain of the partition determines the entries of the adjacency matrix of a graph, $\bbS$. For instance, the entry $\bbS(3,3)$ is obtained by integrating $W(u,v)$ over $I_{3}\times I_{3}$ -- the subdomain determined by the third interval in the equipartition. The finer the partition the more elements $\bbS$ has.
}
\label{fig_graphon_to_graph}
\end{figure}


Graphons are symmetric bounded measurable functions on the unit square, $W:[0,1]^{2}\rightarrow [0,1]$~\cite{lovasz2012large}. Although originally conceived for the study of large graphs, they quickly became valuable tools to analyze data that is defined \textit{on} the graphs and \textit{on} convergent sequences of graphs~\cite{lovasz2012large,graphon_geert_j,graphon_pooling_c,graphon_pooling_j}.

The relationship between graphons and graphs can be established in two directions. On one hand, we can use graphons as generative models of graphs. In this context, discretization methods are applied to the graphon to derive the adjacency matrix of a graph. On the other hand, we can build a graphon representation from a graph~\cite{lovasz2012large,graphon_pooling_c,graphon_pooling_j}. Regarding the first scenario, we highlight the following discretization method:
\smallskip
\begin{list}
      {}
      {\setlength{\labelwidth}{26pt}
       \setlength{\labelsep}{-3pt}
       \setlength{\itemsep}{10pt}
       \setlength{\leftmargin}{26pt}
       \setlength{\rightmargin}{0pt}
       \setlength{\itemindent}{0pt} 
       \let\makelabel=\makeboxlabel
       }
\item[(GD1)]~~Building a \textbf{regular} grid on $[0,1]^{2}$ and associating the area of the graphon $W(u,v)$ in each subdomain of the grid to a given entry in an adjacency matrix $\mathbf{A}$ (see Fig.~\ref{fig_graphon_to_graph}).
\end{list}
\medskip
Although other discretizations such as irregular grids and irregular sampling can be used to generate graphs from graphons, it was formally proved in~\cite{lovasz2012large} and~\cite{graphon_pooling_j} that regular grids have unique attributes in terms of convergence and label invariance/equivariance properties. Additionally, as we will show later, regular partitions provide a natural way to build graphon representations of graphs -- it is indeed the most commonly used and studied method to represent graphs as graphons~\cite{lovasz2012large}.

Given a graph $G=(V(G),E(G),w_{G})$ with adjacency matrix $\bbA$, it is possible to obtain an \textit{induced} graphon, $W_{G}$, by building a uniform (regular) partition of $[0,1]$ with $N=\vert V(G)\vert$ intervals, $I_{i}$, such that
\begin{equation}\label{eq_W_from_G}
W_{G}(u,v) =
                   \sum_{i,j=1}^{N}\bbA (i,j)\chi_{I_{i}}(u)\chi_{I_{j}}(v)
,
\end{equation}
with $\chi_{I_{i}}(u)$ given by
\begin{equation}
\chi_{I_{i}}(u) = 
                         \begin{cases}
                                    1,\quad \text{if}\quad u\in I_i \\
                                    0,\quad \text{otherwise}
                         \end{cases}
                         ,
\end{equation}
and where the subindex in $I_{i}$ follows an ascending ordering that matches the inherited ordering structure of the lower (or upper) boundaries of the $I_{i}$. This is, we have that
\begin{equation}
\inf\{ I_{i} \} < \inf\{ I_{j} \}~\forall~ i<j
.
\end{equation}
This implies that although the node-labeling of the graphs is arbitrary, such labels are mapped to intervals that inherit the ordering structure of $[0,1]$. The node labeled as $i\in V(G)$ is associated with the interval $I_i$ which has a specific location in $[0,1]$. If one relabels the nodes in $V(G)$, the node $i\in V(G)$ (possibly different from the previous scenario) will be associated with the same subset of $[0,1]$. In~\cite{graphon_pooling_j} it is shown that when considering equipartitions, the signal models and operators defined on the graphon are stable with respect to changes in the labeling.


\subsection{Convolutional Signal Processing on Graphons}

The first convolutional signal processing model on graphons was introduced in~\cite{Diao2016ModelfreeCO} for the analysis of clustering methods on large graphs. Indeed, what today is known as a graphon signal is called in~\cite{Diao2016ModelfreeCO} a \textit{node level statistic}, and the so-called Graphon Fourier transform is nothing but a representation introduced in~\cite{Diao2016ModelfreeCO} that leverages the spectral decomposition of the graphon. These concepts and their extensions were later used in~\cite{graphon_geert_j} to perform polynomial filtering in the context of signal processing over large networks, giving rise to the names \textit{graphon signal processing} and \textit{graphon filtering}~\cite{graphon_geert_j}. In what follows, we will describe convolutional signal processing on graphons which we will refer to as Gphon-SP. For the sake of clarity, we represent signals and operators on graphons using italic-bold notation while reserving only-bold symbols for signals and operators on graphs.

For an arbitrary graphon $W(u,v)$, a signal on $W(u,v)$ is a function\footnote{Notice that for the representation of signals and operators in graphs we used bold symbols while for signals and operators on graphons we use italic bold symbols.} $\boldsymbol{x}: [0,1] \rightarrow \mbR$ in $L_{2} ([0,1])$. To emphasize the fact that $\boldsymbol{x}$ is defined on $W(u,v)$ we represent graphon signals by the pair $(W,\boldsymbol{x})$.

To formally define Gphon-SP, we rely on the use of a \textit{graphon shift operator}, $\boldsymbol{T}_{W}: L_{2}([0,1])\to L_{2}([0,1])$, that is used to write filters as polynomials whose independent variable is $\boldsymbol{T}_{W}$. The standard choice for such operator acts on $(W,\boldsymbol{x})$ to produce the signal $(W,\boldsymbol{y})$ according to
\begin{equation}\label{eq_Tw}
\boldsymbol{y}(u)
     =
      \left(\boldsymbol{T}_{W}\boldsymbol{x}\right)(u)
     =
      \int_{0}^{1}W(u,v)\boldsymbol{x}(v)dv.
\end{equation}
As pointed out in~\cite{lovasz2012large}, for every graphon $W$ the operator $\boldsymbol{T}_{W}$ is normal, compact, and Hilbert-Schmidt.

Similar to the GSP scenario, we can leverage the graphon shift operator, $\boldsymbol{T}_W$, to define the notion of filtering~\cite{graphon_pooling_j,graphon_pooling_c,alejopm_algnn_j}. Then, a convolutional filter $\boldsymbol{H}(\boldsymbol{T}_{W})$ on $W(u,v)$ is given by a polynomial operator
\begin{equation}
\boldsymbol{H}
    (\boldsymbol{T}_{W})
            =
             \sum_{k=0}^{K}h_{k}\boldsymbol{T}_{W}^{k}
             , 
\end{equation}
where $\boldsymbol{T}_{W}^{k}$ indicates the $k$-times composition of $\boldsymbol{T}_{W}$. The filter $\boldsymbol{H}(\boldsymbol{T}_{W})$ acts on the signal $(W,\boldsymbol{x})$ to produce the signal $(W,\boldsymbol{y})$ where
\begin{equation}
\boldsymbol{y}
=
\boldsymbol{H}
    (\boldsymbol{T}_{W})
    \boldsymbol{x}
            =
             \sum_{k=0}^{K}
                  h_{k}\boldsymbol{T}_{W}^{k}
                  \boldsymbol{x}
             . 
\end{equation}
Since the eigenvalues of $\boldsymbol{T}_W$ are countable, it is possible to obtain a Fourier decomposition of the graphon signals. If $\lambda_{i}(\boldsymbol{T}_W)$ and $\boldsymbol{\varphi}_{W,i}$ are the eigenvalues and the eigenvectors of $\boldsymbol{T}_{W}$, respectively, we have that the Fourier transform of $(W,\boldsymbol{x})$ is given by the pair $(\widehat{W},\widehat{\boldsymbol{x}})$, where
\begin{equation}
\widehat{\boldsymbol{x}}(j)
              =
                  \int_{0}^{1}\boldsymbol{x}(u)\boldsymbol{\varphi}_{W,j}(u)du
,
\end{equation}
with
$\widehat{\boldsymbol{x}}\in \ell_2 \left( \mathbb{Z} \right)$ 
and where $\widehat{W}$ remarks that $\widehat{\boldsymbol{x}}$ is defined on the eigenvalues of $\boldsymbol{T}_{W}$ and not in $[0,1]$. We can have a representation of a graphon signal $(\boldsymbol{x}, W)$ in terms of its Fourier transform $(\widehat{W},\widehat{\boldsymbol{x}})$ by taking into account that
\begin{equation}
\boldsymbol{x} 
=
\sum_{i=1}^{\infty}
    \widehat{\boldsymbol{x}}
         \left(
                i
         \right)
         \boldsymbol{\varphi}_{W,i}(u)
,
\end{equation}
where the ordering of the eigenvalues $\lambda_{i}(\boldsymbol{T}_W)$ is typically given according to $\vert\lambda_{1}(\boldsymbol{T}_W)\vert\geq\vert\lambda_{2}(\boldsymbol{T}_W)\vert\geq\cdots$. We say that a graphon signal $\boldsymbol{x}$ is bandlimited with bandwidth $\omega$ if $\hat{\boldsymbol{x}}(i) = 0$ for all $\vert\lambda_{i}(\boldsymbol{T}_W)\vert<\omega$. The set of graphon bandlimited signals with bandwidth $\omega$ is represented by $\mathcal{PW}_{\omega}(W)$.


\subsection{Gphon-SP induced by GSP}

Graph signals like graphs, possess graphon-induced representations~\cite{graphon_pooling_j}. If $(G,\bbx)$ is a graph signal, we say that the graphon signal $(W_{G},\boldsymbol{x})$ is induced by $(G,\bbx)$ if $W_{G}$ is induced by $G$ -- in the sense of~\eqref{eq_W_from_G} -- and
\begin{equation}\label{eq_wsig_from_gsig}
\boldsymbol{x}(u)
    =    
    \bbx\left( 
             \left\lceil 
                u
                \left\vert V(G) \right\vert
             \right\rceil 
        \right)
,
\end{equation}
where $u\in [0,1]$ and $\lceil\cdot\rceil$ is the ceiling operator. Then, the graphon signals induced by graph signals are piece-wise constant functions on a regular partition of $[0,1]$. We will also refer to the operation in~\eqref{eq_wsig_from_gsig} as the \textit{stepping} of $\bbx$ and will denote it by $\boldsymbol{x}=\mathsf{step}\left( \bbx \right)$.

With the notions presented above, we formalize the notion of induced filters in the following result from~\cite{graphon_pooling_j}.


\begin{theorem}[\cite{graphon_pooling_j}]
\label{thm_induced_gphon_filtering}
Let $(W_{G} , \boldsymbol{x})$ be a graphon signal induced by the graph signal $(G,\bbx)$. Let $h(t)=\sum_{k=0}^{K-1}h_{k}t^k$ and $\boldsymbol{y} = h\left( \boldsymbol{T}_{W_G} \right)\boldsymbol{x}$, where $\boldsymbol{T}_{W_G}$ is the graphon shift operator in $W_G$. Then, it follows that
\begin{equation}
\boldsymbol{y}
          =
          \mathsf{step}
          \left(
           h\left( 
                 \frac{\bbS_G}{\vert V(G)\vert}
           \right)\bbx
          \right) 
,
\end{equation}
where $\bbS_G$ is the shift operator on $G$.
\end{theorem}


It is important to remark that if the graphon $W_G$ is induced by the graph $G$, we have $\lambda_i(\boldsymbol{T}_{W_G}) = \lambda_i (\bbS_{G})/\vert V(G)\vert$ for $i=1,\ldots, \vert V(G)\vert$ and $\lambda_i(\boldsymbol{T}_{W_G}) = 0$ for all $i>N$~\cite{lovasz2012large,Diao2016ModelfreeCO}. Additionally, if $\bbv_i$ is the $i$-th eigenvector of $\bbS_{G}$, the $i$-th eigenvector of $\boldsymbol{T}_{W_{G}}$ is given by $\boldsymbol{\varphi}_{W_G,i} (u) =\sqrt{\vert V(G)\vert}\sum_{\ell=1}^{\vert V(G)\vert}\chi_{I_\ell}(u)\bbv_i (\ell) = \sqrt{\vert V(G)\vert}\mathsf{step}(\bbv)$~\cite{gao2019graphon}.





\section{Sampling of Signals: From Graphs to Graphons and Back}
\label{sec_samp_on_graphons}

Now we extend concepts of sampling on graphs to graphons. We start extending the notion of removable sets.


\begin{definition}\label{def_remov_set_gphonsp}
	We say that the open set $\ccalS\subset [0,1]$ is a removable set if there exists $\Lambda>0$ such that

	\begin{equation}\label{eq_def_remov_set_gphonsp}
	\left\Vert
	              \boldsymbol{T}_{W}\boldsymbol{x}
	\right\Vert_{2}
	                      <
	                      \Lambda \Vert\boldsymbol{x}\Vert_{2}
	                      ,
	                      ~\forall~\boldsymbol{x}\in L_2 (\ccalS)
	                      ,
	\end{equation}
where $L_2 (\ccalS)$ is the space of square integrable functions whose support is contained in $\ccalS$. The infimum among these values is denoted by $\Lambda_{\ccalS}$.
\end{definition}


It is worth pointing out that a removable set on graphons is assumed to be open. This excludes subsets of $[0,1]$ consisting of collections of isolated points. To show how this notion of removable set characterizes sampling sets, we need to determine how the bandwidth of a graphon signal relates to the application of the operator $\boldsymbol{T}_W$. This is stated in the following theorem.


\begin{theorem}\label{thm_bernstein_ineq}
Let $(W,\boldsymbol{x})$ be a graphon signal. If $(W,\boldsymbol{x})\in\mathcal{PW}_{\omega}(W)$, it follows that
\begin{equation}
 \left\Vert 
            \boldsymbol{T}_{W}\boldsymbol{x}
 \right\Vert_{2}
             \geq
             \omega \Vert \boldsymbol{x} \Vert_{2}
             .
\end{equation}
\end{theorem}

\begin{proof}
See Appendix~\ref{sec_app_Bernstein_ineq}.
\end{proof}


By leveraging the notion of a removable set for graphons and Theorem~\ref{thm_bernstein_ineq} we introduce our first result on the uniqueness sets of graphon signals.


\begin{theorem}\label{thm_uq_sets_gphonsp}
Let $\ccalS \subset [0,1]$, where $\ccalS^c$ is a removable set with constant $\Lambda_{\ccalS^c}$. Let $(W,\boldsymbol{x})\in\mathcal{PW}_{\omega}(W)$ and $\omega>\Lambda_{\ccalS^c}$. If $\boldsymbol{x} = \boldsymbol{y}$ on $\ccalS$ then $\boldsymbol{x} = \boldsymbol{y}$ on $[0,1]$. We refer to $\ccalS$ as a uniqueness set for the graphon signals in $\mathcal{PW}_{\omega}(W)$.
\end{theorem}

\begin{proof}
    See Appendix~\ref{sec_proof_uniqsets_graphons}.
\end{proof}


Theorem~\ref{thm_uq_sets_gphonsp} allows us to characterize subsets of $[0,1]$ that determine in a unique way elements  of the space $\mathcal{PW}_{\omega}(W)$. One particular useful insight from Theorem~\ref{thm_uq_sets_gphonsp} is that when looking for the uniqueness sets of any $\mathcal{PW}_{\omega}(W)$ one has to aim to minimize $\Lambda_{\ccalS^{c}}$. In the light of~\eqref{eq_def_remov_set_gphonsp} (Def.~\ref{def_remov_set_gphonsp}) this implies minimizing
\begin{equation}
\sup_{\boldsymbol{x}\in L_2 (\ccalS^{c})}
     \frac{
	\left\Vert
	              \boldsymbol{T}_{W}\boldsymbol{x}
	\right\Vert_{2}
	}
	{
	\Vert\boldsymbol{x}\Vert_{2}
	}
	 =
	 \Lambda_{\ccalS^{c}} 
	  .
\end{equation}
As one can expect, the larger the size of $\ccalS$ is, the easier it is to reduce the value of $\Lambda_{\ccalS^{c}}$.

In what follows, we derive other results that will allow us to understand how sampling sets from different graphons and graphs interact between them, in particular how uniqueness sets for different spaces of graphon signals can be inter-related when such uniqueness sets are equal in $[0,1]$.

 
\begin{theorem}\label{thm_removable_sets_seq}
Let $W_{1}$ and $W_{2}$ be two graphons and let $\ccalS_{
W_{1}}\subset [0,1]$ and $\ccalS_{W_{2}}\subset [0,1]$ be uniqueness sets for signals in $W_1$ and $W_2$, respectively, with $\ccalS_{W_{1}} = \ccalS_{W_{2}}$. Then, it follows that 
\begin{multline}\label{eq_thm_removable_sets_seq_1}
\max
\left\lbrace 
0
,
    \Lambda_{\ccalS_{W_{2}}^c}
    -
    \left\Vert
       \boldsymbol{T}_{W_{1}}
        -
       \boldsymbol{T}_{W_{2}}
    \right\Vert_2
\right\rbrace
\leq
\Lambda_{\ccalS_{W_{1}}^c}
\leq
\\
\min
\left\lbrace 
     \Vert \boldsymbol{T}_{W_1} \Vert_2
     ,
     \left\Vert
          \boldsymbol{T}_{W_{1}}
           -
           \boldsymbol{T}_{W_{2}}
     \right\Vert_2
     +
     \Lambda_{\ccalS_{W_{2}}^c}
\right\rbrace     
,
\end{multline}
where $\boldsymbol{T}_{W_{1}}$ and $\boldsymbol{T}_{W_{2}}$ are the graphon shift operators of $W_1$ and $W_2$, respectively.
\end{theorem}
 
\begin{proof}
  See Appendix~\ref{proof_thm_removable_sets_seq}.
\end{proof}
 


\begin{figure}
\centering
	\centering


\makeatletter
\newcommand{\gettikzx}[2]{%
	\tikz@scan@one@point\pgfutil@firstofone#1\relax
	\edef#2{\the\pgf@x}%
}
\makeatother


\makeatletter
\newcommand{\gettikzy}[2]{%
	\tikz@scan@one@point\pgfutil@firstofone#1\relax
	\edef#2{\the\pgf@y}%
}
\makeatother


\definecolor{my_blue}{rgb}{0.0314, 0.3569, 1.0000}

\definecolor{my_cp4_col1}{RGB}{255, 86, 87}
\definecolor{my_cp4_col2}{RGB}{55, 108, 138}
\definecolor{my_cp4_col3}{RGB}{242, 217, 187}
\definecolor{my_cp4_col4}{RGB}{99, 143, 169}

\definecolor{my_cp5_col1}{RGB}{253, 231, 37}
\definecolor{my_cp5_col2}{RGB}{94, 201, 98}
\definecolor{my_cp5_col3}{RGB}{33, 145, 140}
\definecolor{my_cp5_col4}{RGB}{59, 82, 139}
\definecolor{my_cp5_col5}{RGB}{68, 1, 84}


\def\scale{1}
\def\unit{ \scale cm}

\def\my_gap_inter_plots{1cm}

\def\mylinewidth{0.5}

\def\myplotdimx{7cm}
\def\myplotdimy{5cm}

\begin{tikzpicture}



%
 \path[draw, -stealth, line width = 2*\mylinewidth, color=my_cp5_col5] 
	    (0,0) to [] node [] {} (\myplotdimx,0);

\path (\myplotdimx,0)  coordinate (p1);	    
\path (p1) node [below]  {$\Lambda_{\mathcal{S}^{c}_{W_2}}$};	    

%
 \path[draw, -stealth, line width = 2*\mylinewidth, color=my_cp5_col5] 
	    (0,0) to [] node [] {} (0,\myplotdimy);

\path (0,\myplotdimy)  coordinate (p1);	    
\path (p1) node [left]  {$\Lambda_{\mathcal{S}^{c}_{W_1}}$};


\def\c{1cm}
\def\b{4cm}
\def\a{1.1*\b}
\def\radiuscirclecomp{0.05}

\tikzstyle{dot} = [circle,
                    minimum width  = 0.15*\unit,
                    fill=black,
                    color=my_cp5_col4,
                    inner sep=0pt,
                    draw,
                    anchor = center ]

\path[draw, fill=my_cp5_col5, opacity=0.4,line width = 2*\mylinewidth, color=my_cp5_col2] 
	    (0,0) -- (\c,0) -- (\b,\b -\c) -- (\b,\a) -- (\a-\c,\a) -- (0,\c) -- (0,0);


\path (0,0) node [dot] (x) {};
\path (x.south)++(0,-0.1) node [below] {$(0,0)$};   
 
\path (\c,0) node [dot] (x) {};
\path (x.south)++(0,0) node [below right] {$(\Vert \boldsymbol{T}_{W_1}-\boldsymbol{T}_{W_2}\Vert_{2},0)$};  
	    
\path (\b,\b -\c) node [dot] (x) {};
\path (x.south)++(0,0) node [below] {$\displaystyle\left(\Vert \boldsymbol{T}_{W_2}\Vert_{2},\Vert \boldsymbol{T}_{W_2}\Vert_{2}-\Vert \boldsymbol{T}_{W_1}-\boldsymbol{T}_{W_2}\Vert_{2}\right)$}; 	

\path (\b,\a) node [dot] (x) {};
\path (x.south)++(0,-0.1) node [right] {$\displaystyle\left(\Vert \boldsymbol{T}_{W_2}\Vert_{2},\Vert \boldsymbol{T}_{W_1}\Vert_{2}\right)$}; 	

\path (\a-\c,\a) node [dot] (x) {};
\path (x.south)++(0,0.2) node [above] {$\displaystyle\left(\Vert \boldsymbol{T}_{W_1}\Vert_{2}-\Vert \boldsymbol{T}_{W_1}-\boldsymbol{T}_{W_2}\Vert_{2},\Vert \boldsymbol{T}_{W_1}\Vert_{2}\right)$}; 
	
\path (0,\c) node [dot] (x) {};	
\path (x.south)++(0,0.1) node [above right] {$\displaystyle\left(0, \Vert \boldsymbol{T}_{W_1}-\boldsymbol{T}_{W_2}\Vert_{2} \right)$};

\end{tikzpicture} 
\caption{
Depiction of the admissible values of $\Lambda_{\ccalS_{W_1}^{c}}$ and $\Lambda_{\ccalS_{W_2}^{c}}$ when $\ccalS_{W_1}=\ccalS_{W_2}$ considering two graphon signal models associated with the graphons $W_{1}$ and $W_{2}$ -- Theorem~\ref{thm_removable_sets_seq}. A small value of $\Vert \boldsymbol{T}_{W_1}-\boldsymbol{T}_{W_1}\Vert_{2}$ naturally narrows the number of admissible values of $\Lambda_{\ccalS_{W_2}^{c}}$ given a fixed value of $\Lambda_{\ccalS_{W_1}^{c}}$.
}
\label{fig_admissible_two_uniqueness_set_W}
\end{figure}


Theorem~\ref{thm_removable_sets_seq} emphasizes insights that are central to our discussion. First, it highlights that a given subset $\ccalS\subset [0,1]$ can be the uniqueness set of graphon signals defined on graphons $W_1$ and $W_{2}$, that can be independent of each other. Second, Theorem~\ref{thm_removable_sets_seq} shows that even though both $W_1$ and $W_2$ are arbitrary, the $\Lambda$-removable constants of $\ccalS$ as a uniqueness set of $W_1$ and $W_2$, are related by a closed form expression. Figure~\ref{fig_admissible_two_uniqueness_set_W} shows a geometric depiction of the admissible values of $\Lambda_{\ccalS_{W_1}}$ and $\Lambda_{\ccalS_{W_2}}$ in~\eqref{eq_thm_removable_sets_seq_1}. Notice that~\eqref{eq_thm_removable_sets_seq_1} can be re-written as
\begin{multline}
\max
\left\lbrace 
0
,
    \Lambda_{\ccalS_{W_{1}}^c}
    -
    \left\Vert
       \boldsymbol{T}_{W_{1}}
        -
       \boldsymbol{T}_{W_{2}}
    \right\Vert_2
\right\rbrace
\leq
\Lambda_{\ccalS_{W_{2}}^c}
\leq
\\
\min
\left\lbrace 
     \Vert \boldsymbol{T}_{W_2} \Vert_2
     ,
     \left\Vert
          \boldsymbol{T}_{W_{1}}
           -
           \boldsymbol{T}_{W_{2}}
     \right\Vert_2
     +
     \Lambda_{\ccalS_{W_{1}}^c}
\right\rbrace     
.
\end{multline}
%
%
%


\begin{figure*}
\centering
	\centering
 \input{./figures/fig_2_tikz_source.tex} 
\caption{
Exemplification of how sampling sets from graphons with a different number of nodes can be identical under a graphon representation. $\ccalS_{G_1}$ is a sampling set on the graph $G_1$ (left) with four nodes. $\ccalS_{G_1}$ contains nodes labeled as $3$ and $4$ and it induces (Def.~\ref{def_induced_gphon_samp_set}) the graphon sampling set $\ccalS_{W_{G_1}}\subset [0,1]$. In the graph $G_2$, the sampling set $\ccalS_{G_2}$ consisting of the nodes labeled as $5$, $6$, $7$, and $8$, induces the graphon sampling set $\ccalS_{W_{G_2}} = \ccalS_{W_{G_1}}$.
}
\label{fig_inter_two_samp_sets_gphonspace}
\end{figure*}


Now we turn our attention to the relationship between removable sets on graphs and their induced graphon counterparts. To this end, we introduce some notation in the following definition.


\begin{definition}\label{def_induced_gphon_samp_set}
Let $G$ be a graph and let $W_{G}$ be its induced graphon. Let $\ccalS_{G}\subset V(G)$ and let $(G,\bbx)$ be a binary graph signal -- $\bbx(i)\in\{ 0,1\}$ -- with $\text{supp}(\bbx)=\ccalS_G$. If $(W_{G},\boldsymbol{x})$ is the induced graphon signal of $(G,\bbx)$ we say that $\ccalS_{W_{G}} = \text{supp}(\boldsymbol{x})$ is the set induced by $\ccalS_{G}$.
\end{definition}


The concept introduced in Definition~\ref{def_induced_gphon_samp_set} allows us to directly interpret and represent subsets of nodes of a graph, $G$, on the induced graphon $W_{G}$. In Fig.~\ref{fig_inter_two_samp_sets_gphonspace} we depict examples of induced sets in $[0,1]$ obtained from subsets of nodes in different graphs. Taking into account this terminology, we are ready to state a result that relates removable sets on graphs with the removable sets of their induced graphon counterparts.


\begin{theorem}\label{thm_removable_sets_G_W}
Let $\ccalS_G\subset V(G)$ be a removable set in $G = (V(G), E(G),w_{G})$, with constant $\Lambda_{\ccalS_{G}}$. Let $W_G$ be the graphon induced by $G$ and let $\ccalS_{W_G}\subset [0,1]$ be the removable set induced by $\ccalS_{G}$. Then, it follows that
\begin{equation}\label{eq_thm_removable_sets_G_W}
   \vert V(G) \vert \Lambda_{\ccalS_{W_G}}
         = 
          \Lambda_{\ccalS_G}
   .
\end{equation}
\end{theorem}

\begin{proof}
   See Appendix~\ref{sec_proof_LambdaG_vs_LambdaW}
\end{proof}


With the quantitative closed form expression between the removable constants of sampling sets in graphs and their induced graphons in~\eqref{eq_thm_removable_sets_G_W}, we can take into account Theorem~\ref{thm_removable_sets_seq} to relate uniqueness sets of signals from different graphs. We emphasize this point in the following theorem.


\begin{theorem}\label{thm_removable_sets_twoG}
Let $G_1$ and $G_2$ be two graphs and let $ W_{G_i}$ be the graphon induced by $G_i$ for $i=1,2$. Let $\ccalS_{G_1}$ and $\ccalS_{G_2}$ be sampling sets in $G_1$ and $G_2$, and let $\ccalS_{W_{G_1}}$ and $\ccalS_{W_{G_2}}$ their induced sets on the graphons $W_{G_1}$ and $W_{G_2}$. If $\ccalS_{W_{G_1}}=\ccalS_{W_{G_2}}$ it follows that
\begin{equation}\label{eq_thm_removable_sets_twoG_1}
\theta_1
       \leq 
           \Lambda_{\ccalS_{G_{1}}^c} 
       \leq    
\theta_2       
,
\end{equation}
where
\begin{equation}\label{eq_thm_removable_sets_twoG_2}
\theta_1
    =
    \max
\left\lbrace 
0
,
    \frac{\vert V(G_1)\vert}{\vert V(G_2)\vert}
    \Lambda_{\ccalS_{G_{2}}^c}
    -
    \vert V(G_1) \vert
     \left\Vert
          \boldsymbol{T}_{W_{1}}
           -
           \boldsymbol{T}_{W_{2}}
     \right\Vert_2
\right\rbrace
,
\end{equation}
%
%
%
%
%
%
\begin{multline}\label{eq_thm_removable_sets_twoG_3}
 \theta_2
    =
    \min
\left\lbrace 
     \Vert \boldsymbol{T}_{W_1} \Vert_2
     \vert V(G_1) \vert
     ,
     \vert V(G_1) \vert 
     \left\Vert
          \boldsymbol{T}_{W_{1}}
           -
           \boldsymbol{T}_{W_{2}}
     \right\Vert_2
\right.
\\
\left.
     +
     \frac{\vert V(G_1)\vert}{\vert V(G_2)\vert}
     \Lambda_{\ccalS_{G_{2}}^c}
\right\rbrace     
.
\end{multline}

\end{theorem}

\begin{proof}
    See Appendix~\ref{proof_thm_removable_sets_twoG}
\end{proof}



\begin{figure}
\centering
	\centering


\makeatletter
\newcommand{\gettikzx}[2]{%
	\tikz@scan@one@point\pgfutil@firstofone#1\relax
	\edef#2{\the\pgf@x}%
}
\makeatother


\makeatletter
\newcommand{\gettikzy}[2]{%
	\tikz@scan@one@point\pgfutil@firstofone#1\relax
	\edef#2{\the\pgf@y}%
}
\makeatother


\definecolor{my_blue}{rgb}{0.0314, 0.3569, 1.0000}

\definecolor{my_cp4_col1}{RGB}{255, 86, 87}
\definecolor{my_cp4_col2}{RGB}{55, 108, 138}
\definecolor{my_cp4_col3}{RGB}{242, 217, 187}
\definecolor{my_cp4_col4}{RGB}{99, 143, 169}

\definecolor{my_cp5_col1}{RGB}{253, 231, 37}
\definecolor{my_cp5_col2}{RGB}{94, 201, 98}
\definecolor{my_cp5_col3}{RGB}{33, 145, 140}
\definecolor{my_cp5_col4}{RGB}{59, 82, 139}
\definecolor{my_cp5_col5}{RGB}{68, 1, 84}


\def\scale{1}
\def\unit{ \scale cm}

\def\my_gap_inter_plots{1cm}

\def\mylinewidth{0.5}

\def\myplotdimx{7cm}
\def\myplotdimy{4cm}

\begin{tikzpicture}



%
 \path[draw, -stealth, line width = 2*\mylinewidth, color=my_cp5_col5] 
	    (0,0) to [] node [] {} (\myplotdimx,0);

\path (\myplotdimx,0)  coordinate (p1);	    
\path (p1) node [below]  {$\Lambda_{\mathcal{S}^{c}_{G_2}}$};	    

%
 \path[draw, -stealth, line width = 2*\mylinewidth, color=my_cp5_col5] 
	    (0,0) to [] node [] {} (0,\myplotdimy);

\path (0,\myplotdimy)  coordinate (p1);	    
\path (p1) node [left]  {$\Lambda_{\mathcal{S}^{c}_{G_1}}$};


\def\radiuscirclecomp{0.05}

\tikzstyle{dot} = [circle,
                    minimum width  = 0.15*\unit,
                    fill=black,
                    color=my_cp5_col4,
                    inner sep=0pt,
                    draw,
                    anchor = center ]

\def\Qa{1}
\def\Qb{1.1}
\def\c{0.3}
\def\a{3}
\def\b{2*\a}

\path[draw, fill=my_cp5_col5, opacity=0.4,line width = 2*\mylinewidth, color=my_cp5_col2] 
	    (0,0) -- (\b*\c,0) -- (\Qb*\b,\Qb*\a-\a*\c) -- (\Qb*\b,\Qa*\a) -- (\b*\Qa-\b*\c,\Qa*\a) -- (0,\a*\c) -- (0,0);

\path (5.3,1)  coordinate (p1);	    
\path (p1)+(0,0) node []  {$N_{2}= 2N_{1}$};

%
%
%
%
%
%


\def\b{0.6*\a}

\path[draw, fill=my_cp5_col5, opacity=0.4,line width = 2*\mylinewidth, color=my_cp5_col4] 
	    (0,0) -- (\b*\c,0) -- (\Qb*\b,\Qb*\a-\a*\c) -- (\Qb*\b,\Qa*\a) -- (\b*\Qa-\b*\c,\Qa*\a) -- (0,\a*\c) -- (0,0);

\path (0.5,3.2)  coordinate (p1);	    
\path (p1)+(1,0.2) node []  {$N_{2} = 0.6N_{1}$};


%
 \path[draw, dashed, line width = 2*\mylinewidth, color=my_cp5_col5] 
	    (0,0) to [] node [] {} (\myplotdimy,\myplotdimy);

\path (\myplotdimy,\myplotdimy)  coordinate (p1);	    
\path (p1)+(-0.1,0) node []  {$\Lambda_{\mathcal{S}^{c}_{G_1}}=\Lambda_{\mathcal{S}^{c}_{G_2}}$};	    

%

\end{tikzpicture} 
\caption{
Depiction of the two cases of admissible regions in~\eqref{eq_thm_removable_sets_twoG_1} in Theorem~\ref{thm_removable_sets_twoG}. Letting $N_{1}=\vert V(G_1)\vert$ and $N_{2}=\vert V(G_2)\vert$ 
we consider fixed values for $\Vert \boldsymbol{T}_{W_1}-\boldsymbol{T}_{W_2}\Vert_{2}$, $\Vert \boldsymbol{T}_{W_1}\Vert_{2}$ and $\Vert \boldsymbol{T}_{W_2}\Vert_{2}$. As can be observed when the value of $N_2$ increases with respect to $N_1$ the possible admissible values of $\Lambda_{\ccalS_{G_1}^{c}}$ are limited to a fraction those possible values of $\Lambda_{\ccalS_{G_2}^{c}}$.
}
\label{fig_admissible_two_uniqueness_set_G}
\end{figure}


Theorem~\ref{thm_removable_sets_twoG} has fundamental implications in concrete scenarios where we wish to study and compare the uniqueness sets on two graphs whose global behavior is similar. First, we emphasize that Theorem~\ref{thm_removable_sets_twoG} is most useful when
$
\left\Vert
          \boldsymbol{T}_{W_{1}}
           -
           \boldsymbol{T}_{W_{2}}
\right\Vert_2
$
is small. When this is the case, for a fixed value of $\vert V(G_1) \vert$ we observe that if $\vert V(G_2)\vert \gg \vert V(G_1)\vert$ the value of $\theta_2$ becomes a fraction of $\Lambda_{S_{G_2}^{c}}$. This implies that the bandwidth of signals on $G_1$ uniquely represented in $\ccalS_{G_1}$ is larger than the bandwidth of signals on $G_2$ that are uniquely represented in $\ccalS_{G_2}$. This phenomenon is illustrated in Fig.~\ref{fig_admissible_two_uniqueness_set_G}. From this point, the fundamental aspect to highlight is that having a universal graphon representation allows us to guarantee that the properties of a subset of nodes in a graph are \textit{inherited} by another graph whose structural properties are similar, and such structural similarity is measured by $\Vert \boldsymbol{T}_{W_{G_1}} - \boldsymbol{T}_{W_{G_2}}\Vert_{2}$. The implications of this observation are important since they mean that one can translate or transfer the results obtained by a sampling method between different graphs for which $\Vert \boldsymbol{T}_{W_{G_1}} - \boldsymbol{T}_{W_{G_2}}\Vert_{2}$ is small. In a practical scenario, this means one can compute a sampling set in a small graph and then reuse it to find an approximately optimal sampling set in a graph of large size. We leverage this in the development of Algorithm~\ref{alg_sampt_method} which we will introduce later.

Using Theorem~\ref{thm_removable_sets_twoG} we can also study the behavior of uniqueness sets of sequences of graphs that share common structural properties and that converge to a certain limit. In the following result, we provide some insights in this direction.


\begin{theorem}\label{thm_uniq_remov_sequence}
Let $\{ G_i \}_{i=1}^{\infty}$ be a sequence of graphs obtained from a graphon $W$ using GD1 and let $W_{G_i}$ be the induced graphon by $G_i$ for each $i\in\mbN$. Let $\ccalS_{G_i}$ be a uniqueness subset on $G_i$ for $i\in\mbN$ and let $\ccalS_{W_{G_i}}$ be its induced subset in $[0,1]$. If $\ccalS_{W_{G_i}}=\ccalS_{W}$ for all $i$ with $\ccalS_W$ fixed, and
$     
\left\Vert
          \boldsymbol{T}_{W}
           -
           \boldsymbol{T}_{W_{G_i}}
\right\Vert_2 \to 0
$
as $i\to\infty$, it follows that
\begin{equation}
\lim_{i\to\infty}
      \Lambda_{\ccalS_{W_{G_i}}^{c}}
      =
      \Lambda_{\ccalS_{W}^{c}}
      .
\end{equation}
\end{theorem}

\begin{proof}
    See Appendix~\ref{proof_thm_uniq_remov_sequence}
\end{proof}


Theorem~\ref{thm_uniq_remov_sequence} formalizes the fact that when sequences of graphs are similar on the limit, the uniqueness sets converge as well when represented on a graphon domain. Although this result is somewhat intuitive, it formally shows how such convergence is quantified.


\subsection{Approximation of Optimal Sampling Sets in Large Graphs}

Taking into account the results introduced above, we introduce Algorithm~\ref{alg_sampt_method} intending to show in concrete steps how to reuse optimal sampling sets from small graphs in large graphs.


\begin{algorithm}
\caption{}\label{alg_sampt_method}
\begin{algorithmic}
\Require Uniqueness set $\ccalS_{G_\ell}$ for specific $G_\ell$, size of desired sampling set $m$
\Ensure Sampling set $\ccalS_{G_k}$ for arbitrary $G_k$ with $\vert \ccalS_{G_k}\vert = m$
\State Get $\ccalS_{G_\ell}$ using for instance~\cite{ortega_proxies}
\State Map $\ccalS_{G_\ell}$ into $[0,1]$ to obtain $\ccalS_{W_{G_\ell}}$
\State Build equipartition of $[0,1]$ into $\vert V(G_k)\vert$ intervals $I_i$
\State $\texttt{aux} \gets 0$
\State $\ccalS \gets \emptyset$
\State $\ccalA \gets \emptyset$
\For{$i=1,\dots,\vert V(G_k)\vert$}
\If{$I_i \subset\ccalS_{W_{G_\ell}}$}
      \State $\ccalS \gets \ccalS\bigcup \{ i \}$
      \State $\texttt{aux} \gets \texttt{aux}+1$
             \If{$\texttt{aux} = m$}
                 \State Break
             \EndIf    
\ElsIf{$I_i \bigcap\ccalS_{W_{G_\ell}}\neq\emptyset$}  
      \State $\ccalA \gets \ccalA\bigcup \{ i \}$
\EndIf
\EndFor
\State Choose $\ccalU\subset\ccalA$ with $\vert \ccalU\vert =m-\texttt{aux}$
\State $\ccalS \gets \ccalS\bigcup\ccalU$
\end{algorithmic}
\end{algorithm}


As can be observed, Algorithm~\ref{alg_sampt_method} requires the optimal sampling set, $\ccalS_{G_\ell}$, of a small graph, $G_{\ell}$. The goal is to determine an approximately optimal sampling set in a graph $G_{k}$, where $\vert V(G_k)\vert>\vert V(G_\ell)\vert$ assuming $G_{k}$ shares structural properties with $G_{\ell}$. Then, the algorithm proceeds as follows. We obtain $\ccalS_{W_{G_\ell}}$ which is the set in $[0,1]$ induced by $\ccalS_{G_\ell}$. Then, we build a regular partition of $[0,1]$ into $\vert V(G_k)\vert$ intervals that we denote by $I_i$. We select the subindices of those $I_{i}$ that are contained in $\ccalS_{W_{G_\ell}}$ and store them in the set $\ccalS$. If $\vert \ccalS\vert = m$ then the approximately optimal sampling set in $G_{k}$ is given by $\ccalS_{G_k}=\ccalS$. If $\vert \ccalS\vert<m$ we look for those intervals $I_i$ that are not contained in $\ccalS_{W_{G_\ell}}$ but whose intersection with $\ccalS_{W_{G_\ell}}$ is nonempty, and we store the subindices of those $I_i$ in $\ccalA$. Then, we choose a subset $\ccalU\subset\ccalA$ to complete $\ccalS$.

It is important to emphasize that Algorithm~\ref{alg_sampt_method} runs on values of $m$ that do not exceed excessively the original sampling rate. For instance, if it happens that $\vert \ccalS_{G_\ell}\vert$ equals $5\%$ of $\vert V(G_\ell)\vert$, then Algorithm~\ref{alg_sampt_method} is expected to run on values of $m$ that do not exceed $5\%$ of $\vert V(G_k)\vert$. If it is the case that $m$ is larger than $0.05\vert V(G_k)\vert$, it is expected that at least $\ccalA$ is nonempty, i.e. the intervals that are not contained in $\ccalS_{W_{G_\ell}}$ intersect with it. Notice that these aspects are contained in the results presented in previous subsections.


\begin{remark}\label{rmk_node_labeling_in_algorithm}\normalfont 

We emphasize that the quality of the sampling sets obtained in Algorithm~\ref{alg_sampt_method} rely on the results/theorems derived above. Consequently, running Algorithm~\ref{alg_sampt_method} makes sense when $\Vert \boldsymbol{T}_{W_{G_1}} - \boldsymbol{T}_{W_{G_2}}\Vert_{2}$ is small. In fact, if that is the case, the quality of a sampling set approximately transfers between the two graphs. This scenario includes cases where the graphs are obtained from totally different means and may have arbitrary node labeling. Additionally, notice that when the graphs under consideration are obtained from the same graphon using the discretization method (GD1), we end up naturally in a scenario where $\Vert \boldsymbol{T}_{W_{G_1}} - \boldsymbol{T}_{W_{G_2}}\Vert_{2}$ is small, which is partly explained by how (GD1) preserves well spectral attributes of graphons and its stability to node relabelings~\cite{graphon_pooling_j}.

\end{remark}



\subsection{ Edge Dropping Effects on the Sampling Sets}
\label{sub_sec_edgedroppings}


\begin{figure*}
	\centering
	\centering
	\input{./figures/fig_5_tikz_source.tex} 
	\caption{Depiction of the effects of node relabeling and edge dropping on the induced representation of sampling sets in the graphon domain. The graph $G_1$ (left) is a node-relabeled version of the graph $G_2$ (center). Since the sampling sets $\ccalS_{G_1}\subset V(G_1)$ and $\ccalS_{G_2}\subset V(G_2)$ contain the labels $\{ 5, 6, 7, 8 \}$, they induce the same graphon sampling set in $[0,1]$, i.e. $\ccalS_{W_{G_1}}=\ccalS_{W_{G_2}}$. The graph $G_3$ (right) is obtained from $G_2$ (center) by dropping some edges and keeping the same node labeling. This guarantees that $\ccalS_{W_{G_2}}=\ccalS_{W_{G_3}}$.}
	\label{fig_relabel_edgedrop_samp_gphon}
\end{figure*}


The results presented in Theorem~\ref{thm_removable_sets_twoG} allow us to study and quantify the effects of edge dropping on a given sampling set. To see this, let $G_1$ be a graph and let $G_2$ be the graph obtained from $G_1$ after dropping some edges and where we preserve the same node labeling -- see Fig.~\ref{fig_relabel_edgedrop_samp_gphon}. Using the notation $V(G_1)=V(G_2)=\ccalV$, let $\ccalS_{G_1}\subset\ccalV$ be a sampling set in $G_1$ and let $\ccalS_{G_2}=\ccalS_{G_1}$ be a sampling set in $G_2$. As shown in~\cite{graphon_pooling_j} the effects of edge dropping can be quantified by the term $\Vert \boldsymbol{T}_{W_{G_1}} - \boldsymbol{T}_{W_{G_2}} \Vert_{2}$. If  $\Vert \boldsymbol{T}_{W_{G_1}} - \boldsymbol{T}_{W_{G_2}} \Vert_{2}<\epsilon$, then in the light of Theorem~\ref{thm_removable_sets_twoG} it follows that
\begin{multline}\label{eq_edge_dropping_sec_1}
    \max
\left\lbrace 
0
,
    \Lambda_{\ccalS_{G_{2}}^c}
    -
    \vert \ccalV \vert
    \epsilon
\right\rbrace
       \leq 
           \Lambda_{\ccalS_{G_{1}}^c} 
       \leq    
\\
    \min
\left\lbrace 
     \Vert \boldsymbol{T}_{W_1} \Vert_2
     \vert \ccalV \vert
     ,
     \vert \ccalV \vert 
\epsilon
\right.
\left.
     +
     \Lambda_{\ccalS_{G_{2}}^c}
\right\rbrace        
,
\end{multline}
which implies that the removable constants that characterize the sampling sets are stable\footnote{This is the same notion of stability to perturbations/deformations considered considered in~\cite{alejopm_algnn_j,alejopm_algnn_c,alejopm_algnn_nc_j,alejopm_agggnn_j}.} to edge droppings in the sense that the change in the constant $\Lambda_{\ccalS^{c}_{G_1}}$ is proportional to the size of the edge dropped.


\subsection{Node Relabeling  Effects on the Sampling Sets}
\label{sub_sec_noderelabeling}

We can also leverage Theorem~\ref{thm_removable_sets_twoG} to understand the effect of node relabeling on the constants that characterize sampling sets. We can measure node relabeling with the term $\Vert \boldsymbol{T}_{W_{G_1}} - \boldsymbol{T}_{W_{G_2}} \Vert_{2}$~\cite{graphon_pooling_j}.  To see this, let us first introduce some notation. Let $G_1$ be a graph and let $G_2$ be the graph obtained from $G_1$ 
by node relabeling. This is, $V(G_1)$ and $V(G_2)$ are identical as sets, but the elements in $V(G_1)$ and $V(G_2)$ have different labels. We indicate this with the symbols $V(G_1)\equiv(\ccalV,\boldsymbol{\ell}_{1})$ and $V(G_2)\equiv(\ccalV,\boldsymbol{\ell}_{2})$, where $\ccalV$ is the set of nodes and $\boldsymbol{\ell}_{1}, \boldsymbol{\ell}_{2}$ indicate particular labelings of the elements in $\ccalV$ considering the labels $\{ 1, 2, \ldots, \vert \ccalV\vert \}$. Let $\ccalS_{G_1}$ and $\ccalS_{G_2}$ be sampling sets in $G_1$ and $G_2$, respectively. Additionally, let us consider that the difference between the labelings $\boldsymbol{\ell}_1$ and $\boldsymbol{\ell}_2$ is measured by $\Vert \boldsymbol{T}_{W_{G_1}} - \boldsymbol{T}_{W_{G_2}} \Vert_{2}<\delta$. Then, using Theorem~\ref{thm_removable_sets_twoG} we can see that if $\ccalS_{G_1}\equiv (\ccalS_1,\boldsymbol{\ell}_{0})$ and $\ccalS_{G_2}\equiv (\ccalS_2,\boldsymbol{\ell}_{0})$ we have
\begin{multline}\label{eq_edge_dropping_sec_1}
    \max
\left\lbrace 
0
,
    \Lambda_{\ccalS_{G_{2}}^c}
    -
    \vert \ccalV \vert
    \delta
\right\rbrace
       \leq 
           \Lambda_{\ccalS_{G_{1}}^c} 
       \leq    
\\
    \min
\left\lbrace 
     \Vert \boldsymbol{T}_{W_1} \Vert_2
     \vert \ccalV \vert
     ,
     \vert \ccalV \vert 
\delta
\right.
\left.
     +
     \Lambda_{\ccalS_{G_{2}}^c}
\right\rbrace        
,
\end{multline}
where $\ccalS_1 \subset \ccalV$ and $\ccalS_2 \subset\ccalV$ are not necessarily equal, and $\boldsymbol{\ell}_{0}$ is a particular labeling of $\ccalS_1$ and $\ccalS_2$. This is, it is precisely the fact that we are using the same labeling, $\boldsymbol{\ell}_{0}$, for both sets what guarantees that $\ccalS_{W_{G_1}} = \ccalS_{W_{G_2}}$ and therefore allows us to leverage Theorem~\ref{thm_removable_sets_twoG} -- see Fig.~\ref{fig_relabel_edgedrop_samp_gphon}. Then, in this context we can argue that the constants that characterize sampling sets are stable to relabelings of the node set as long as the collection of labels used in the sampling sets remains the same, even when the node set changes itself. Such stability guarantees that the change in the sampling set is proportional to the change in the relabeling given by $\delta$.

\section{Numerical Simulations}
\label{sec_num_sim}


\begin{figure*}
%
%
\centering
\begin{subfigure}{.32\linewidth}
	   \centering
          \includegraphics[width=1\textwidth]{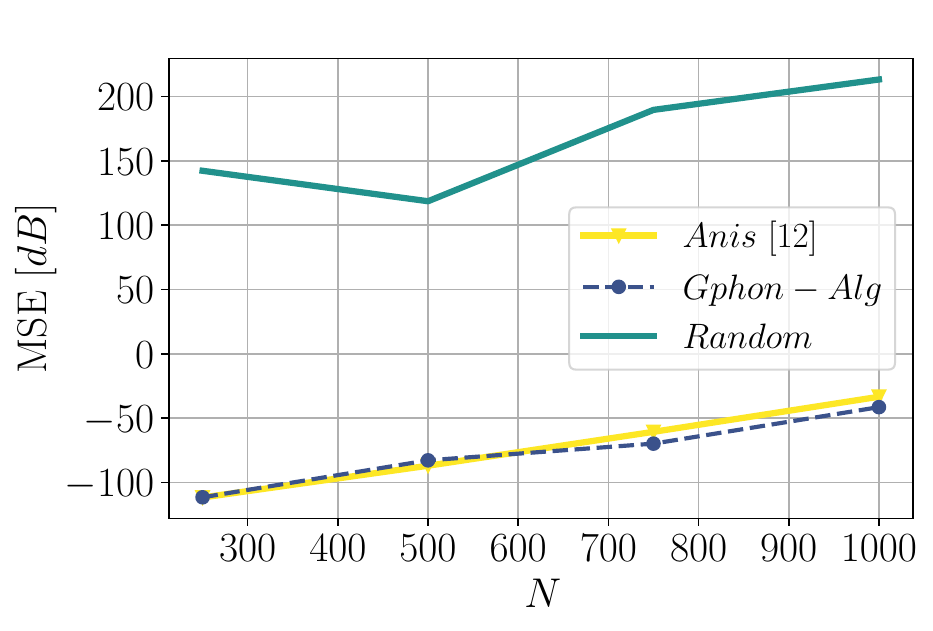} 
\end{subfigure}
\begin{subfigure}{.32\linewidth}
		\centering
            \includegraphics[width=1\textwidth]{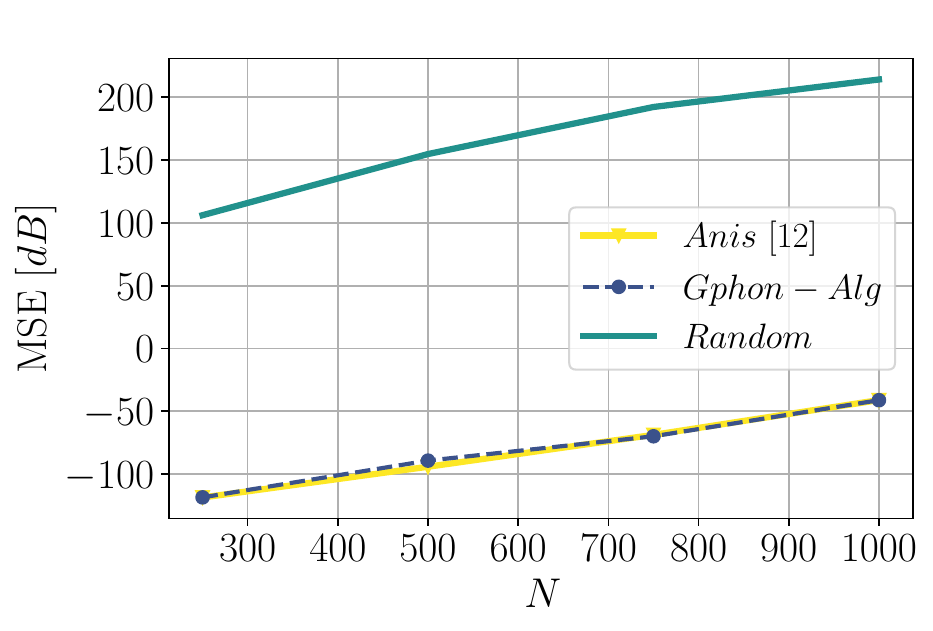} 
\end{subfigure}
    	\begin{subfigure}{.32\linewidth}
		\centering
\includegraphics[width=1\textwidth]{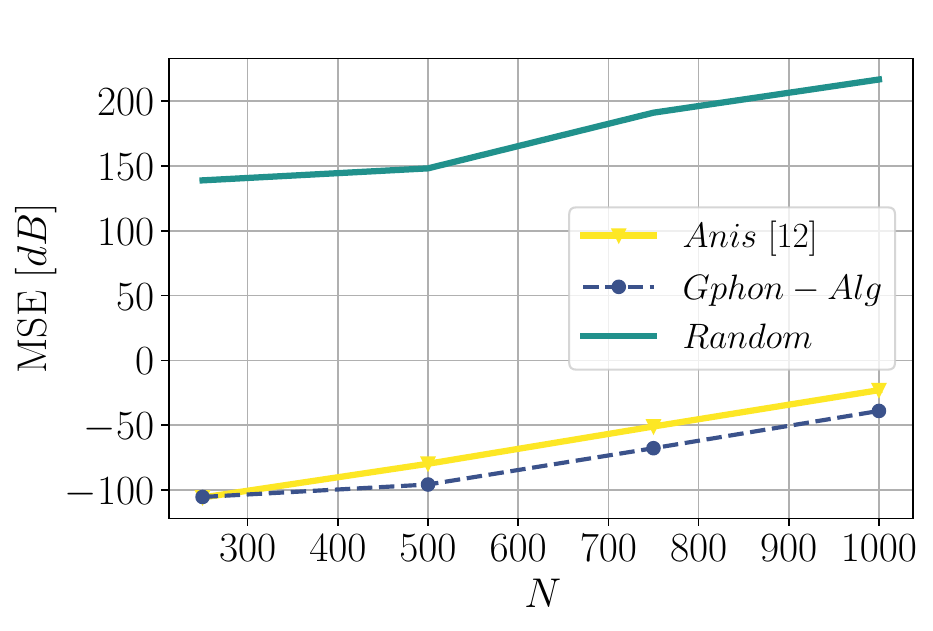} 
\end{subfigure}
%
%
\centering
\begin{subfigure}{.32\linewidth}
        \centering
        \includegraphics[width=1\textwidth]{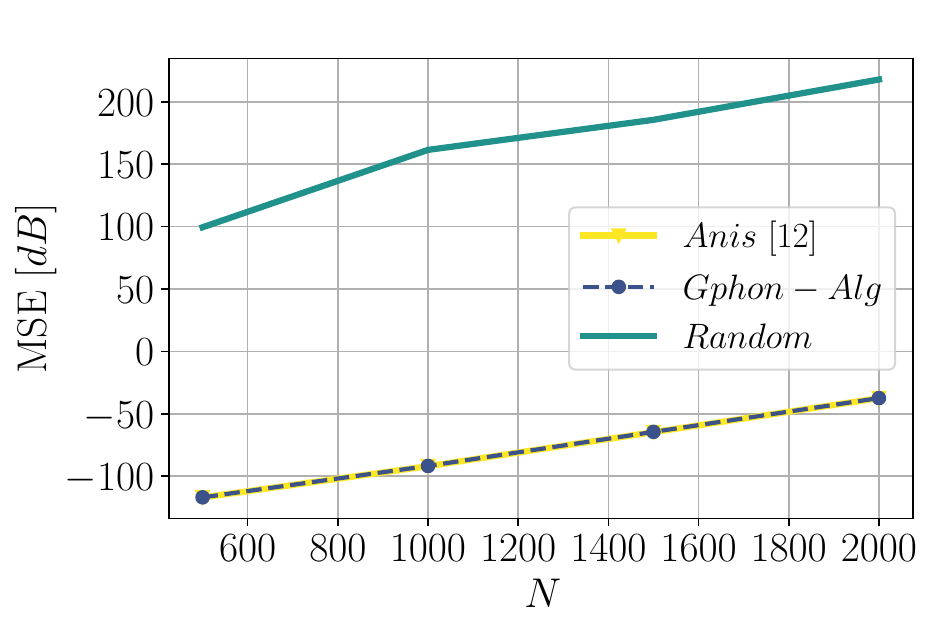} 
        \caption*{$W_{1}(u,v)=(u+v)/2$}
\end{subfigure}
\begin{subfigure}{.32\linewidth}
		\centering
            \includegraphics[width=1\textwidth]{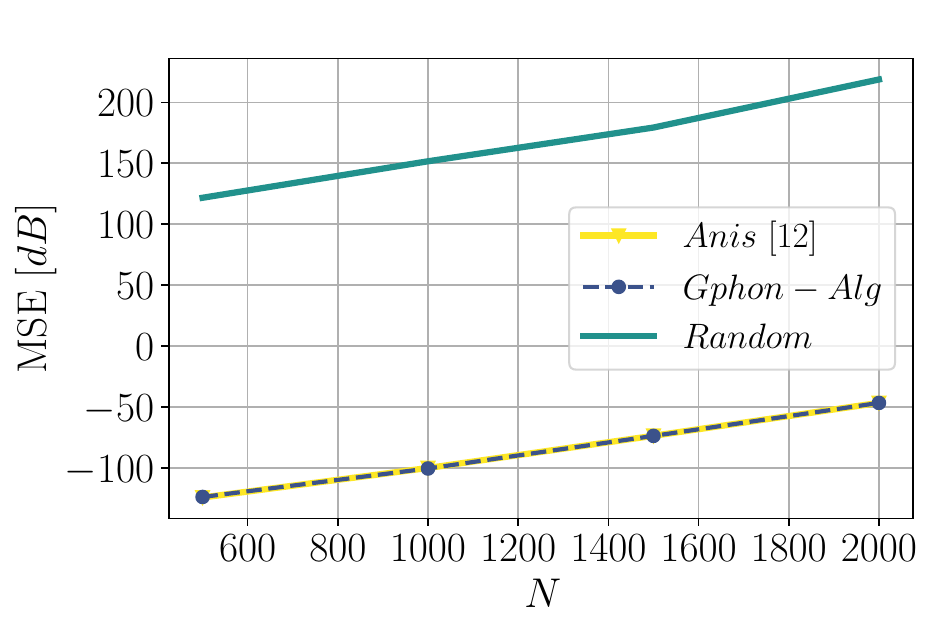} 
            \caption*{$W_{2}(u,v)=\left(u^{2}+v^{2}\right)/2$}
\end{subfigure}
\begin{subfigure}{.32\linewidth}
		\centering
            \includegraphics[width=1\textwidth]{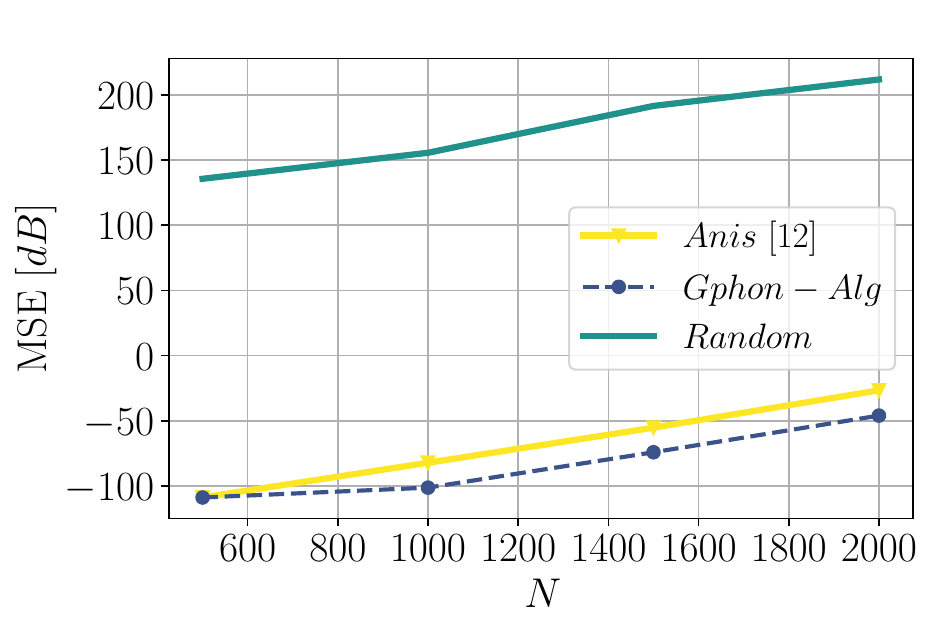} 
            \caption*{$W_{3}(u,v)=1-\max(u,v)$}
\end{subfigure}
    \caption{Reconstruction error of sampled bandlimited signals on graphs derived from a graphon. Each column is associated with experiments performed with a particular graphon, $W_1$ for the left column, $W_2$ for the centered column, and $W_3$ for the column on the right. In the first row, the graphs generated have $250, 500, 750$ and $1000$ nodes. In the second row, the graphs generated have $500, 1000, 1500$ and $2000$ nodes. The graphs are obtained from the graphon using the discretization method (GD1). The bandwidth model considered in this figure is~\textbf{BWM1}. The axis $N$ indicates the number of nodes in the graph. The MSE value depicted is averaged over $50$, which is the number of signals used for each reconstruction. The sampling rate is $5\%$.}
    \label{fig_error_rec_exp_1}
\end{figure*}



\begin{figure*}
%
%
    \centering
       \begin{subfigure}{.32\linewidth}
	   \centering
          \includegraphics[width=1\textwidth]{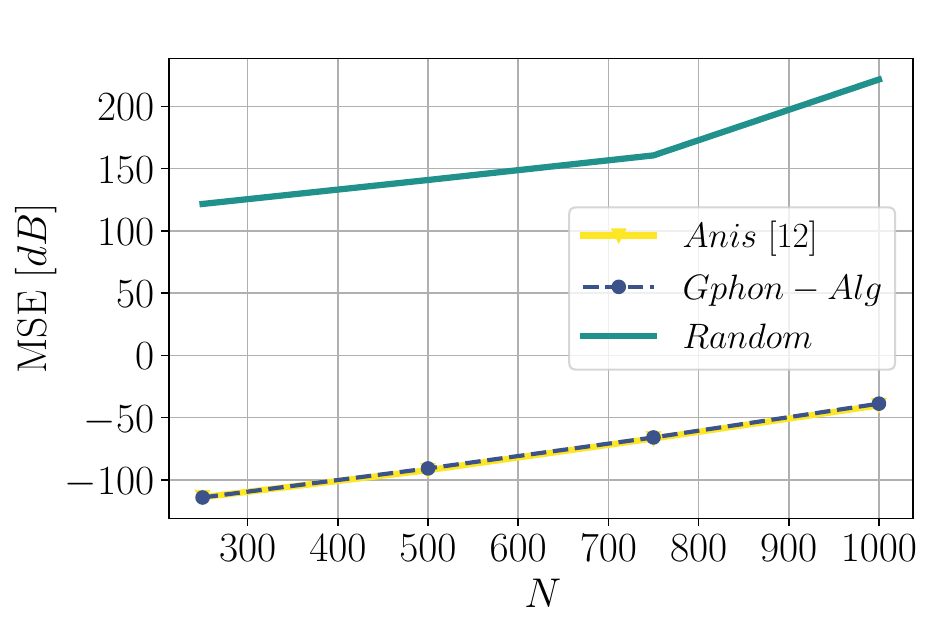} 
       \end{subfigure}
    	\begin{subfigure}{.32\linewidth}
		\centering
            \includegraphics[width=1\textwidth]{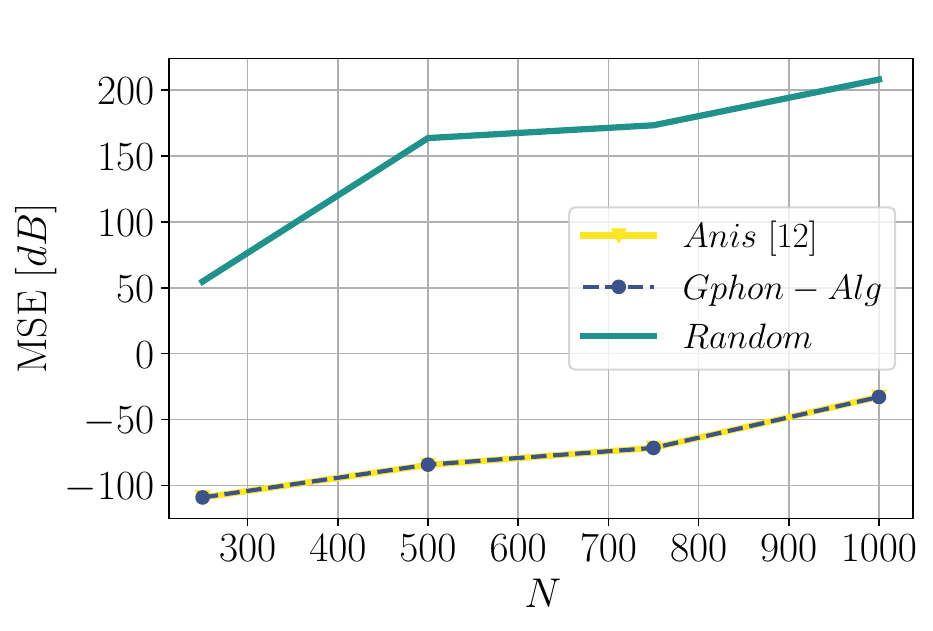} 
\end{subfigure}
    	\begin{subfigure}{.32\linewidth}
		\centering
  \includegraphics[width=1\textwidth]{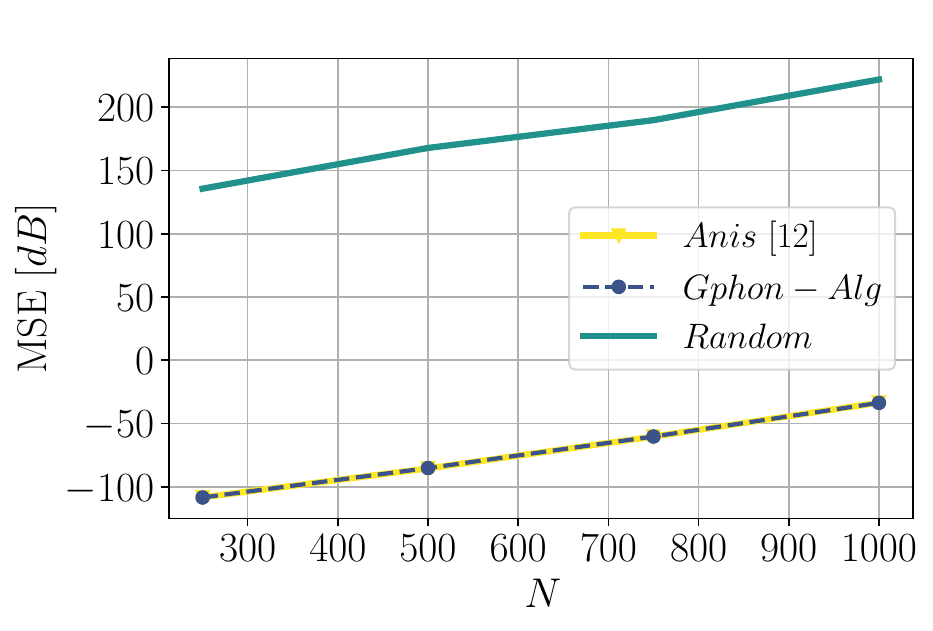} 
\end{subfigure}
%
%
\centering
\begin{subfigure}{.32\linewidth}
        \centering
        \includegraphics[width=1\textwidth]{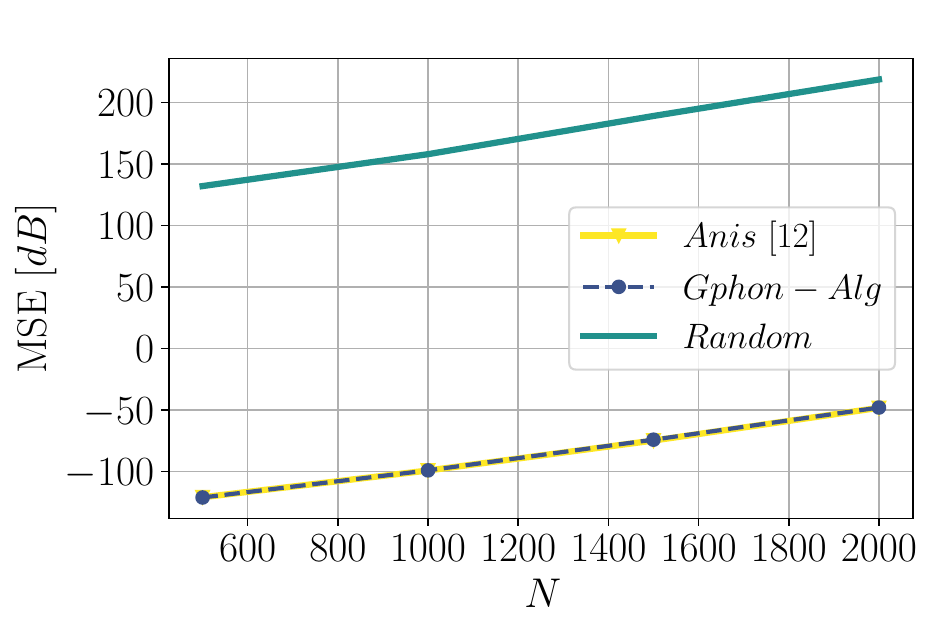} 
        \caption*{$W_{1}(u,v)=(u+v)/2$}
\end{subfigure}
\begin{subfigure}{.32\linewidth}
		\centering
            \includegraphics[width=1\textwidth]{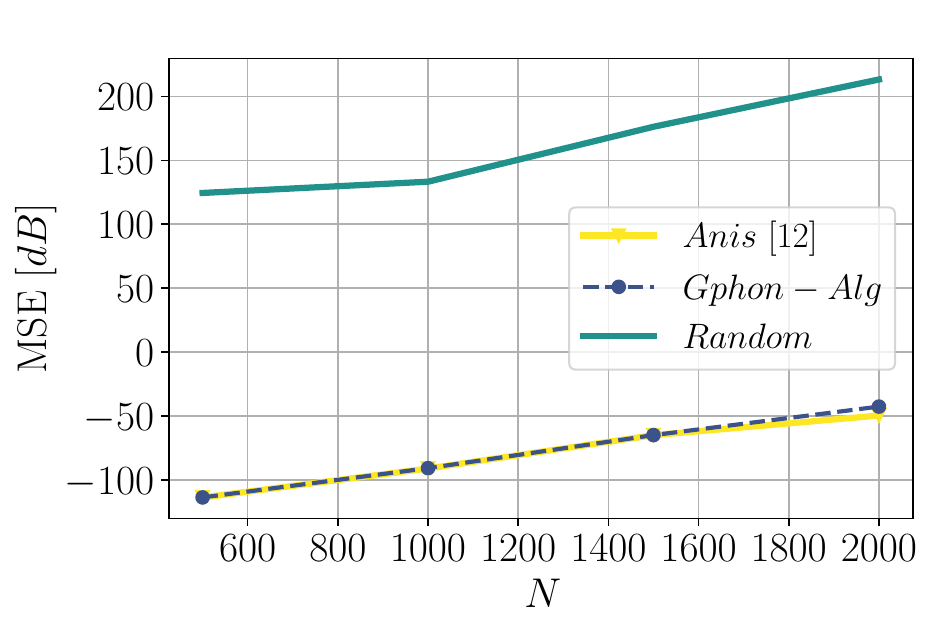} 
            \caption*{$W_{2}(u,v)=\min(x,y)*(1-\max(x,y))$}
\end{subfigure}
\begin{subfigure}{.32\linewidth}
		\centering
            \includegraphics[width=1\textwidth]{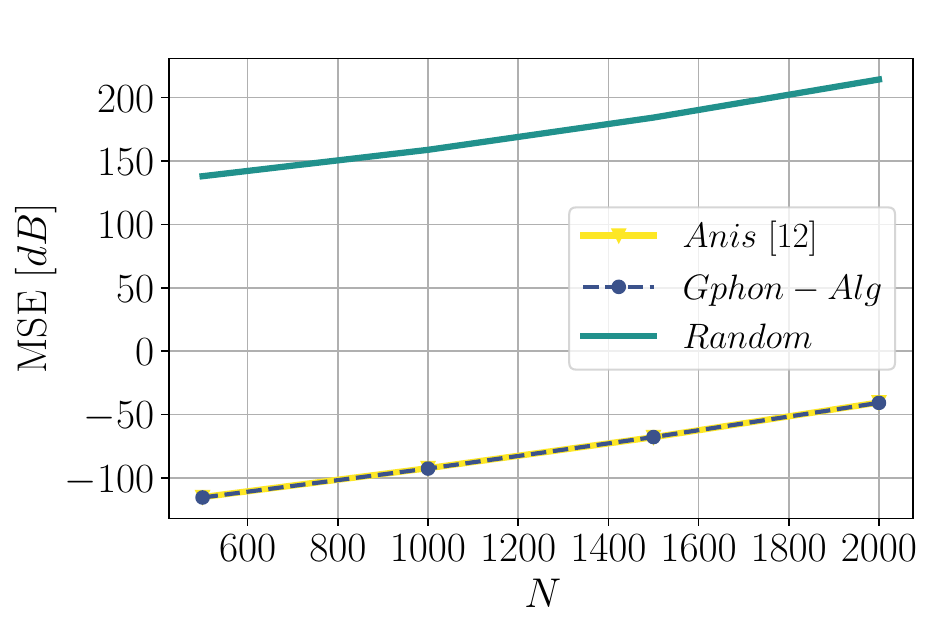} 
            \caption*{$W_{3}(u,v)=1-\max(u,v)$}
\end{subfigure}
    \caption{Reconstruction error of sampled bandlimited signals on graphs derived from a graphon. Each column is associated with experiments performed with a particular graphon, $W_1$ for the left column, $W_2$ for the centered column, and $W_3$ for the column on the right. In the first row, the graphs generated have $250, 500, 750$ and $1000$ nodes. In the second row, the graphs generated have $500, 1000, 1500$ and $2000$ nodes. The graphs are obtained from the graphon using the discretization method (GD1). The bandwidth model considered in this figure is~\textbf{BWM2}. The axis $N$ indicates the number of nodes in the graph. The MSE value depicted is averaged over $50$, which is the number of signals used for each reconstruction. The sampling rate is $5\%$.}
    \label{fig_error_rec_exp_2}
\end{figure*}



\begin{figure*}
%
%
\centering
       \begin{subfigure}{.32\linewidth}
	   \centering
          \includegraphics[width=1\textwidth]{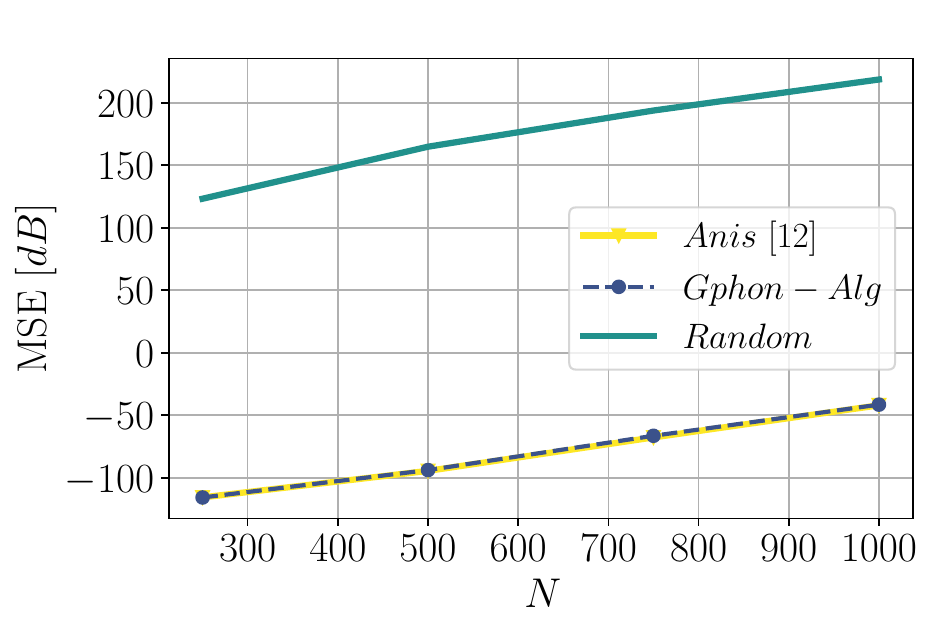} 
       \end{subfigure}
    	\begin{subfigure}{.32\linewidth}
		\centering
            \includegraphics[width=1\textwidth]{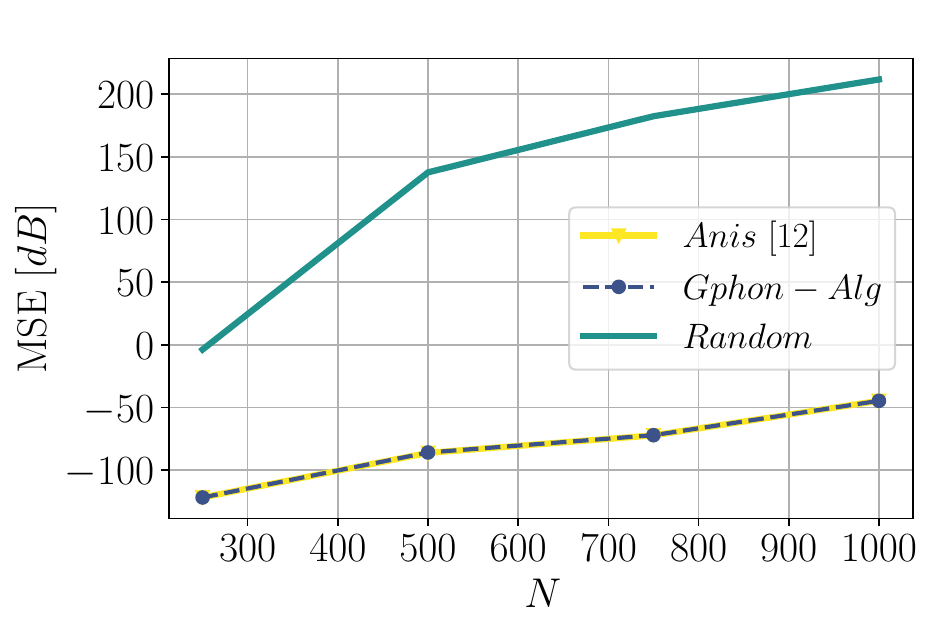} 
\end{subfigure}
    	\begin{subfigure}{.32\linewidth}
		\centering
  \includegraphics[width=1\textwidth]{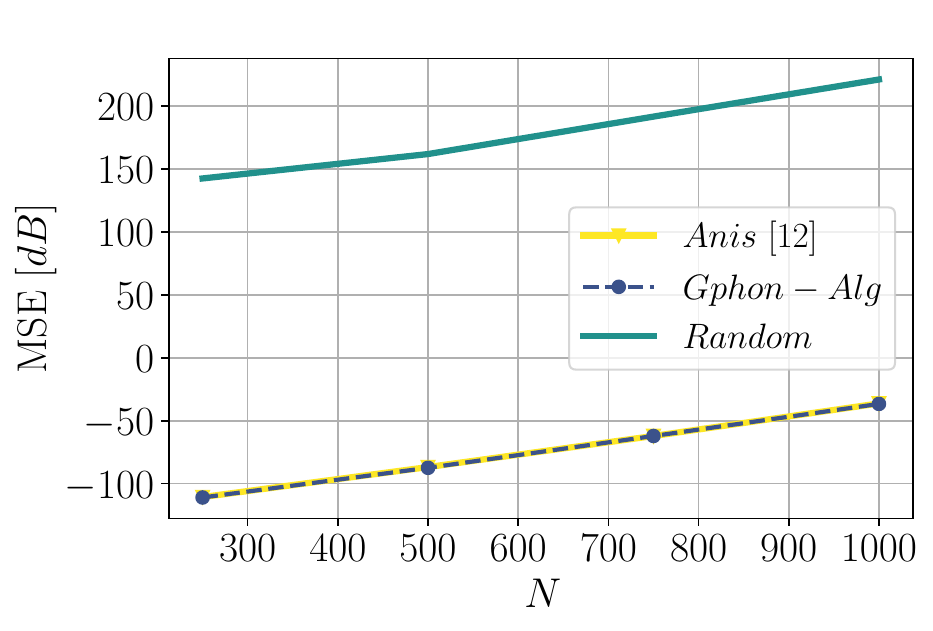} 
\end{subfigure}
%
%
\centering
\begin{subfigure}{.32\linewidth}
        \centering
        \includegraphics[width=1\textwidth]{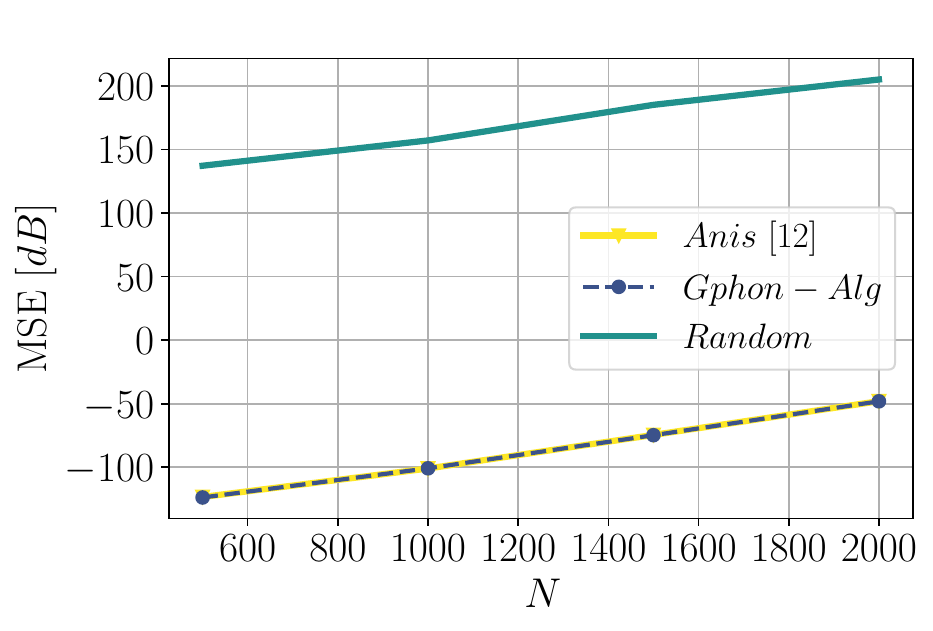} 
        \caption*{$W_{1}(u,v)=(u+v)/2$}
\end{subfigure}
\begin{subfigure}{.32\linewidth}
		\centering
            \includegraphics[width=1\textwidth]{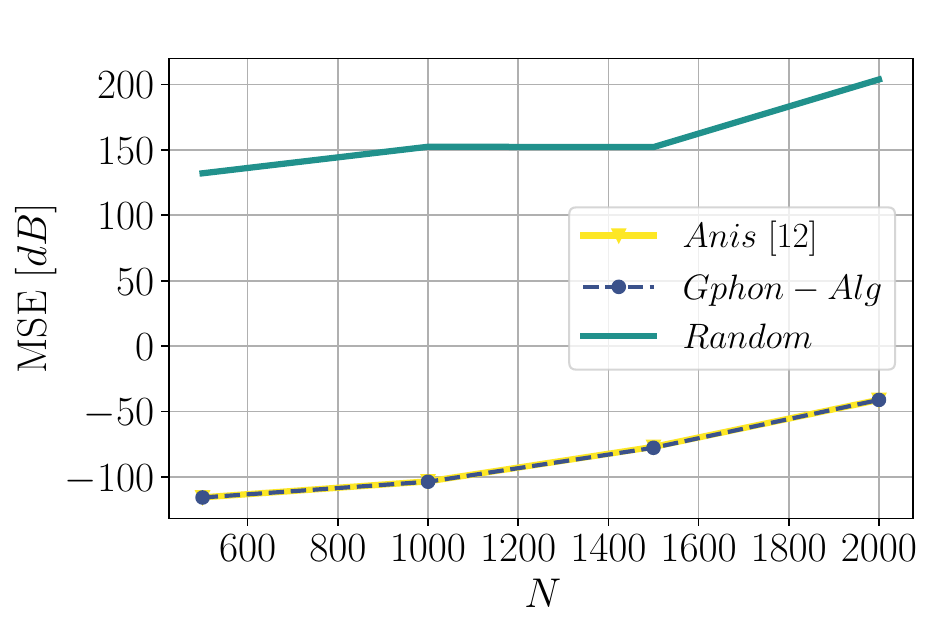} 
            \caption*{$W_{2}(u,v)=\min(x,y)*(1-\max(x,y))$}
\end{subfigure}
\begin{subfigure}{.32\linewidth}
		\centering
            \includegraphics[width=1\textwidth]{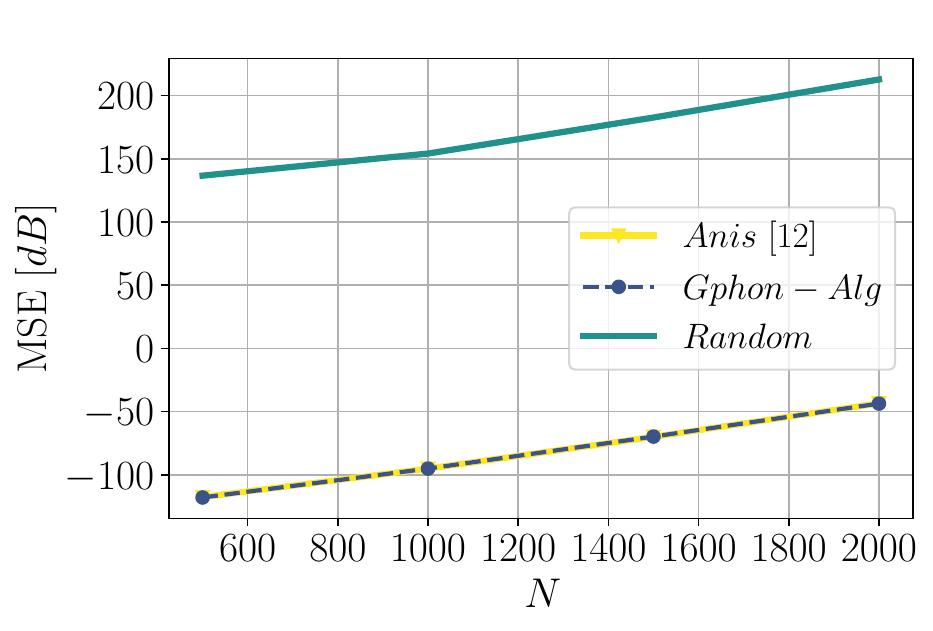} 
            \caption*{$W_{3}(u,v)=1-\max(u,v)$}
\end{subfigure}
    \caption{Reconstruction error of sampled bandlimited signals on graphs derived from a graphon. Each column is associated with experiments performed with a particular graphon, $W_1$ for the left column, $W_2$ for the centered column, and $W_3$ for the column on the right. In the first row, the graphs generated have $250, 500, 750$ and $1000$ nodes. In the second row, the graphs generated have $500, 1000, 1500$ and $2000$ nodes. The graphs are obtained from the graphon using the discretization method (GD1). The bandwidth model considered in this figure is~\textbf{BWM3}. The axis $N$ indicates the number of nodes in the graph. The MSE value depicted is averaged over $50$, which is the number of signals used for each reconstruction. The sampling rate is $5\%$.}
    \label{fig_error_rec_exp_3}
\end{figure*}



\begin{figure*}
%
%
\centering
       \begin{subfigure}{.32\linewidth}
	   \centering
          \includegraphics[width=1\textwidth]{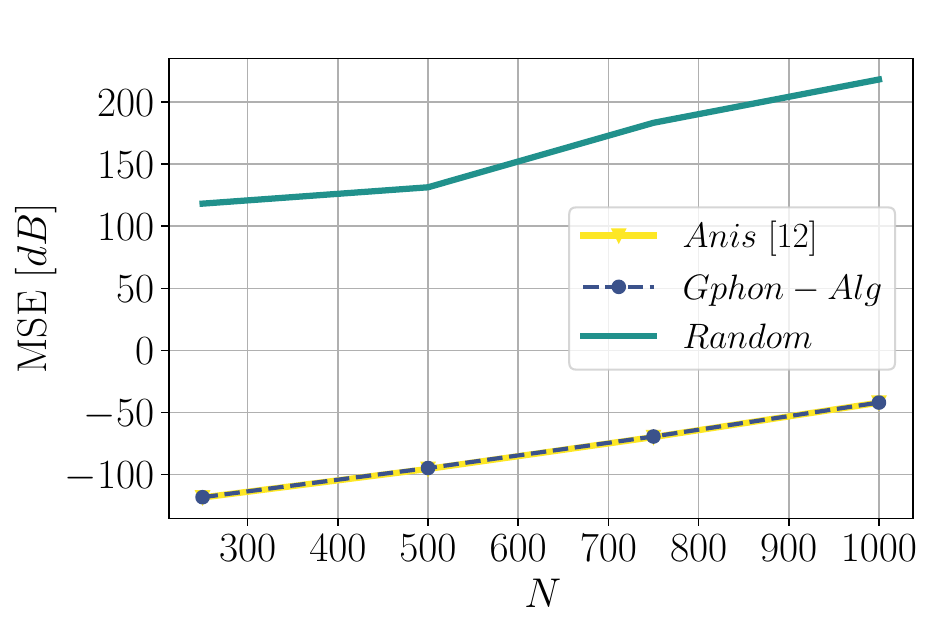} 
       \end{subfigure}
    	\begin{subfigure}{.32\linewidth}
		\centering
            \includegraphics[width=1\textwidth]{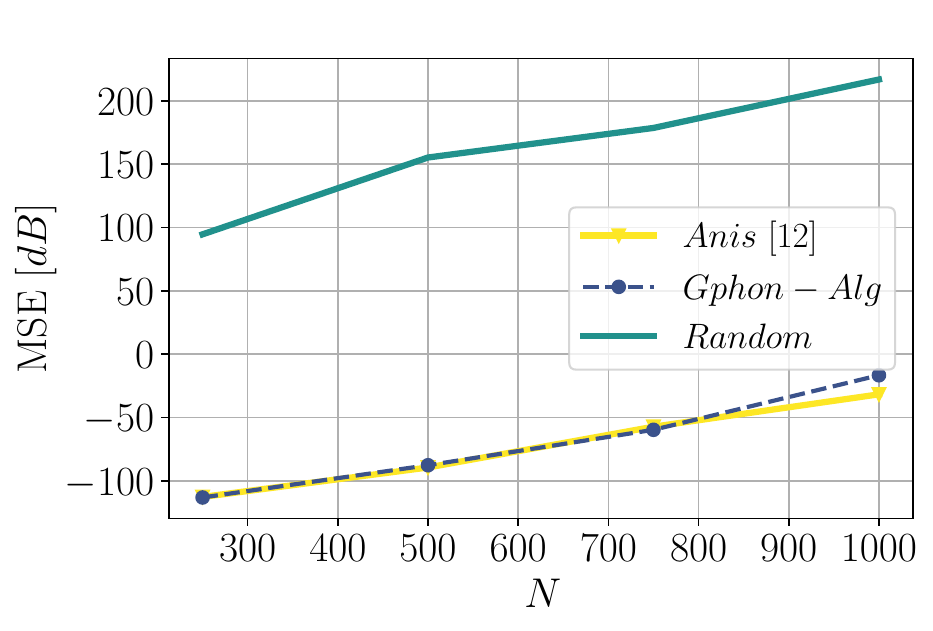} 
\end{subfigure}
    	\begin{subfigure}{.32\linewidth}
		\centering
  \includegraphics[width=1\textwidth]{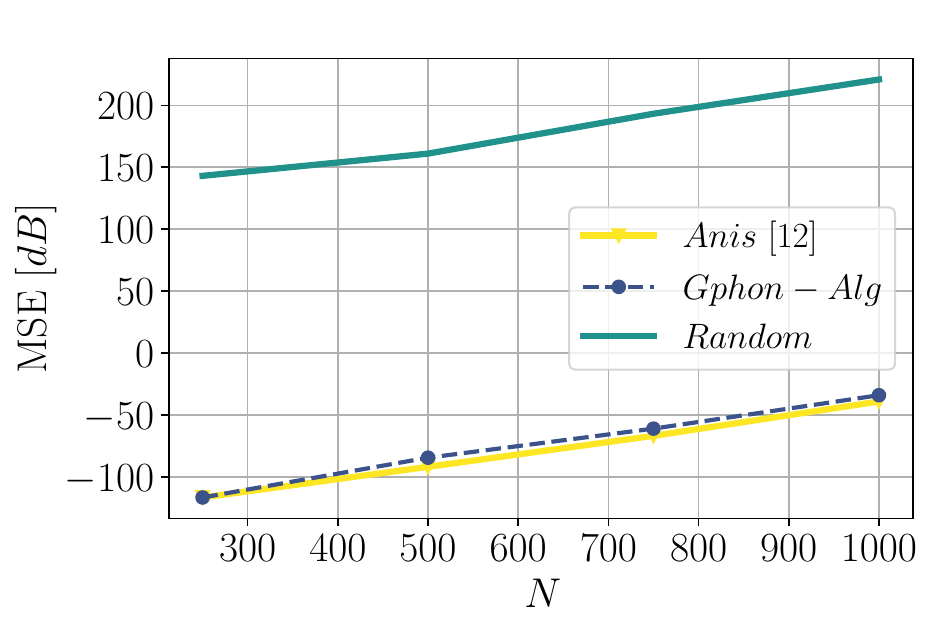} 
\end{subfigure}
%
%
\centering
\begin{subfigure}{.32\linewidth}
        \centering
        \includegraphics[width=1\textwidth]{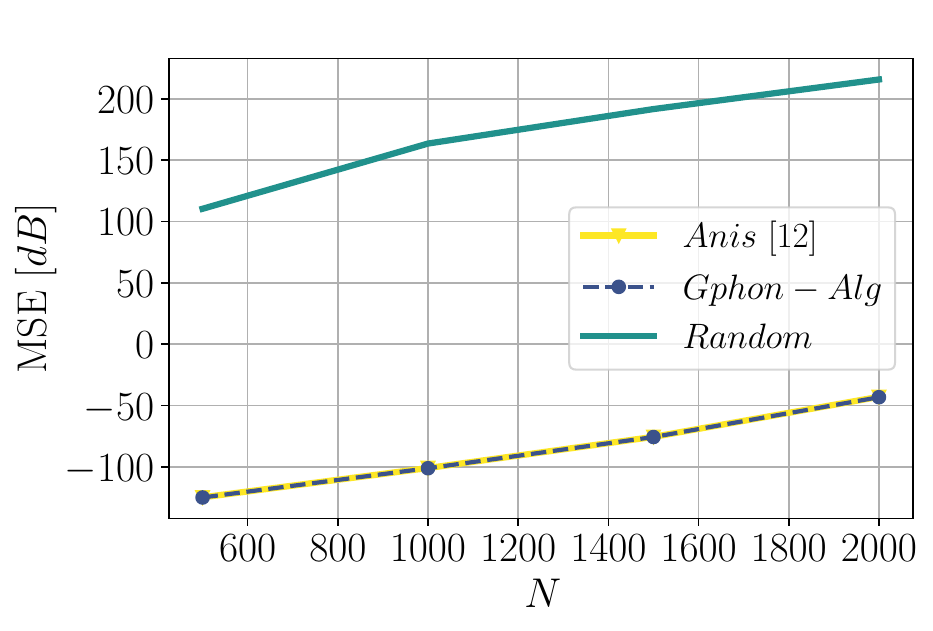} 
        \caption*{$W_{1}(u,v)=(u+v)/2$}
\end{subfigure}
\begin{subfigure}{.32\linewidth}
		\centering
            \includegraphics[width=1\textwidth]{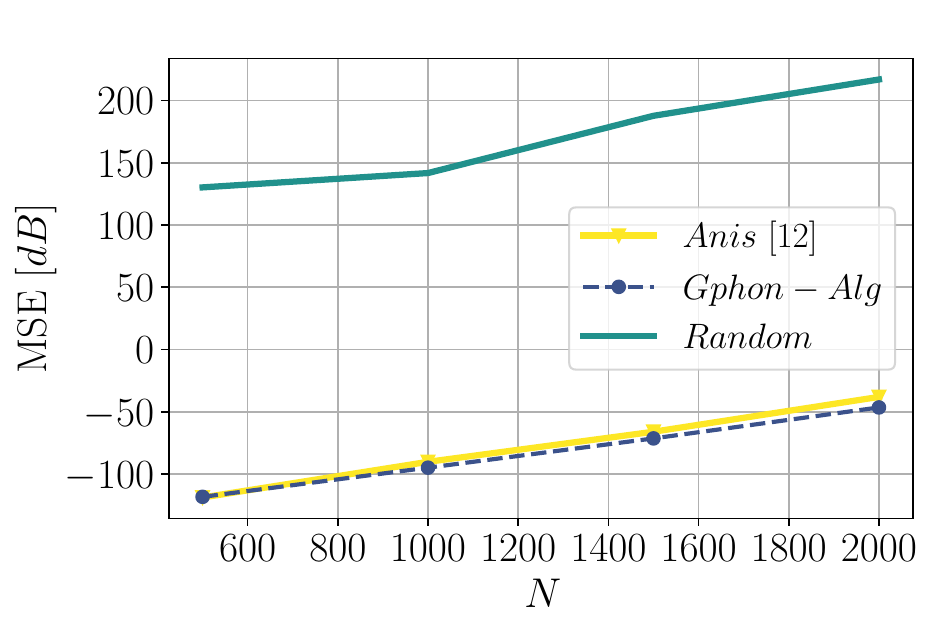} 
            \caption*{$W_{2}(u,v)=\min(x,y)*(1-\max(x,y))$}
\end{subfigure}
\begin{subfigure}{.32\linewidth}
		\centering
            \includegraphics[width=1\textwidth]{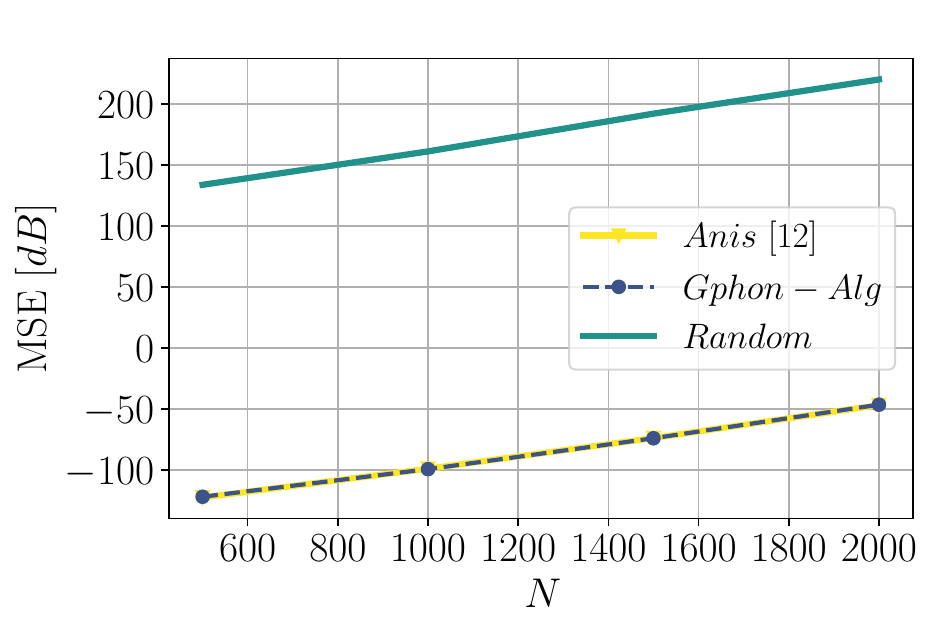} 
            \caption*{$W_{3}(u,v)=1-\max(u,v)$}
\end{subfigure}
    \caption{Reconstruction error of sampled bandlimited signals on graphs derived from a graphon. Each column is associated with experiments performed with a particular graphon, $W_1$ for the left column, $W_2$ for the centered column, and $W_3$ for the column on the right. In the first row, the graphs generated have $250, 500, 750$ and $1000$ nodes. In the second row, the graphs generated have $500, 1000, 1500$ and $2000$ nodes. The graphs are obtained from the graphon using the discretization method (GD1). The bandwidth model considered in this figure is~\textbf{BWM4}. The axis $N$ indicates the number of nodes in the graph. The MSE value depicted is averaged over $50$, which is the number of signals used for each reconstruction. The sampling rate is $5\%$.}
    \label{fig_error_rec_exp_4}
\end{figure*}



\begin{figure*}
%
%
\centering
       \begin{subfigure}{.32\linewidth}
	   \centering
          \includegraphics[width=1\textwidth]{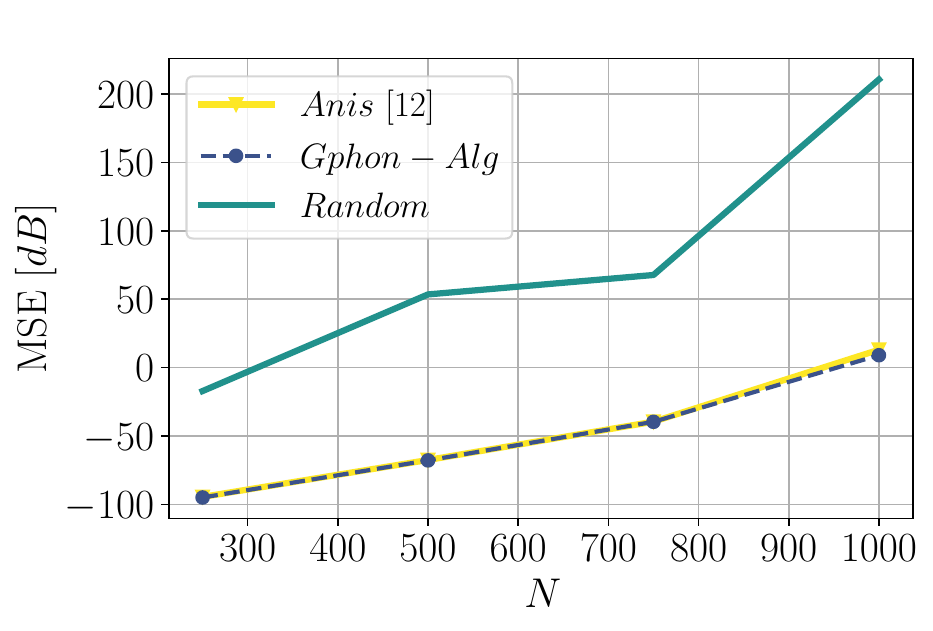} 
       \end{subfigure}
    	\begin{subfigure}{.32\linewidth}
		\centering
            \includegraphics[width=1\textwidth]{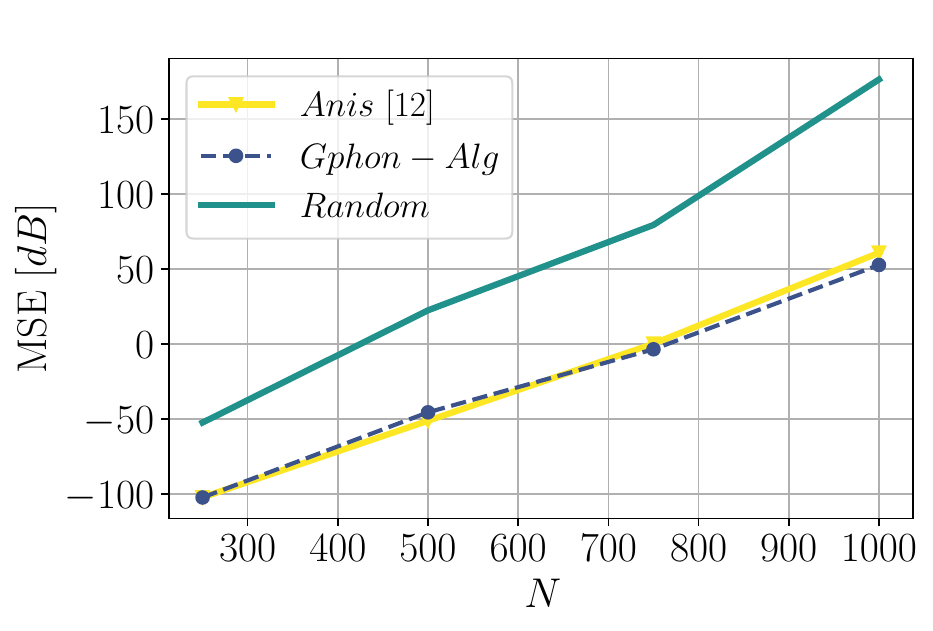} 
\end{subfigure}
    	\begin{subfigure}{.32\linewidth}
		\centering
  \includegraphics[width=1\textwidth]{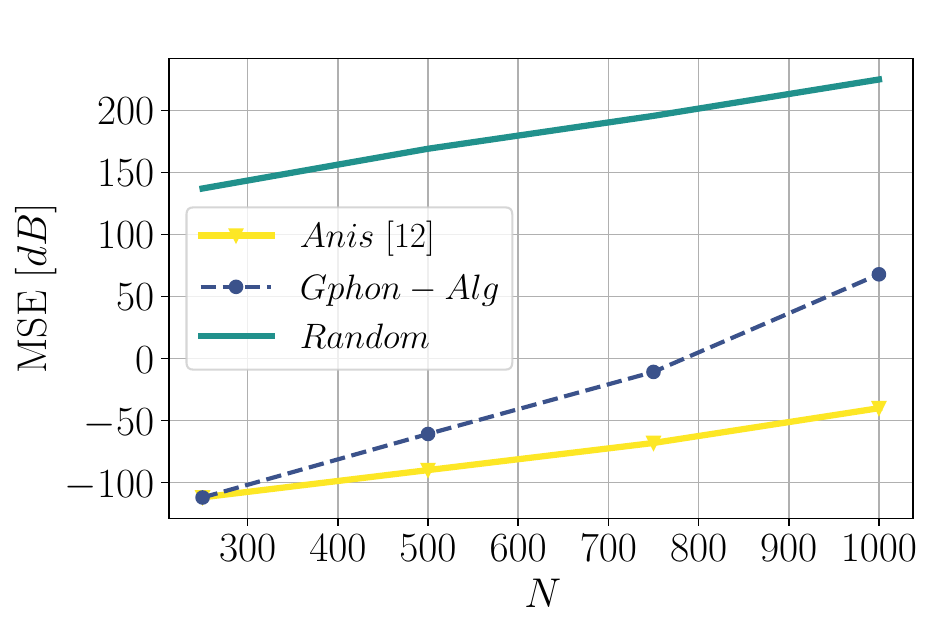} 
\end{subfigure}
%
%
\centering
\begin{subfigure}{.32\linewidth}
        \centering
        \includegraphics[width=1\textwidth]{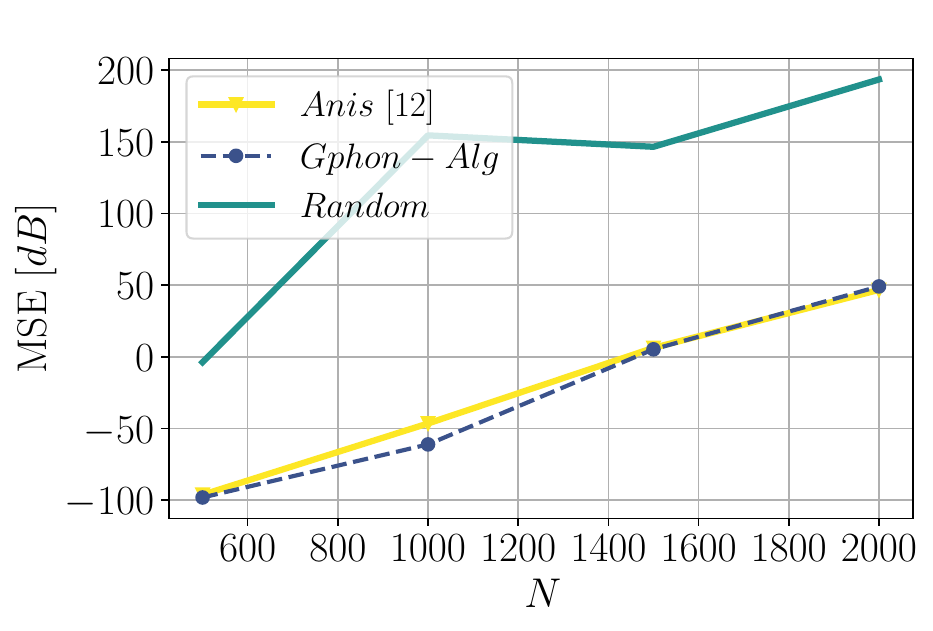} 
        \caption*{$W_{1}(u,v)=\left\vert\sin\left(100uv\right)\right\vert$}
\end{subfigure}
\begin{subfigure}{.32\linewidth}
		\centering
            \includegraphics[width=1\textwidth]{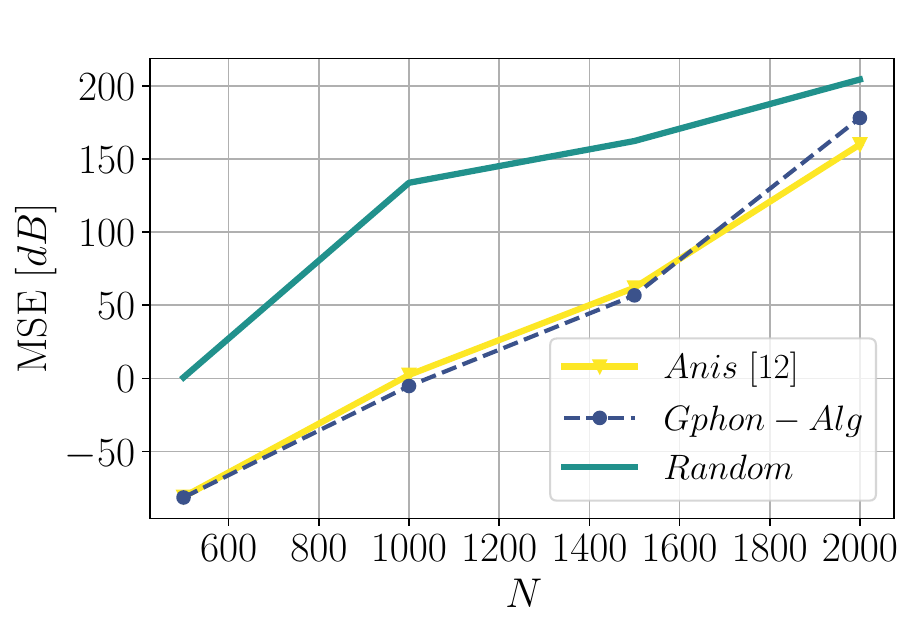} 
            \caption*{$W_{2}(u,v)=\left\vert\sin\left(64uv\right)\right\vert/2+\left\vert\cos\left(64uv\right)\right\vert/2$}
\end{subfigure}
\begin{subfigure}{.32\linewidth}
		\centering
            \includegraphics[width=1\textwidth]{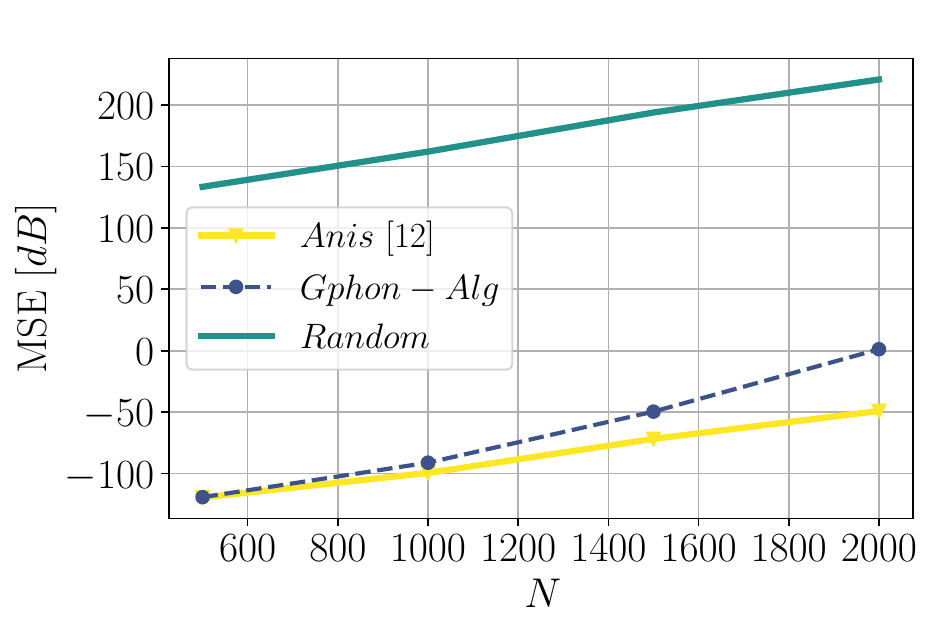} 
            \caption*{$W_{3}(u,v)=\left\vert\sin\left(10uv\right)\right\vert/2+\left\vert\cos\left(10uv\right)\right\vert/2$}
\end{subfigure}
    \caption{Reconstruction error of sampled bandlimited signals on graphs derived from a graphon. Each column is associated with experiments performed with a particular graphon, $W_1$ for the left column, $W_2$ for the centered column, and $W_3$ for the column on the right. In the first row, the graphs generated have $250, 500, 750$ and $1000$ nodes. In the second row, the graphs generated have $500, 1000, 1500$ and $2000$ nodes. The graphs are obtained from the graphon using the discretization method (GD1). The bandwidth model considered in this figure is~\textbf{BWM2}. The axis $N$ indicates the number of nodes in the graph. The MSE value depicted is averaged over $50$, which is the number of signals used for each reconstruction. The sampling rate is $5\%$.}
    \label{fig_error_rec_exp_5}
\end{figure*}



\begin{figure*}
%
%
\centering
       \begin{subfigure}{.32\linewidth}
	   \centering
          \includegraphics[width=1\textwidth]{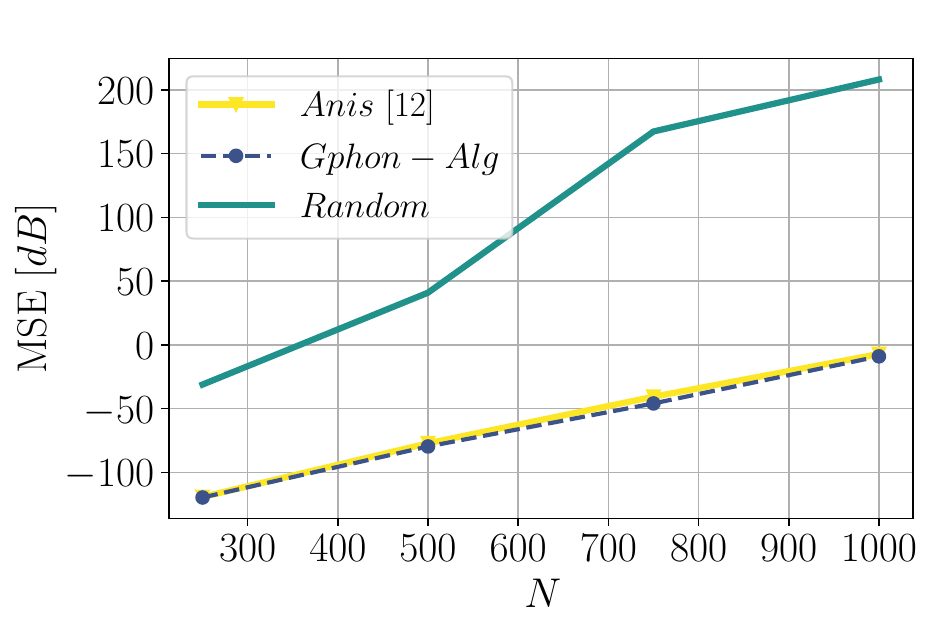} 
       \end{subfigure}
    	\begin{subfigure}{.32\linewidth}
		\centering
            \includegraphics[width=1\textwidth]{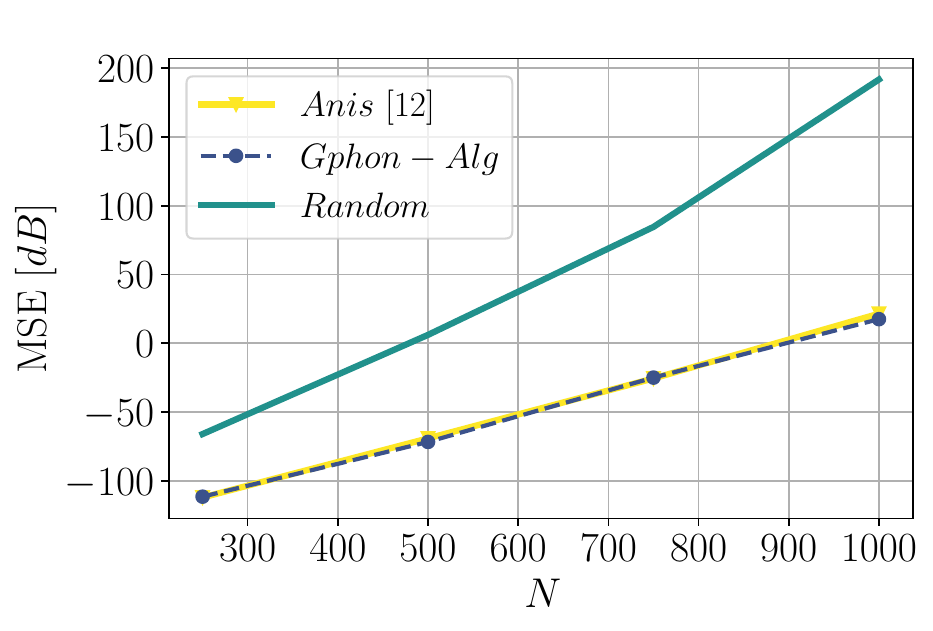} 
\end{subfigure}
    	\begin{subfigure}{.32\linewidth}
		\centering
  \includegraphics[width=1\textwidth]{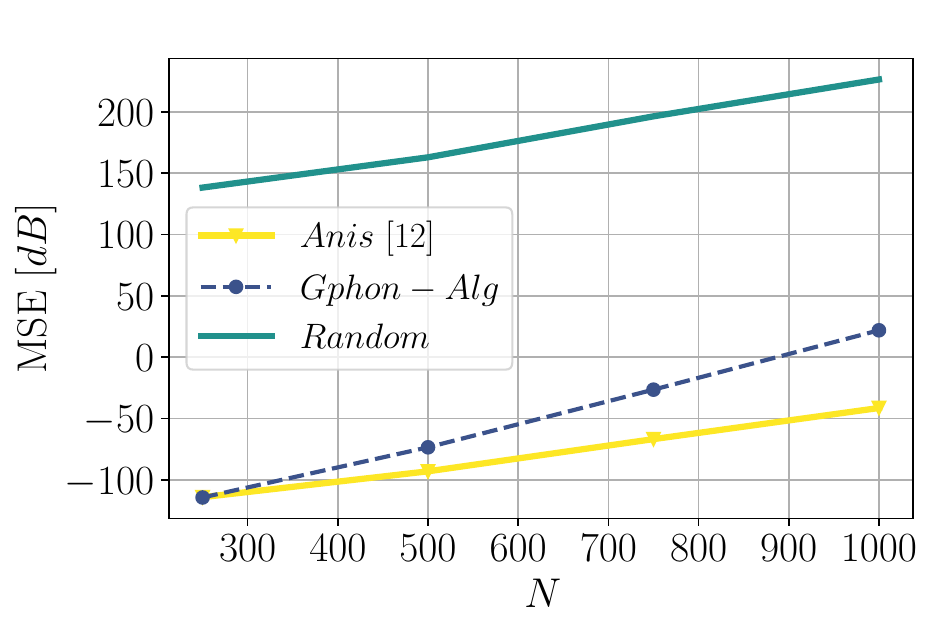} 
\end{subfigure}
%
%
\centering
\begin{subfigure}{.32\linewidth}
        \centering
        \includegraphics[width=1\textwidth]{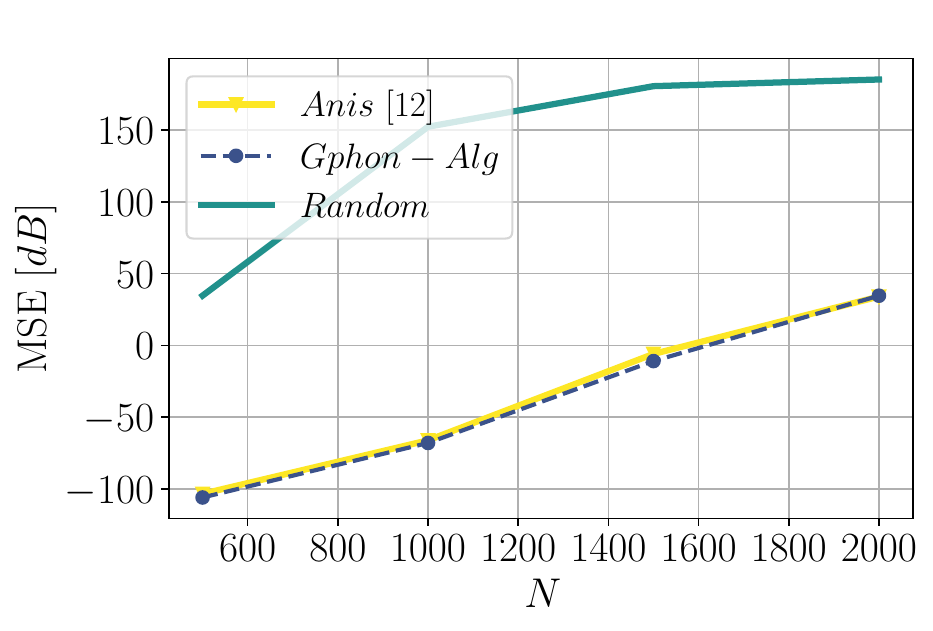} 
        \caption*{$W_{1}(u,v)=\left\vert\sin\left(100uv\right)\right\vert$}
\end{subfigure}
\begin{subfigure}{.32\linewidth}
		\centering
            \includegraphics[width=1\textwidth]{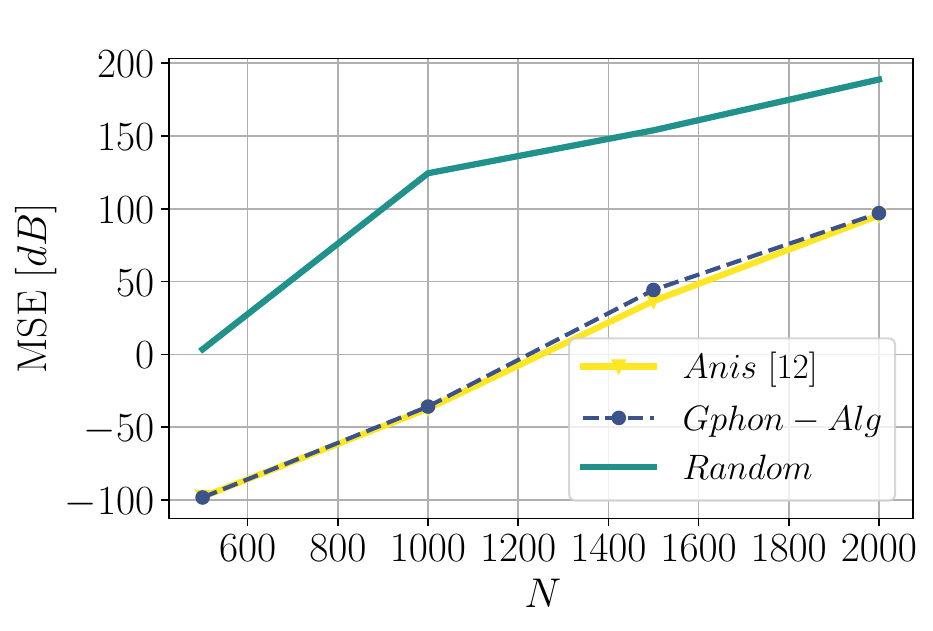} 
            \caption*{$W_{2}(u,v)=\left\vert\sin\left(64uv\right)\right\vert/2+\left\vert\cos\left(64uv\right)\right\vert/2$}
\end{subfigure}
\begin{subfigure}{.32\linewidth}
		\centering
            \includegraphics[width=1\textwidth]{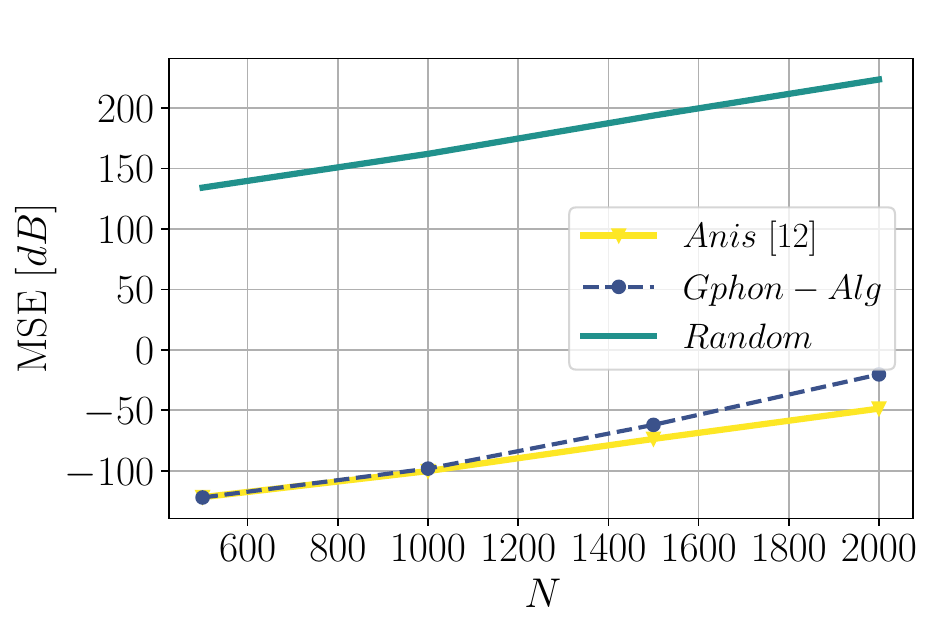} 
            \caption*{$W_{3}(u,v)=\left\vert\sin\left(10uv\right)\right\vert/2+\left\vert\cos\left(10uv\right)\right\vert/2$}
\end{subfigure}
    \caption{Reconstruction error of sampled bandlimited signals on graphs derived from a graphon. Each column is associated with experiments performed with a particular graphon, $W_1$ for the left column, $W_2$ for the centered column, and $W_3$ for the column on the right. In the first row, the graphs generated have $250, 500, 750$ and $1000$ nodes. In the second row, the graphs generated have $500, 1000, 1500$ and $2000$ nodes. The graphs are obtained from the graphon using the discretization method (GD1). The bandwidth model considered in this figure is~\textbf{BWM3}. The axis $N$ indicates the number of nodes in the graph. The MSE value depicted is averaged over $50$, which is the number of signals used for each reconstruction. The sampling rate is $5\%$.}
    \label{fig_error_rec_exp_6}
\end{figure*}



\begin{figure*}
%
%
\centering
       \begin{subfigure}{.32\linewidth}
	   \centering
          \includegraphics[width=1\textwidth]{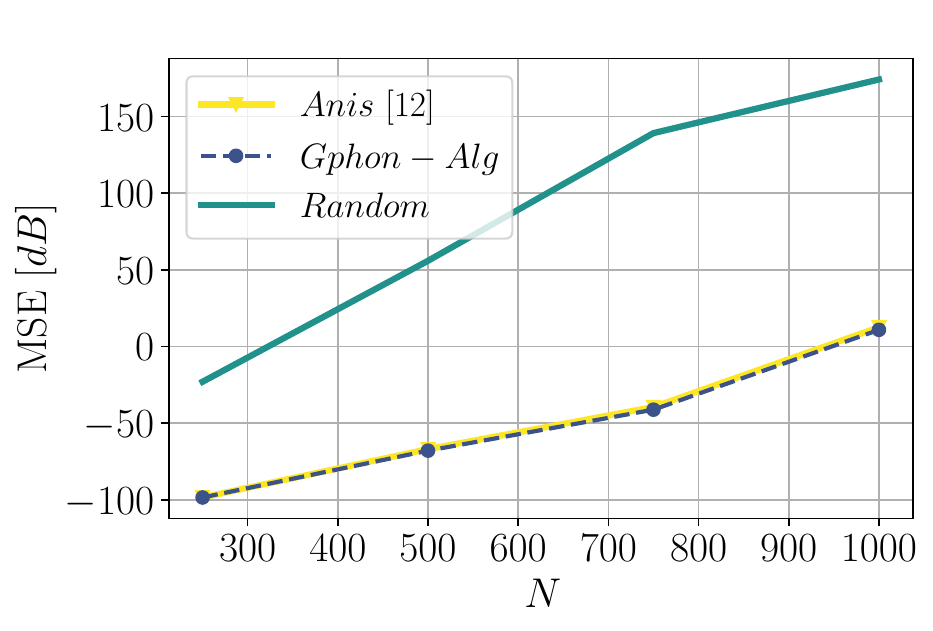} 
       \end{subfigure}
    	\begin{subfigure}{.32\linewidth}
		\centering
            \includegraphics[width=1\textwidth]{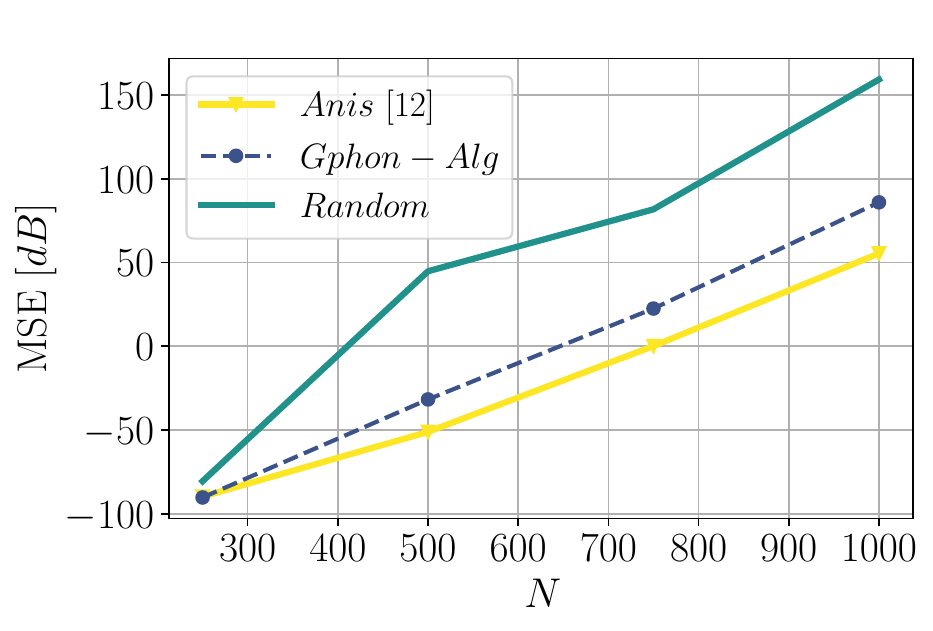} 
\end{subfigure}
    	\begin{subfigure}{.32\linewidth}
		\centering
  \includegraphics[width=1\textwidth]{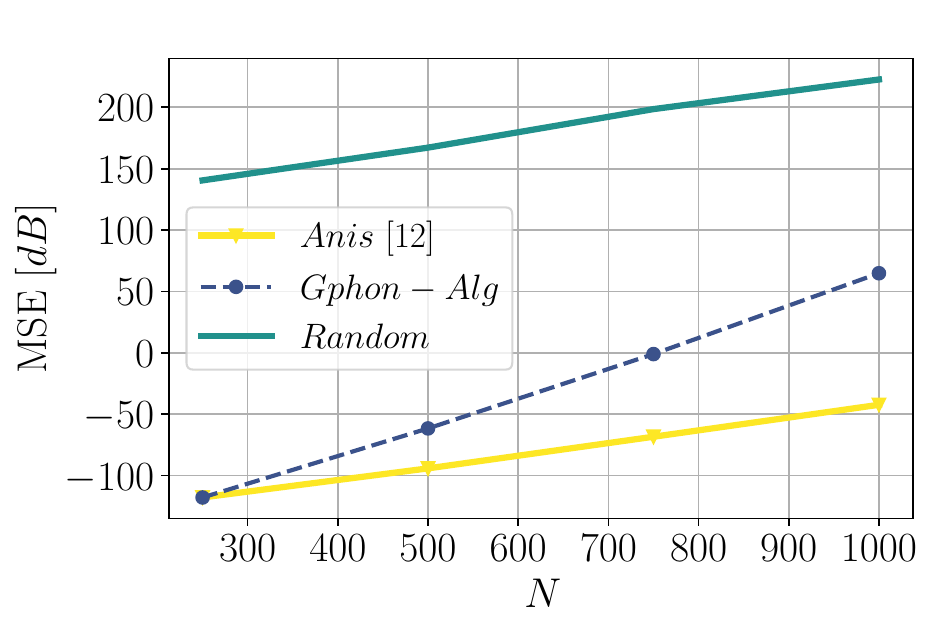} 
\end{subfigure}
%
%
\centering
\begin{subfigure}{.32\linewidth}
        \centering
        \includegraphics[width=1\textwidth]{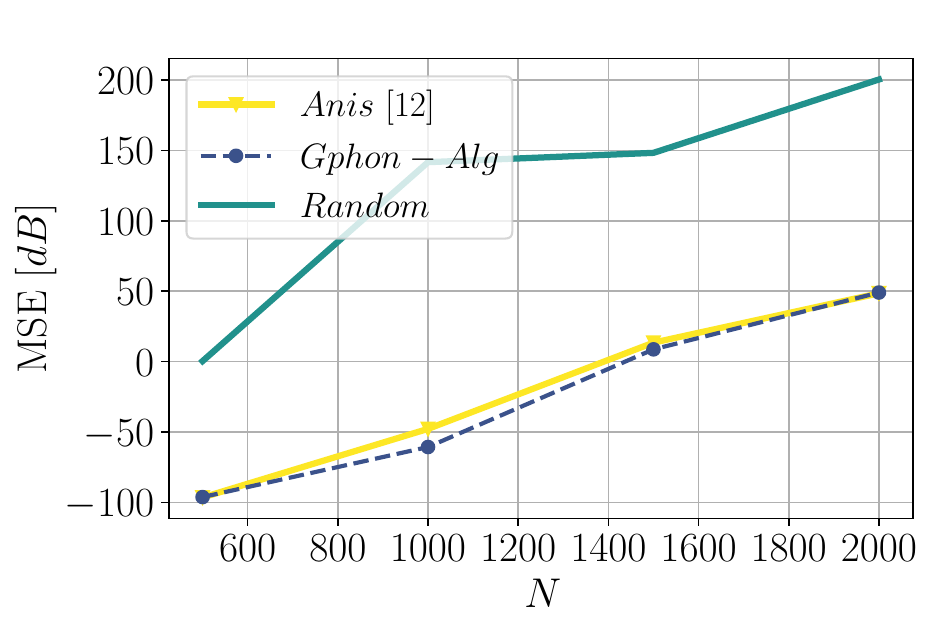} 
        \caption*{$W_{1}(u,v)=\left\vert\sin\left(100uv\right)\right\vert$}
\end{subfigure}
\begin{subfigure}{.32\linewidth}
		\centering
            \includegraphics[width=1\textwidth]{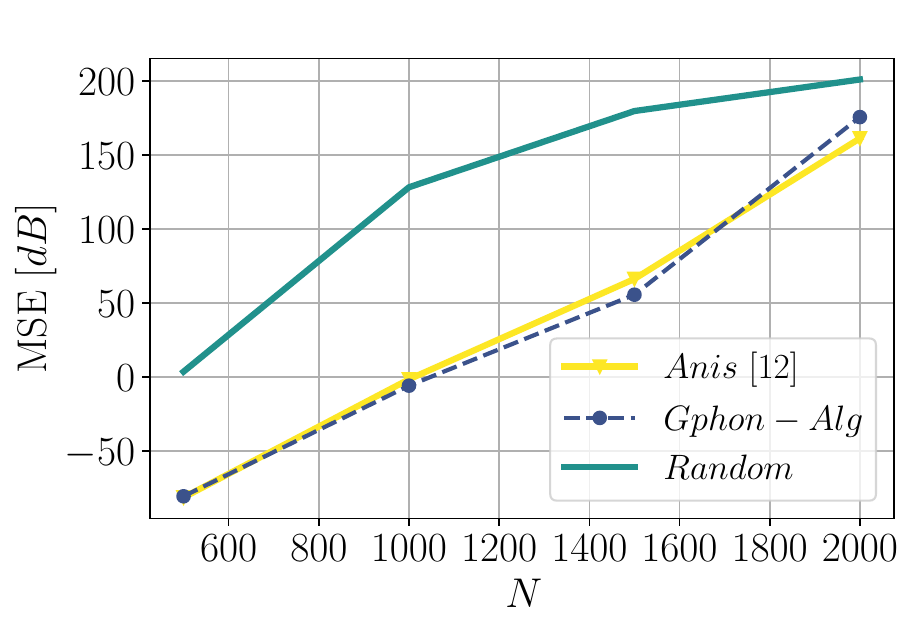} 
            \caption*{$W_{2}(u,v)=\left\vert\sin\left(64uv\right)\right\vert/2+\left\vert\cos\left(64uv\right)\right\vert/2$}
\end{subfigure}
\begin{subfigure}{.32\linewidth}
		\centering
            \includegraphics[width=1\textwidth]{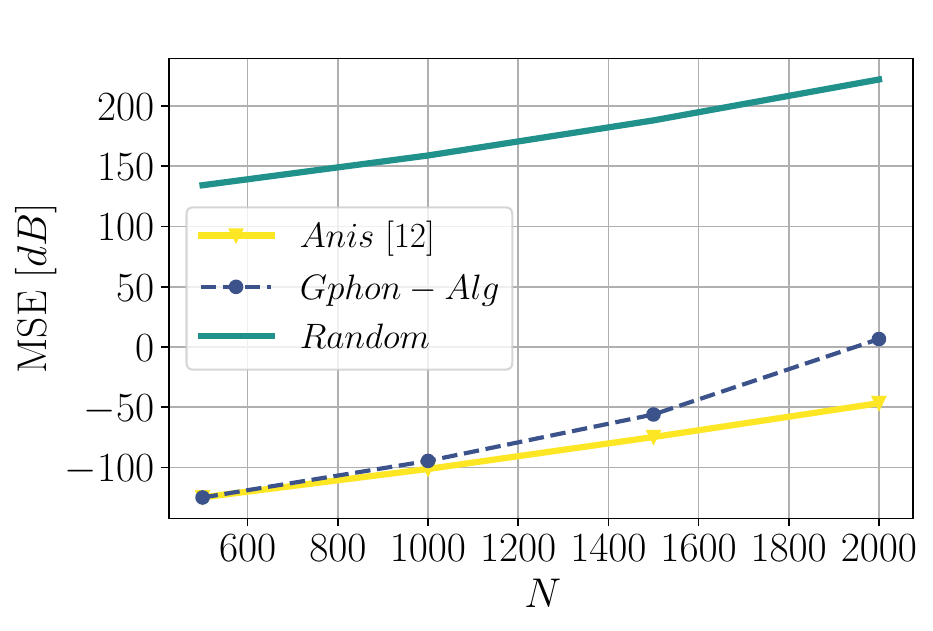} 
            \caption*{$W_{3}(u,v)=\left\vert\sin\left(10uv\right)\right\vert/2+\left\vert\cos\left(10uv\right)\right\vert/2$}
\end{subfigure}
    \caption{Reconstruction error of sampled bandlimited signals on graphs derived from a graphon. Each column is associated with experiments performed with a particular graphon, $W_1$ for the left column, $W_2$ for the centered column, and $W_3$ for the column on the right. In the first row, the graphs generated have $250, 500, 750$ and $1000$ nodes. In the second row, the graphs generated have $500, 1000, 1500$ and $2000$ nodes. The graphs are obtained from the graphon using the discretization method (GD1). The bandwidth model considered in this figure is~\textbf{BWM4}. The axis $N$ indicates the number of nodes in the graph. The MSE value depicted is averaged over $50$, which is the number of signals used for each reconstruction. The sampling rate is $5\%$.}
    \label{fig_error_rec_exp_7}
\end{figure*}


In this section, we provide a set of numerical experiments that will allow us to validate the insights and results presented in Section~\ref{sec_samp_on_graphons}. We aim to numerically verify how the properties of sampling sets are approximately inherited by families of graphs whose structural properties are tied to a graphon. This is the underlying principle captured in Algorithm~\ref{alg_sampt_method}. We start by presenting the general steps of the experiments and later on, we present the specific details of the experiment.

%
%
\begin{itemize}
\item First, we consider several graphon models, $\{ W_{i}(u,v) \}_{i}$, and we generate a collection of graphs, $\left\lbrace G_{j}^{W_{i}} \right\rbrace_{j}$, for each graphon, where
\begin{equation}
\left\vert 
        V\left(
             G_{j}^{W_{i}}
          \right)
\right\vert
         <
\left\vert 
        V\left(
             G_{j+1}^{W_{i}}
          \right)
\right\vert  
.
\end{equation}
\item On each graph, $G_{j}^{W_{i}}$, we generate random bandlimited signals with a fixed bandwidth. Given such bandwidth and a sampling rate, we determine sampling sets utilizing three methods: random uniform sampling, Anis et al.~\cite{ortega_proxies}, and Algorithm~\ref{alg_sampt_method}. 
\item Then, we perform sampling and average the reconstruction error over the number of bandlimited signals used. To determine the initial sampling sets in the Algorithm~\ref{alg_sampt_method}, we start computing the optimal sampling set on the smallest graph obtained from each graphon, $G_{1}^{W_{i}}$, using Anis~\cite{ortega_proxies}. Then, we use the steps in Algorithm~\ref{alg_sampt_method} to determine the sampling sets for larger graphs, i.e. $G_{2}^{W_{i}}$, $G_{3}^{W_{i}}$ and so on.
\end{itemize}

\medskip

The specific details of the graphons, graphs, signals and methods used in the experiments are the following.

\medskip


\noindent\textbf{Graphon models}: $W(u,v)=u+v$, $W(u,v)=(u^{2}+v^{2})/2$, $W(u,v)=1-\max(x,y)$, $W(u,v)=\min(x,y)*(1-\max(x,y))$, $W(u,v)=\left\vert\sin\left(100uv\right)\right\vert$, $W(u,v)=\left\vert\sin\left(64uv\right)\right\vert/2+\left\vert\cos\left(64uv\right)\right\vert/2$, and $W(u,v)=\left\vert\sin\left(10uv\right)\right\vert/2+\left\vert\cos\left(10uv\right)\right\vert/2$.

\medskip
\noindent\textbf{Discretization method and graph generation}: The graphs, $G_{j}^{W_i}$, are obtained from each $W_{i}(u,v)$ using the equipartition method (GD1) -- Section~\ref{sec_gphon_sp}. Two scenarios are considered for each graphon. In the first scenario, we generate graphs with $250, 500, 750$ and $1000$ nodes. In the second scenario we generate graphs with $500, 1000, 1500$ and $2000$ nodes.

\medskip
\noindent\textbf{Sampling rate and bandwidth}: The sampling rate, $m$, is chosen to be $5\%$ of the total number of nodes in the graph: $m=\left\lfloor 0.05\left\vert V\left( G_{j}^{W_i}\right) \right\vert \right\rceil$. The bandwidth of the signals, $\omega$, is selected such that signals are low pass and also according to the following models:

\begin{itemize}
\item \textbf{BWM1}. The bandwidth of the signals is equal to the sampling rate:
       \begin{equation}
       	k_\omega = m,~\quad m=\left\lfloor 0.05\left\vert V\left( G_{j}^{W_i}\right) \right\vert \right\rceil.
       \end{equation}      
\item \textbf{BWM2}. The bandwidth is $90\%$ of the sampling rate:
       \begin{equation}
       	k_\omega = \left\lfloor 0.9m \right\rceil,~\quad m=\left\lfloor 0.05\left\vert V\left( G_{j}^{W_i}\right) \right\vert \right\rceil.
       \end{equation}
\item \textbf{BWM3}. The bandwidth is $85\%$ of the sampling rate:
       \begin{equation}
       	k_\omega = \left\lfloor 0.85m \right\rceil,~\quad m=\left\lfloor 0.05\left\vert V\left( G_{j}^{W_i}\right) \right\vert \right\rceil.
       \end{equation}
\item \textbf{BWM4}. A random signal is filtered in frequency with the filter
       \begin{equation}
           h(k)
                 =
                  \begin{cases}
			          1, &  k\leq k_{\omega} \\
                        e^{-4(k - k_{\omega})}, &  k>k_{\omega}
		       \end{cases}
               ,
       \end{equation}
       where 
       \begin{equation}
       	k_\omega = \left\lfloor 0.9m \right\rceil,~\quad m=\left\lfloor 0.05\left\vert V\left( G_{j}^{W_i}\right) \right\vert \right\rceil.
       \end{equation}
\end{itemize}

\noindent The Fourier coefficients of the signals in $\mathcal{PW}_{\omega}\left( G_{j}^{W_i} \right)$ are generated using the
Gaussian distribution $\ccalN (1,0.52)$. For each bandwidth model we generate $50$ signals to run the experiment.

\medskip
\noindent\textbf{Sampling methods}: Random uniform sampling, Anis et al.~\cite{ortega_proxies}, and Algorithm~\ref{alg_sampt_method}. We add noise to the samples considering an $SNR=20dB$. Note that in Algorithm~\ref{alg_sampt_method} we use the sampling sets determined with~\cite{ortega_proxies} in the smallest graph generated from the graphon and then approximate the sampling sets in the larger graphs.

\medskip
\noindent\textbf{Reconstruction method}: To reconstruct the sampled signals we use the reconstruction method used in~\cite{ortega_proxies,tsitsverobarbarossa,7581102,alejopm_phdthesis,alejopm_BN_j} and that we describe as follows. Let $\left(G_{j}^{W_i}, \bbx_{rec} \right)$ be the reconstructed signal from the samples in $\ccalS\subset V\left(G_{j}^{W_i} \right)$  obtained from the graph signal $\left(G_{j}^{W_i}, \bbx \right)$. Let $\bbU_{k_{\omega}}$ the matrix whose columns are the eigenvectors that span $\mathcal{PW}_{\omega}\left( G_{j}^{W_i} \right)$. Then, we have
\begin{equation}
\bbx_{rec}
      =
        \argmin_{\bbz \in span(\mathbf{U}_{k_\omega})}
               \left \Vert 
                       \mathbf{M}\bbz
                        -
                        \bbx(\ccalS)
               \right\Vert^{2}_{2}
      =
        \mathbf{U}_{k_\omega}\left( 
                                     \mathbf{M}\mathbf{U}_{k_\omega}
                                \right)^{\dagger}\bbx(\ccalS) 
                                ,
\label{eq_basicrec}
\end{equation}
where $\bbx(\ccalS)=\mathbf{M}\bbx$ and $\mathbf{M}$ is the matrix with entries are given by $\mathbf{M} = [\boldsymbol{\delta}_{s_{1}},\ldots,\boldsymbol{\delta}_{s_{m}}]^{\mathsf{T}}$,
$\boldsymbol{\delta}_{s_i}$ is the $N-$ dimensional Kronecker column vector centered at $s_{i}\in\ccalS$ and $(\cdot)^{\dagger}$ indicates the pseudo inverse matrix.

\medskip
\noindent\textbf{Measuring the error}: To evaluate the error we use the mean square error (MSE). We generate $50$ signals for each space $\mathcal{PW}_{\omega}$ to perform the experiments and average the MSE.

\medskip
\noindent\textbf{Diversity of the scenarios considered in the experiments}: It is important to emphasize that the models~\textbf{BWM1}-\textbf{BWM4} represent standard scenarios considered in~\cite{ortega_proxies,puysampling,chensamplingongraphs,tremblayAB17,tsitsverobarbarossa,JAYAWANT2022108436,alejopm_phdthesis,alejopm_cographs_c,alejopm_BN_j}
for the evaluation of the quality of a sampling set. In all cases, there is exact or approximate knowledge of the bandwidth of the signals that are being sampled and, consequently, knowledge of the sampling rate to be used. We emphasize that the experiments allow us to properly compare the quality of sampling sets generated generated by different methods. This is central to our goal of showing that the optimality of good sampling sets in small graphs is transferable to larger graphs with similar structural properties. Additionally, please notice that the graphons used include typical graphons considered in the literature and more exotic graphons such as those including cosine functions which exhibit high variability in $[0,1]^{2}$.




\bigskip\medskip

It is important to remark that the sampling method in~\cite{ortega_proxies} is one among many that can be used to find the sampling sets~\cite{puysampling,chensamplingongraphs,tremblayAB17,tsitsverobarbarossa,JAYAWANT2022108436,alejopm_phdthesis,alejopm_cographs_c,alejopm_BN_j}. The point we want to emphasize in our experiments is not about the benefits of any particular sampling technique over the other, but instead on how the sampling sets obtained by any given close to optimal method, are inherited by families of graphs whose structural properties are captured by a graphon representation.

The results of our experiments are depicted in Fig.~\ref{fig_error_rec_exp_1}, Fig.~\ref{fig_error_rec_exp_2}, Fig.~\ref{fig_error_rec_exp_3}, Fig.~\ref{fig_error_rec_exp_4}, Fig.~\ref{fig_error_rec_exp_5}, Fig.~\ref{fig_error_rec_exp_6} and Fig.~\ref{fig_error_rec_exp_7}. In these figures, each column contains the results of the experiments performed with a graphon, $W_1$ for the left column, $W_2$ for the centered column, and $W_3$ for the column on the right. In the first row, we have the results for the graphs with $250, 500, 750$ and $1000$ nodes. In the second row, we have the results for the graphs with $500, 1000, 1500$ and $2000$ nodes. As can be observed, the goodness of the sampling sets obtained in the small graphs with~\cite{ortega_proxies} are well preserved when re-used under the Algorithm~\ref{alg_sampt_method}. This behavior is consistent for all the graphons and bandwidth models considered. Notice also that those graphons with a high variability in $[0,1]^{2}$ -- like the ones with cosine functions -- naturally induce more signaificant differences between the graphs generated (and their shift operators) which affects the preservation of the goodness of the sampling sets across the different graphs.

Notice that for \textbf{BWM1} and $W_{3}(u,v)$ in Fig.~\ref{fig_error_rec_exp_1}, one can see a small difference between \textit{Gphon-Alg} and Anis~\cite{ortega_proxies}. We argue that this phenomena occurs as a consequence of the conditioning of the matrix in the reconstruction given by~\eqref{eq_basicrec}. Anis method aims to minimize the effects of the noise in the reconstruction, providing a sampling set inside a class of multiple optimal sampling sets, that are naturally associated to different condition numbers in the reconstruction matrix. The algorithm we propose aims to preserve the goodness of the optimal sampling sets determined by Anis in the small graph, when reused and adapted in larger graphs. Therefore the difference observed in Fig.~\ref{fig_error_rec_exp_1} occurs because although both sampling sets are inside the optimal class, they lead to reconstruction matrices different condition numbers.


\begin{remark}\normalfont
Due to the limited space, we include extra experiments and results in the supplementary material. Among the extra experiments included are the results for the scenario \textbf{BWM1} for the sinusoidal type graphons, and the evaluation of the sampling methods when noise is also added to the graphs generated from the underlying graphon.
\end{remark}




\section{Conclusions and Discussion}
\label{sec_conclusions}

We introduced the notion of removable and uniqueness sets for signal processing models on graphons. These notions allowed us to characterize the properties of subsets of $[0,1]$ that determine uniquely bandlimited graphon signals (Theorem~\ref{thm_uq_sets_gphonsp}). By building on these concepts, we studied the properties of uniqueness sets on families of graphs obtained from graphons using equipartition discretization methods. We showed that the goodness of sampling sets, represented in the graphon space, is approximately inherited by those graphs whose structural properties are similar -- i.e. small values of $\Vert \boldsymbol{T}_{W_{G_1}} - \boldsymbol{T}_{W_{G_2}}\Vert_{2}$ in Theorems~\ref{thm_removable_sets_seq} and~\ref{thm_removable_sets_twoG} --.

We proposed an algorithm (Algorithm~\ref{alg_sampt_method}) that leverages the optimal sampling sets of small graphs to generate approximately optimal sampling sets in larger graphs whose structural properties are tied to the small graph on the graphon space. We also showed that the uniqueness sets of a convergent sequence of graphs constitute a convergent sequence as well as long as the induced graphon representations of such sets are identical in $[0,1]$ (Theorem~\ref{thm_uniq_remov_sequence}). We proved that uniqueness sets are stable to node relabeling and edge droppings when the induced set representation in the graphon space is preserved (Sections~\ref{sub_sec_edgedroppings} and~\ref{sub_sec_noderelabeling}). This notion of stability guarantees that the change in constants that characterize the uniqueness sets is proportional to the measure of the relabeling and the edge droppings.

It is important to remark that while $\Vert \boldsymbol{T}_{W_{G_1}} - \boldsymbol{T}_{W_{G_2}}\Vert_{2}$ being small is a fundamental requirement for the bounds derived to be useful, such condition is guaranteed when the graphs under study are generated from the same graphon using the method (GD1). Additionally, notice that even when the graphs considered are generated by totally arbitrary means the term $\Vert \boldsymbol{T}_{W_{G_1}} - \boldsymbol{T}_{W_{G_2}}\Vert_{2}$ provides a formal measure of similarity that, if small, guarantees transferability of the uniqueness sets between different graphs.

The numerical experiments performed validated the results derived, showing that one can use optimal sampling techniques to estimate the uniqueness sets in graphs of medium or small size and then use such sets as a base to approximate close to optimal sampling sets of larger graphs. The results of the experiments performed are consistent for several graphon models, several bandlimited signal models, and when noise is added to the samples of the signals.

The contributions presented in this paper provide the guarantees necessary to overcome the current computational cost limitations associated with optimal sampling approaches. This is achieved, not by competing with such methods, but instead by formally showing that the properties of the optimal sampling sets can be reused to find approximately optimal sampling sets in larger graphs.

It is important to remark that Theorem~\ref{thm_removable_sets_seq} is valid to understand how identical uniqueness sets interrelate in the graphon space when considering arbitrary graphons. There is no assumption on the properties of the graphons themselves, which guarantees its applicability with those graphons that are not necessarily Lipschitz. This sets the starting point of an interesting research direction studying the interaction between interpolation methods on the graphon space and their relationship with reconstructions of signals on graphs. This may lead to better reconstruction approaches on large graphs in the presence of noise.

Finally, one particular topic that opens up for future exciting research directions is 
the study of sampling sets when considering a probabilistic interpretation of the graphon as a generator of graphs. We foresee that the insights in such problem could lead to specific characterizations of \textit{probably optimal} sampling sets, and they may lead to specific characterizations of the statistical moments of the removable sets constants when seen as a probability density function.



\appendices



\section{Proof of Theorem~\ref{thm_bernstein_ineq}}
\label{sec_app_Bernstein_ineq}
\begin{proof}

Using the spectral theorem  and taking into account that $\boldsymbol{x}$ is $\omega$-bandlimited we have

\begin{equation}
\boldsymbol{T}_{W}\boldsymbol{x}
                           =
                           \sum_{i=1}^{i_\omega}
                             \lambda_{i}\left(
                                          \boldsymbol{T}_{W}
                                         \right)
                                              \langle
                                                    \boldsymbol{x}
                                                     ,
                                             \boldsymbol{\varphi}_{W,i} 
                                              \rangle
                                              \boldsymbol{\varphi}_{W,i}
                                              ,
\end{equation}
where $i_{\omega}$ is the largest integer such that
$                    
\lambda_{i}\left(
                \boldsymbol{T}_{W}
           \right)
           >
           \omega
           .
$
Then, it follows that
\begin{equation}
\left\Vert 
        \boldsymbol{T}_{W}\boldsymbol{x}
\right\Vert^{2}_{2}
         =
\left\langle
         \boldsymbol{T}_{W}\boldsymbol{x}
          ,
         \boldsymbol{T}_{W}\boldsymbol{x}
\right\rangle_{L_2}
               =
               \sum_{i=1}^{i_\omega} 
                  \lambda_{i}\left(
                                   \boldsymbol{T}_{W}
                              \right)^{2} 
                                                                          \langle
                                                                                      \boldsymbol{x}
                                                                                      ,
                                                                                      \boldsymbol{\varphi}_{W,i} 
                                                                          \rangle_{L_2}^2
                                                                          .
\end{equation}
Since $\vert\lambda_{1}(\boldsymbol{T}_{W})\vert \geq \ldots \geq \vert\lambda_{i_\omega} (\boldsymbol{T}_{W})\vert \geq \omega$, we have
\begin{equation}
               \sum_{i=1}^{i_\omega}
                  \lambda_{i}\left( 
                                \boldsymbol{T}_{W}
                              \right)^{2} 
               \langle
               \boldsymbol{x}
               ,
               \boldsymbol{\varphi}_{W,i} 
               \rangle^2
               \geq
                        \omega^2 
                              \sum_{i=1}^{i_\omega} 
                              \langle
                              \boldsymbol{x}
                              ,
                              \boldsymbol{\varphi}_{W,i} 
                              \rangle^2
                              =
                              \omega^2 
                                 \Vert 
                                    \boldsymbol{x} 
                                 \Vert_{2}^2
                              ,
\end{equation}
which leads to 
$
\left\Vert 
         \boldsymbol{T}_{W}\boldsymbol{x}
\right\Vert_{2}^2
                \geq
                    \omega^2 
                       \Vert 
                          \boldsymbol{x} 
                       \Vert_{2}^2
$. Taking square root on both sides of this inequality we have
\begin{equation}
\left\Vert 
        \boldsymbol{T}_{W}\boldsymbol{x}
\right\Vert_{2}
            \geq
                \omega 
                    \Vert 
                        \boldsymbol{x} 
                    \Vert_{2}
                  .
\end{equation}

\end{proof}


\section{Proof Theorem~\ref{thm_uq_sets_gphonsp}}\label{sec_proof_uniqsets_graphons}

\begin{proof}
Let $(W,\boldsymbol{x}), (W,\boldsymbol{y})\in\mathcal{PW}_{\omega}(W)$. Since $\mathcal{PW}_{\omega}(W)$ is a vector space we have that $(W,\boldsymbol{x}-\boldsymbol{y})\in\mathcal{PW}_{\omega}(W)$. Then, using Theorem~\ref{thm_bernstein_ineq} it follows that
	\begin{equation}\label{eq_proof_uniqsets_0}
	\left\Vert 
	                \boldsymbol{T}_{W} \left( \boldsymbol{x} - \boldsymbol{y}\right)
	\right\Vert_{2}
	                   \geq
	                   \omega
	                   \Vert 
                              \boldsymbol{x} 
                              - 
                              \boldsymbol{y} 
                          \Vert_{2}
	                   .
	\end{equation}
    If $\boldsymbol{x} = \boldsymbol{y}$ on $\ccalS$, then $\boldsymbol{x} - \boldsymbol{y} = 0$ on $\ccalS$. If we assume $\boldsymbol{x} - \boldsymbol{y} \neq \boldsymbol{0} $ on $\ccalS^c$, then using of the concept of removable sets we have
    \begin{equation}\label{eq_proof_uniqsets_1}
    	\left\Vert 
    	                 \boldsymbol{T}_{W} \left( \boldsymbol{x} - \boldsymbol{y}\right)
    	\right\Vert_{2}
    	<
    	\Lambda_{\ccalS^c}
    	\Vert 
    	       \boldsymbol{x}
                  - 
                  \boldsymbol{y} 
    	\Vert_{2}
    	.
    \end{equation}
    Since by hypothesis $\omega>\Lambda_{\ccalS^c}$,~\eqref{eq_proof_uniqsets_1} leads to a contradiction. Therefore, $\boldsymbol{x} - \boldsymbol{y} = \boldsymbol{0} $.
	\end{proof}





\section{Proof of Theorem~\ref{thm_removable_sets_G_W}}\label{sec_proof_LambdaG_vs_LambdaW}

\begin{proof}

First let $\boldsymbol{x}\in L_{2}(\ccalS_{W_{G}})$ and $\bbx\in L_{2}(\ccalS_{G})$, where $(W_{G},\boldsymbol{x})$ is the graphon signal induced by $(G,\bbx)$. We start taking into account that
\begin{equation}\label{eq_sec_proof_LambdaG_vs_LambdaW_0}
\boldsymbol{T}_{W_G} \boldsymbol{x}
                        =
                           \sum_{i=1}^{\infty} 
                                       \lambda_i (\boldsymbol{T}_{W_G}) \boldsymbol{\varphi}_{W_{G},i} 
                                                     \left\langle 
                                                                  \boldsymbol{x}, \boldsymbol{\varphi}_{W_G , i}
                                                     \right\rangle
                                                     .
\end{equation}
From~\cite{Diao2016ModelfreeCO,lovasz2012large} we know that if the graphon $W_G$ is induced by the graph $G$ we have
\begin{equation}\label{eq_eigenvalues_G_WG}
\lambda_i \left( \boldsymbol{T}_{W_G} \right)
                          =
                             \frac{\lambda_i \left( \bbS_G\right)}{N}
                             ,
\end{equation}
\begin{equation}\label{eq_eigenvectors_G_WG}
\boldsymbol{\varphi}_{W_G , i} (t)
            =
            \sqrt{N}\sum_{\ell=1}^{N}\chi_{I_\ell}(t)\bbv_{G,i} (\ell)
            ,
\end{equation}
where $N=\vert V(G)\vert$ and $\bbv_{G,i} $ are the eigenvectors of $\bbS_{G}$. Substituting~\eqref{eq_eigenvalues_G_WG} and~\eqref{eq_eigenvectors_G_WG} into~\eqref{eq_sec_proof_LambdaG_vs_LambdaW_0} we have
\begin{multline}
\boldsymbol{T}_{W_G} \boldsymbol{x}
                        =
                          \sum_{i=1}^{\infty} 
                                    \frac{\lambda_i (\bbS_G)}{N}
                                                \left( 
                                                         \sqrt{N}\sum_{\ell=1}^{N}\chi_{I_\ell}(u)\bbv_{G,i} (\ell)
                                                \right) 
                                                \\
                                                         \int_{0}^{1}
                                                                     \boldsymbol{x}(t)
                                                                     \sqrt{N}\sum_{\ell=1}^{N}\chi_{I_\ell}(t)\bbv_{G,i} (\ell)
                                                          dt           
.
\end{multline}
Grouping and simplifying terms we have
\begin{multline*}
\boldsymbol{T}_{W_G} \boldsymbol{x}
                        =
                        \\
                           \sum_{i=1}^{\infty} 
                                     \lambda_i (\bbS_G)
\left( 
         \sum_{\ell=1}^{N}\chi_{I_\ell}(t)\bbv_{G,i} (\ell)
\right) 
\left(
          \sum_{\ell=1}^{N}
                               \left(
                                          \int_{I_\ell}\boldsymbol{x}(t)dt   
                               \right)        
\bbv_{G,i} (\ell)      
\right)  
.
\end{multline*}
Now, taking into account that
$
    \int_{I_\ell}\boldsymbol{x}(t) dt = \bbx(\ell)/N
    ,
$
we have that
\begin{multline}
\boldsymbol{T}_{W_G} \boldsymbol{x}
=
\\
\frac{1}{N}
\sum_{i=1}^{\infty} 
              \lambda_i (\bbS_G)
                       \left( 
                              \sum_{\ell=1}^{N}\chi_{I_\ell}(t)\bbv_{G,i} (\ell)
                       \right) 
                       \left(
                               \sum_{\ell=1}^{N}
                                              \bbx(\ell)      
                                              \bbv_{G,i} (\ell)      
                       \right)  
\\
= 
\frac{1}{N}
               \sum_{i=1}^{\infty} 
                                 \lambda_i (\bbS_G)
\left( 
           \sum_{\ell=1}^{N}\chi_{I_\ell}(t)\bbv_{G,i} (\ell)
\right) 
\left\langle
        \bbx , \bbv_{G,i} 
\right\rangle                     
.
\end{multline}
Then, it follows that
\begin{multline}
\left\Vert 
       \boldsymbol{T}_{W_G} \boldsymbol{x}
\right\Vert_{2}^{2}
          =
\left\langle
          \bbT_{W_G} \boldsymbol{x}
          ,
          \bbT_{W_G} \boldsymbol{x} 
\right\rangle    
         =   
         \\
\frac{1}{N}
              \sum_{\ell=1}^{N}
                         \left( 
                                 \frac{1}{N}
                                       \sum_{i=1}^{N}
                                                 \lambda_i (\bbS_G)
                                                 \bbv_{G,i}(\ell)
                                                 \left\langle \bbx, \bbv_{G,i}\right\rangle
                         \right)^2
                         .
\end{multline}
Now, expanding the product and organizing terms we have
\begin{multline}
\left\Vert 
       \boldsymbol{T}_{W_G} \boldsymbol{x}
\right\Vert_{2}^{2}
=
\frac{1}{N^3}
         \sum_{\ell=1}^{N}
                  \left( 
                            \sum_{i=1}^{N}
                                        \lambda_i (\bbS_G)^2 \bbv_{G,i}(\ell)^2 \langle \bbx, \bbv_{G,i}\rangle^2
                             \right.
                             \\           
                             \left.           
                            +
                            \sum_{i\neq j}
                                     \lambda_i (\bbS_G) \lambda_j (\bbS_G)
                                     \bbv_{G,i}(\ell) 
                                     \bbv_{G,j}(\ell)     
                                     \langle \bbx, \bbv_{G,i}\rangle   
                                     \langle \bbx, \bbv_{G,j}\rangle           
                  \right)
                  .
\end{multline}
Reorganizing the summation it follows that
\begin{multline*}
\left\Vert 
        \boldsymbol{T}_{W_G} \boldsymbol{x}
\right\Vert_{2}^{2}
=
\frac{1}{N^3}
\left( 
\sum_{i=1}^{N}
\lambda_i (\bbS_G)^2 
                            \left(\sum_{\ell=1}^{N} \bbv_{G,i}(\ell)^2 \right)
                \langle \bbx, \bbv_{G,i}\rangle^2
\right.
\\           
\left.           
+
\sum_{i\neq j}
            \lambda_i (\bbS_G) \lambda_j (\bbS_G)
                     \left( 
                               \sum_{\ell=1}^{N}
                                \bbv_{G,i}(\ell) 
                                \bbv_{G,j}(\ell)  
                      \right)   
\langle \bbx, \bbv_{G,i}\rangle   
\langle \bbx, \bbv_{G,j}\rangle           
\right)
.
\end{multline*}
Since
\begin{equation}
\sum_{\ell=1}^{N} \bbv_{G,i}(\ell)^2  
             =
             \Vert \bbv_{G,i} \Vert^2 
             =
             1
             ,
\end{equation}
\begin{equation}
           \sum_{\ell=1}^{N}
                               \bbv_{G,i}(\ell) 
                               \bbv_{G,j}(\ell)  
                         =
                                \langle
                                         \bbv_{G,i} , \bbv_{G,j}
                                \rangle  
                         =
                             0
                             ,         
\end{equation}
it follows that
\begin{multline}\label{eq_SWnorm_SGnorm}
\left\Vert 
      \boldsymbol{T}_{W_G} \boldsymbol{x}
\right\Vert_{2}^{2}
=
\frac{1}{N^3}
\left( 
\sum_{i=1}^{N}
\lambda_i (\bbS_G)^2 
\langle \bbx, \bbv_{G,i}\rangle^2                 
\right)
\\
=
\frac{1}{N^3}
\Vert \bbS_{G} \bbx \Vert_{2}^{2}
.
\end{multline}
Replacing~\eqref{eq_SWnorm_SGnorm} and $\Vert \boldsymbol{x}\Vert_{2}^{2} = \Vert \bbx \Vert_{2}^{2} /N$ into 
$
\left\Vert 
\boldsymbol{T}_{W_G} \boldsymbol{x}
\right\Vert_{2}
<
\Lambda_{\ccalS_{W_G}}
\left\Vert
\boldsymbol{x}
\right\Vert_{2}
,
$
we have
\begin{equation}
\frac{1}{N^{\frac{3}{2}}}
      \Vert 
            \bbS_{G} \bbx 
      \Vert_{2}
      <
       \frac{\Lambda_{\ccalS_{W_G}} }{\sqrt{N}}  
        \Vert \bbx \Vert_{2} 
        ,
\end{equation}
and simplifying terms we have
$
\Vert 
\bbS_{G} \bbx 
\Vert_{2}
<
N\Lambda_{\ccalS_{W_G}}
\Vert \bbx \Vert_{2} 
.
$
Since
\begin{equation}
\sup_{\bbx\in L_{2}(\ccalS_{G})}\left(
                         \frac{ \Vert \bbS_G \bbx \Vert_{2}}
                                {\Vert \bbx \Vert_{2}}
                  \right) 
                          = 
                            \Lambda_{\ccalS_G}
\end{equation}
we have
$
\Lambda_{\ccalS_G} 
               \leq 
                      N \Lambda_{\ccalS_{W_G}}
.
$

Now, replacing~\eqref{eq_SWnorm_SGnorm} and $\Vert \boldsymbol{x}\Vert_{2}^{2} = \Vert \bbx \Vert_{2}^{2} /N$ into 
$
\left\Vert 
\bbS_{G} \bbx
\right\Vert_{2}
<
\Lambda_{\ccalS_G}
\left\Vert
\bbx
\right\Vert_{2}
,
$
we have
\begin{equation}
N^{\frac{3}{2}}\left\Vert 
                                \boldsymbol{T}_{W_G} \boldsymbol{x}
                       \right\Vert_{2}
                       <
                       \Lambda_{\ccalS_G}
                       \sqrt{N}
                       \Vert
                               \boldsymbol{x}
                       \Vert_{2}
                       ,
\end{equation}
and simplifying terms we have
\begin{equation}
\left\Vert 
       \boldsymbol{T}_{W_G} \boldsymbol{x}
\right\Vert_{2}
<
\frac{\Lambda_{\ccalS_G}}{N}
\Vert
    \boldsymbol{x}
\Vert_{2}
.
\end{equation}
Since
\begin{equation}
\sup_{\boldsymbol{x}\in L_{2}(\ccalS_{W_G})}
         \left(\frac{\left\Vert 
                            \boldsymbol{T}_{W_G} \boldsymbol{x}
                         \right\Vert_{2}
                        }
                       {
                        \left\Vert
                               \boldsymbol{x}
                         \right\Vert_{2}
                        }
\right)
=
\Lambda_{\ccalS_{W_G}}
,
\end{equation}
it follows that
$
N\Lambda_{\ccalS_{W_G}} 
           \leq 
                 \Lambda_{\ccalS_G}
.
$

Then, since from previous results we have
$
\Lambda_{\ccalS_G} 
               \leq 
N \Lambda_{\ccalS_{W_G}}
$
it follows that $\Lambda_{\ccalS_G} = N\Lambda_{\ccalS_{W_G}}$.
\end{proof}


\section{Proof of~Theorem~\ref{thm_removable_sets_seq}}
\label{proof_thm_removable_sets_seq}
\begin{proof}

Since $\ccalS_{W_1}^{c} = \ccalS_{W_2}^{c}$ we denote this set by $\ccalS_{W}^{c}$. Let $\boldsymbol{x}\in L_{2}(\ccalS_{W}^{c})$. We start taking into account that
\begin{equation}\label{eq_proof_thm_removable_sets_seq_1}
\left\Vert 
          \boldsymbol{T}_{W_{1}}\boldsymbol{x}    
\right\Vert_2
     -
\left\Vert 
          \boldsymbol{T}_{W_{2}}\boldsymbol{x}    
\right\Vert_2     
\leq
\left\Vert
           \boldsymbol{T}_{W_{1}}\boldsymbol{x}
           -
           \boldsymbol{T}_{W_{2}}\boldsymbol{x}
\right\Vert_2
.
\end{equation}
Applying the operator norm property on the right-hand side of \eqref{eq_proof_thm_removable_sets_seq_1} we have
\begin{equation}\label{eq_proof_thm_removable_sets_seq_2}
\left\Vert 
          \boldsymbol{T}_{W_{1}}\boldsymbol{x}    
\right\Vert_2
-
\left\Vert 
          \boldsymbol{T}_{W_{2}}\boldsymbol{x}    
\right\Vert_2     
\leq
\left\Vert
            \boldsymbol{T}_{W_{1}}
             -
            \boldsymbol{T}_{W_{2}}
\right\Vert_2
\Vert
     \boldsymbol{x}
\Vert_2
.
\end{equation}
Then, dividing~\eqref{eq_proof_thm_removable_sets_seq_2} by $\Vert \boldsymbol{x}\Vert_{2}$ and rearranging the terms we have
\begin{equation}\label{eq_proof_thm_removable_sets_seq_3}
\frac{
\left\Vert 
             \boldsymbol{T}_{W_{1}}\boldsymbol{x}    
\right\Vert_2
}{\Vert \boldsymbol{x}\Vert_2}
\leq
\left\Vert
\boldsymbol{T}_{W_{1}}
-
\boldsymbol{T}_{W_{2}}
\right\Vert_2
+
\frac{
\left\Vert 
              \boldsymbol{T}_{W_{2}}\boldsymbol{x}    
\right\Vert_2  
}{\Vert \boldsymbol{x}\Vert_2}
.
\end{equation}
Considering the supremum over $\boldsymbol{x}$ on both sides of~\eqref{eq_proof_thm_removable_sets_seq_3} we reach
\begin{equation}\label{eq_proof_thm_removable_sets_seq_4} 
\Lambda_{\ccalS_{W_{1}}^c}
\leq
\left\Vert
\boldsymbol{T}_{W_{1}}
-
\boldsymbol{T}_{W_{2}}
\right\Vert_2
+
\Lambda_{\ccalS_{W_{2}}^c}
.
\end{equation}
By symmetry, we can substitute subindices in~\eqref{eq_proof_thm_removable_sets_seq_4} to obtain
\begin{equation}\label{eq_proof_thm_removable_sets_seq_5} 
\Lambda_{\ccalS_{W_{2}}^c}
\leq
\left\Vert
\boldsymbol{T}_{W_{1}}
-
\boldsymbol{T}_{W_{2}}
\right\Vert_2
+
\Lambda_{\ccalS_{W_{1}}^c}
.
\end{equation}
Now, we combine~\eqref{eq_proof_thm_removable_sets_seq_4} and~\eqref{eq_proof_thm_removable_sets_seq_5} to obtain
\begin{equation}
\Lambda_{\ccalS_{W_{2}}^c}
-
\left\Vert
\boldsymbol{T}_{W_{1}}
-
\boldsymbol{T}_{W_{2}}
\right\Vert_2
\leq
\Lambda_{\ccalS_{W_{1}}^c}
\leq
\left\Vert
\boldsymbol{T}_{W_{1}}
-
\boldsymbol{T}_{W_{2}}
\right\Vert_2
+
\Lambda_{\ccalS_{W_{2}}^c}
,
\end{equation}
which is equivalent to
\begin{equation}
\Lambda_{\ccalS_{W_{1}}^c}
-
\left\Vert
\boldsymbol{T}_{W_{1}}
-
\boldsymbol{T}_{W_{2}}
\right\Vert_2
\leq
\Lambda_{\ccalS_{W_{2}}^c}
\leq
\left\Vert
\boldsymbol{T}_{W_{1}}
-
\boldsymbol{T}_{W_{2}}
\right\Vert_2
+
\Lambda_{\ccalS_{W_{1}}^c}
.
\end{equation}
Since $0\leq\Lambda_{\ccalS_{W_1}^{c}}\leq \Vert \boldsymbol{T}_{W_1}\Vert_{2}$ and $0\leq\Lambda_{\ccalS_{W_2}^{c}}\leq \Vert \boldsymbol{T}_{W_2}\Vert_{2}$ we have that
\begin{multline}
\max
\left\lbrace 
0
,
    \Lambda_{\ccalS_{W_{2}}^c}
    -
    \left\Vert
       \boldsymbol{T}_{W_{1}}
        -
       \boldsymbol{T}_{W_{2}}
    \right\Vert_2
\right\rbrace
\leq
\Lambda_{\ccalS_{W_{1}}^c}
\leq
\\
\min
\left\lbrace 
     \Vert \boldsymbol{T}_{W_1} \Vert_2
     ,
     \left\Vert
          \boldsymbol{T}_{W_{1}}
           -
           \boldsymbol{T}_{W_{2}}
     \right\Vert_2
     +
     \Lambda_{\ccalS_{W_{2}}^c}
\right\rbrace     
,
\end{multline}
and
\begin{multline}
\max
\left\lbrace 
0
,
    \Lambda_{\ccalS_{W_{1}}^c}
    -
    \left\Vert
       \boldsymbol{T}_{W_{1}}
        -
       \boldsymbol{T}_{W_{2}}
    \right\Vert_2
\right\rbrace
\leq
\Lambda_{\ccalS_{W_{2}}^c}
\leq
\\
\min
\left\lbrace 
     \Vert \boldsymbol{T}_{W_2} \Vert_2
     ,
     \left\Vert
          \boldsymbol{T}_{W_{1}}
           -
           \boldsymbol{T}_{W_{2}}
     \right\Vert_2
     +
     \Lambda_{\ccalS_{W_{1}}^c}
\right\rbrace     
,
\end{multline}
\end{proof}


\section{Proof of Theorem~\ref{thm_removable_sets_twoG}}
\label{proof_thm_removable_sets_twoG}

In the light of Theorem~\ref{thm_removable_sets_G_W} we have that $\Lambda_{\ccalS_{W_{G_1}}^{c}}=\Lambda_{\ccalS_{G_1}}/\vert V(G_1)\vert$ and $\Lambda_{\ccalS_{W_{G_2}}^{c}}=\Lambda_{\ccalS_{G_2}}/\vert V(G_2)\vert$. Replacing these identities in~\eqref{eq_proof_thm_removable_sets_seq_4} and~\eqref{eq_proof_thm_removable_sets_seq_5} it follows that
\begin{equation}\label{eq_proof_thm_removable_sets_twoG_1}
\Lambda_{\ccalS_{G_{1}}^c}
\leq
\vert V(G_1)\vert
\left\Vert
\boldsymbol{T}_{W_{1}}
-
\boldsymbol{T}_{W_{2}}
\right\Vert_2
+
\Lambda_{\ccalS_{G_{2}}^c}
\frac{\vert V(G_1)\vert}{\vert V(G_2)\vert}
,
\end{equation}
and
\begin{equation}\label{eq_proof_thm_removable_sets_twoG_2}
\Lambda_{\ccalS_{G_{2}}^c}
\leq
\vert V(G_2)\vert
\left\Vert
\boldsymbol{T}_{W_{1}}
-
\boldsymbol{T}_{W_{2}}
\right\Vert_2
+
\Lambda_{\ccalS_{G_{1}}^c}
\frac{\vert V(G_2)\vert}{\vert V(G_1)\vert}
.
\end{equation}
Combining~\eqref{eq_proof_thm_removable_sets_twoG_1} and~\eqref{eq_proof_thm_removable_sets_twoG_2} it follows that
\begin{multline}
\Lambda_{\ccalS_{G_{2}}^c}
\frac{\vert V(G_1)\vert}{\vert V(G_2)\vert}
-
\vert V(G_1)\vert
\left\Vert
\boldsymbol{T}_{W_{1}}
-
\boldsymbol{T}_{W_{2}}
\right\Vert_2
\\
\leq 
\Lambda_{\ccalS_{G_{1}}^c}
\leq
\\
\vert V(G_1)\vert
\left\Vert
\boldsymbol{T}_{W_{1}}
-
\boldsymbol{T}_{W_{2}}
\right\Vert_2
+
\Lambda_{\ccalS_{G_{2}}^c}
\frac{\vert V(G_1)\vert}{\vert V(G_2)\vert}
,
\end{multline}
and taking into account that $0\leq\Lambda_{\ccalS_{W_1}^{c}}\leq \Vert \boldsymbol{T}_{W_1}\Vert_{2}$ and $0\leq\Lambda_{\ccalS_{W_2}^{c}}\leq \Vert \boldsymbol{T}_{W_2}\Vert_{2}$ we obtain
\begin{multline}
\Lambda_{\ccalS_{G_{2}}^c}
\frac{\vert V(G_1)\vert}{\vert V(G_2)\vert}
-
\vert V(G_1)\vert
\left\Vert
\boldsymbol{T}_{W_{1}}
-
\boldsymbol{T}_{W_{2}}
\right\Vert_2
\\
\leq 
\Lambda_{\ccalS_{G_{1}}^c}
\leq
\\
\vert V(G_1)\vert
\left\Vert
\boldsymbol{T}_{W_{1}}
-
\boldsymbol{T}_{W_{2}}
\right\Vert_2
+
\Lambda_{\ccalS_{G_{2}}^c}
\frac{\vert V(G_1)\vert}{\vert V(G_2)\vert}
,
\end{multline}


\section{Proof of Theorem~\ref{thm_uniq_remov_sequence}}
\label{proof_thm_uniq_remov_sequence}

In the light of Theorem~\ref{thm_removable_sets_seq} we have that
\begin{multline}
\max
\left\lbrace 
0
,
    \Lambda_{\ccalS_{W}^c}
    -
    \left\Vert
       \boldsymbol{T}_{W_{i}}
        -
       \boldsymbol{T}_{W}
    \right\Vert_2
\right\rbrace
\leq
\Lambda_{\ccalS_{W_{i}}^c}
\leq
\\
\min
\left\lbrace 
     \Vert \boldsymbol{T}_{W_1} \Vert_2
     ,
     \left\Vert
          \boldsymbol{T}_{W_{i}}
           -
           \boldsymbol{T}_{W}
     \right\Vert_2
     +
     \Lambda_{\ccalS_{W}^c}
\right\rbrace     
.
\end{multline}
Applying the limit 
$
    \left\Vert
       \boldsymbol{T}_{W_{i}}
        -
       \boldsymbol{T}_{W}
    \right\Vert_2 \to 0
$
on both sides of the inequality we reach $\Lambda_{\ccalS_{W_i}^{c}}=\Lambda_{\ccalS_{W}^{c}}$.

\bibliography{bibliography}

@article{puysampling,
title = "Random sampling of bandlimited signals on graphs",
journal = "Applied and Computational Harmonic Analysis",
volume = "44",
number = "2",
pages = "446 - 475",
year = "2018",
issn = "1063-5203",
doi = "https://doi.org/10.1016/j.acha.2016.05.005",
url = "http://www.sciencedirect.com/science/article/pii/S1063520316300215",
author = "G.~Puy and N.~Tremblay and R.~Gribonval and P.~Vandergheynst"
}

@INPROCEEDINGS{anis_conf, 
author={A. Anis and A. Gadde and A. Ortega}, 
booktitle={2014 IEEE International Conference on Acoustics, Speech and Signal Processing (ICASSP)}, 
title={Towards a sampling theorem for signals on arbitrary graphs}, 
year={2014}, 
volume={}, 
number={}, 
pages={3864-3868}, 
doi={10.1109/ICASSP.2014.6854325}, 
ISSN={1520-6149}, 
month={May},}

@ARTICLE{segarra_samlocagreg, 
author={A. G. Marques and S. Segarra and G. Leus and A. Ribeiro}, 
journal={IEEE Transactions on Signal Processing}, 
title={Sampling of Graph Signals With Successive Local Aggregations}, 
year={2016}, 
volume={64}, 
number={7}, 
pages={1832-1843}, 
doi={10.1109/TSP.2015.2507546}, 
ISSN={1053-587X}, 
month={April},}

@ARTICLE{chensamplingongraphs, 
author={S. Chen and R. Varma and A. Sandryhaila and J. Kovačević}, 
journal={IEEE Transactions on Signal Processing}, 
title={Discrete Signal Processing on Graphs: Sampling Theory}, 
year={2015}, 
volume={63}, 
number={24}, 
pages={6510-6523}, 
doi={10.1109/TSP.2015.2469645}, 
ISSN={1053-587X}, 
month={Dec},}

@article{tremblayAB17,
  author    = {N.~Tremblay and
               P.~O. Amblard and
               S. Barthelm{\'{e}}},
  title     = {Graph sampling with determinantal processes},
  journal   = {CoRR},
  volume    = {abs/1703.01594},
  year      = {2017},
  url       = {http://arxiv.org/abs/1703.01594},
  bibsource = {dblp computer science bibliography, http://dblp.org}
}

@article{pesensonams1,
	author = {Pesenson, Isaac},
	year = {2011},
	month = {11},
	pages = {},
	title = {Sampling in Paley-Wiener spaces on combinatorial graphs},
	volume = {361},
	journal = {Transactions of The American Mathematical Society - TRANS AMER MATH SOC},
	doi = {10.1090/S0002-9947-09-04937-X}
}

@Article{pesenson2009,
	author="Pesenson, I.",
	title="Variational Splines and Paley--Wiener Spaces onÂ Combinatorial Graphs",
	journal="Constructive Approximation",
	year="2009",
	month="Feb",
	day="01",
	volume="29",
	number="1",
	pages="1--21",
	issn="1432-0940",
	doi="10.1007/s00365-007-9004-9",
	url="https://doi.org/10.1007/s00365-007-9004-9"
}

@ARTICLE{pesenson2019,
       author = {{Pesenson}, I~Z.},
        title = "{Average sampling and average splines on combinatorial graphs}",
      journal = {arXiv e-prints},
         year = "2019",
        month = "Jan",
          eid = {arXiv:1901.08726},
        pages = {arXiv:1901.08726},
archivePrefix = {arXiv},
       eprint = {1901.08726},
 primaryClass = {math.FA},
       adsurl = {https://ui.adsabs.harvard.edu/\#abs/2019arXiv190108726P}
}

@article{fuhrpesenson,
	author = {H.~Fuhr and I.~Z. Pesenson},
	title = {Poincaré and Plancherel Polya Inequalities in Harmonic Analysis on Weighted Combinatorial Graphs},
	journal = {SIAM Journal on Discrete Mathematics},
	volume = {27},
	number = {4},
	pages = {2007-2028},
	year = {2013},
}

@INPROCEEDINGS{pesensonschrodinger, 
	author={I. Z. Pesenson}, 
	booktitle={2015 International Conference on Sampling Theory and Applications (SampTA)}, 
	title={Sampling solutions of Schrodinger equations on combinatorial graphs}, 
	year={2015}, 
	pages={82-85}, 
	doi={10.1109SAMPTA.2015.7148855}, 
	month={May},}

@article{pesenson2010,
	author={Pesenson, I.~Z.
	and Pesenson, Meyer Z.},
	title={Sampling Filtering and Sparse Approximations on Combinatorial Graphs},
	journal={Journal of Fourier Analysis and Applications},
	year={2010},
	month={Dec},
	day={01},
	volume={16},
	number={6},
	pages={921--942},
	issn={1531-5851},
}

@Article{pesenson_sampvechilbert,
author="Pesenson, I.",
title="Sampling of Band-Limited Vectors",
journal="Journal of Fourier Analysis and Applications",
year="2001",
month="Jan",
day="01",
volume="7",
number="1",
pages="93--100",
issn="1531-5851",
doi="10.1007/s00041-001-0007-9",
url="https://doi.org/10.1007/s00041-001-0007-9"
}

@Article{pesenson_maniflPW,
author="Pesenson, I.",
title="Poincar{\'e}-Type inequalities and reconstruction of Paley-Wiener functions on manifolds",
journal= "The Journal of Geometric Analysis",
year="2004",
month="Mar",
day="01",
volume="14",
number="1",
pages="101--121",
issn="1559-002X",
doi="10.1007/BF02921868",
url="https://doi.org/10.1007/BF02921868"
}

@article{feich_pesenson_samplhyperbolic,
author = {Feichtinger, H. and Pesenson, I.},
year = {2011},
month = {05},
pages = {},
title = {A Reconstruction Method for Band-Limited Signals on the Hyperbolic Plane},
volume = {4},
booktitle = {Sampl. Theory Signal Image Process.}
}

@INPROCEEDINGS{pesenriemman2017, 
author={I. Z. Pesenson}, 
booktitle={2017 International Conference on Sampling Theory and Applications (SampTA)}, 
title={Sampling and Weyl's Law on compact Riemannian manifolds}, 
year={2017}, 
volume={}, 
number={}, 
pages={92-95}, 
doi={10.1109/SAMPTA.2017.8024469}, 
ISSN={}, 
month={July},}

@article{ortega_proxies, 
author={A. Anis and A. Gadde and A. Ortega}, 
journal={IEEE Transactions on Signal Processing}, 
title={Efficient Sampling Set Selection for Bandlimited Graph Signals Using Graph Spectral Proxies}, 
year={2016}, 
volume={64}, 
number={14}, 
pages={3775-3789}, 
doi={10.1109/TSP.2016.2546233}, 
ISSN={1053-587X}, 
month={July},}

@book{ortega2022introduction,
	title={Introduction to Graph Signal Processing},
	author={Ortega, A.},
	isbn={9781108428132},
	lccn={2021038900},
	url={https://books.google.com/books?id=v7htEAAAQBAJ},
	year={2022},
	publisher={Cambridge University Press}
}

@ARTICLE{tsitsverobarbarossa, 
author={M. {Tsitsvero} and S. {Barbarossa} and P. {Di Lorenzo}}, 
journal={IEEE Transactions on Signal Processing}, 
title={Signals on Graphs: Uncertainty Principle and Sampling}, 
year={2016}, 
volume={64}, 
number={18}, 
pages={4845-4860}, 
doi={10.1109/TSP.2016.2573748}, 
ISSN={1053-587X}, 
month={Sep.},}

@INPROCEEDINGS{6638704, 
author={S. K. {Narang} and A. {Gadde} and A. {Ortega}}, 
booktitle={2013 IEEE International Conference on Acoustics, Speech and Signal Processing}, 
title={Signal processing techniques for interpolation in graph structured data}, 
year={2013}, 
volume={}, 
number={}, 
pages={5445-5449}, 
doi={10.1109/ICASSP.2013.6638704}, 
ISSN={1520-6149}, 
month={May},}

@incollection{bookgspchapsampling,
  author      = "P. Di Lorenzo and S. Barbarossa and and P. Banelli",
  title       = "Sampling and Recovery of Graph Signals",
  editor      = "P. Djuric and C.Richard",
  booktitle   = "Cooperative and Graph Signal Processing",
  publisher   = "Elsevier",
  address     = " ",
  year        = 2018,
  pages       = " ",
  chapter     = " " ,
}

@ARTICLE{7055883, 
author={X. {Wang} and P. {Liu} and Y. {Gu}}, 
journal={IEEE Transactions on Signal Processing}, 
title={Local-Set-Based Graph Signal Reconstruction}, 
year={2015}, 
volume={63}, 
number={9}, 
pages={2432-2444}, 
doi={10.1109/TSP.2015.2411217}, 
ISSN={1053-587X}, 
month={May},}

@ARTICLE{8047995, 
author={L. F. O. {Chamon} and A. {Ribeiro}}, 
journal={IEEE Transactions on Signal Processing}, 
title={Greedy Sampling of Graph Signals}, 
year={2018}, 
volume={66}, 
number={1}, 
pages={34-47}, 
doi={10.1109/TSP.2017.2755586}, 
ISSN={1053-587X}, 
month={Jan},}

@ARTICLE{7581102, 
author={S. {Chen} and R. {Varma} and A. {Singh} and J. {Kovačević}}, 
journal={IEEE Transactions on Signal and Information Processing over Networks}, 
title={Signal Recovery on Graphs: Fundamental Limits of Sampling Strategies}, 
year={2016}, 
volume={2}, 
number={4}, 
pages={539-554}, 
doi={10.1109/TSIPN.2016.2614903}, 
ISSN={2373-776X}, 
month={Dec},}

@article{JAYAWANT2022108436,
title = {Practical graph signal sampling with log-linear size scaling},
journal = {Signal Processing},
volume = {194},
pages = {108436},
year = {2022},
issn = {0165-1684},
doi = {https://doi.org/10.1016/j.sigpro.2021.108436},
url = {https://www.sciencedirect.com/science/article/pii/S0165168421004734},
author = {Ajinkya Jayawant and Antonio Ortega}
}

@phdthesis{alejopm_phdthesis,
	author={Parada-Mayorga,Alejandro},
	year={2019},
	title={Blue Noise and Optimal Sampling on Graphs},
	journal={ProQuest Dissertations and Theses},
	pages={239},
	isbn={9781687974693},
	language={English},
	url={https://proxy.library.upenn.edu/login?url=https://www.proquest.com/dissertations-theses/blue-noise-optimal-sampling-on-graphs/docview/2307785173/se-2?accountid=14707},
}

@INPROCEEDINGS{alejopm_BN_c1,
author={Parada-Mayorga, Alejandro and Lau, Daniel L. and Giraldo, Jhony H. and Arce, Gonzalo R.},
booktitle={2019 13th International conference on Sampling Theory and Applications (SampTA)}, 
title={Blue-Noise Sampling of Signals on Graphs}, 
year={2019},
volume={},
number={},
pages={1-5},
doi={10.1109/SampTA45681.2019.9030829}}

@INPROCEEDINGS{alejopm_BN_c2,
author={Parada-Mayorga, Alejandro and Lau, Daniel L. and Giraldo, Jhony H. and Arce, Gonzalo R.},
booktitle={2019 IEEE Data Science Workshop (DSW)}, 
title={Sampling of Graph Signals with Blue Noise Dithering}, 
year={2019},
volume={},
number={},
pages={150-154},
doi={10.1109/DSW.2019.8755603}}

@INPROCEEDINGS{alejopm_cographs_c,
author={Guillot, Dominique and Parada-Mayorga, Alejandro and Cioaba, Sebastian and Arce, Gonzalo R.},
booktitle={2019 IEEE Data Science Workshop (DSW)}, 
title={Optimal Sampling Sets in Cographs}, 
year={2019},
volume={},
number={},
pages={165-169},
doi={10.1109/DSW.2019.8755569}}

@ARTICLE{alejopm_BN_j,
author={Parada-Mayorga, Alejandro and Lau, Daniel L. and Giraldo, Jhony H. and Arce, Gonzalo R.},
journal={IEEE Transactions on Signal and Information Processing over Networks}, 
title={Blue-Noise Sampling on Graphs}, 
year={2019},
volume={5},
number={3},
pages={554-569},
doi={10.1109/TSIPN.2019.2922852}}

@ARTICLE{graphon_pooling_j,
  author={Parada-Mayorga, Alejandro and Wang, Zhiyang and Ribeiro, Alejandro},
  journal={IEEE Transactions on Signal Processing}, 
  title={Graphon Pooling for Reducing Dimensionality of Signals and Convolutional Operators on Graphs}, 
  year={2023},
  volume={71},
  number={},
  pages={3577-3591},
  doi={10.1109/TSP.2023.3318471}}

@INPROCEEDINGS{graphon_pooling_c,
  author={Parada-Mayorga, Alejandro and Ruiz, Luana and Ribeiro, Alejandro},
  booktitle={2020 28th European Signal Processing Conference (EUSIPCO)}, 
  title={Graphon Pooling in Graph Neural Networks}, 
  year={2021},
  volume={},
  number={},
  pages={860-864},
  doi={10.23919/Eusipco47968.2020.9287735}}

@ARTICLE{graphon_geert_j,
author={Morency, Matthew W. and Leus, Geert},
journal={IEEE Transactions on Signal Processing}, 
title={Graphon Filters: Graph Signal Processing in the Limit}, 
year={2021},
volume={69},
number={},
pages={1740-1754},
doi={10.1109/TSP.2021.3061575}}

@article{Diao2016ModelfreeCO,
	title={Model-free consistency of graph partitioning},
	author={Peter Diao and Dominique Guillot and Apoorva Khare and Bala Rajaratnam},
	journal={arXiv: Combinatorics},
	year={2016}
}

@book{lovasz2012large,
  author="Lov{\'a}sz, L.",
  title="Large networks and graph limits",
  volume="60",
  year="2012",
  publisher="American Mathematical Society"
}

@book{gao2019graphon,
	title={Graphon Control Theory for Linear Systems on Complex Networks and Related Topics},
	author={Gao, S.},
	series={McGill theses},
	url={https://books.google.com/books?id=MtUgzQEACAAJ},
	year={2019},
	publisher={McGill University Libraries}
}

@ARTICLE{gdsp_moura,
author={Sandryhaila, Aliaksei and Moura, José M. F.},
journal={IEEE Transactions on Signal Processing}, 
title={Discrete Signal Processing on Graphs}, 
year={2013},
volume={61},
number={7},
pages={1644-1656},
doi={10.1109/TSP.2013.2238935}}

@ARTICLE{alejopm_algnn_j,
author={Parada-Mayorga, Alejandro and Ribeiro, Alejandro},
journal={IEEE Transactions on Signal Processing}, 
title={Algebraic Neural Networks: Stability to Deformations}, 
year={2021},
volume={69},
number={},
pages={3351-3366},
doi={10.1109/TSP.2021.3084537}}

@INPROCEEDINGS{alejopm_algnn_c,
	author={Parada-Mayorga, Alejandro and Ribeiro, Alejandro},
	booktitle={ICASSP 2021 - 2021 IEEE International Conference on Acoustics, Speech and Signal Processing (ICASSP)}, 
	title={Stability of Algebraic Neural Networks to Small Perturbations}, 
	year={2021},
	volume={},
	number={},
	pages={5205-5209},
	doi={10.1109/ICASSP39728.2021.9414604}}

@ARTICLE{alejopm_algnn_nc_j,
author={Parada-Mayorga, Alejandro and Butler, Landon and Ribeiro, Alejandro},
journal={IEEE Transactions on Signal Processing}, 
title={Convolutional Filters and Neural Networks With Noncommutative Algebras}, 
year={2023},
volume={71},
number={},
pages={2683-2698},
doi={10.1109/TSP.2023.3293716}}

@ARTICLE{alejopm_agggnn_j,
author={Parada-Mayorga, Alejandro and Wang, Zhiyang and Gama, Fernando and Ribeiro, Alejandro},
journal={IEEE Transactions on Signal and Information Processing over Networks}, 
title={Stability of Aggregation Graph Neural Networks}, 
year={2023},
volume={},
number={},
pages={1-16},
doi={10.1109/TSIPN.2023.3341408}}

@INPROCEEDINGS{owerko1,
  author={Owerko, Damian and Gama, Fernando and Ribeiro, Alejandro},
  booktitle={ICASSP 2020 - 2020 IEEE International Conference on Acoustics, Speech and Signal Processing (ICASSP)}, 
  title={Optimal Power Flow Using Graph Neural Networks}, 
  year={2020},
  volume={},
  number={},
  pages={5930-5934},
  doi={10.1109/ICASSP40776.2020.9053140}}

@INPROCEEDINGS{owerko2,
  author={Owerko, Damian and Gama, Fernando and Ribeiro, Alejandro},
  booktitle={2018 IEEE Global Conference on Signal and Information Processing (GlobalSIP)}, 
  title={Predicting Power Outages Using Graph Neural Networks}, 
  year={2018},
  volume={},
  number={},
  pages={743-747},
  doi={10.1109/GlobalSIP.2018.8646486}}

@INPROCEEDINGS{owerko3,
  author={Owerko, Damian and Gama, Fernando and Ribeiro, Alejandro},
  booktitle={ICASSP 2024 - 2024 IEEE International Conference on Acoustics, Speech and Signal Processing (ICASSP)}, 
  title={Unsupervised Optimal Power Flow Using Graph Neural Networks}, 
  year={2024},
  volume={},
  number={},
  pages={6885-6889},
  doi={10.1109/ICASSP48485.2024.10446827}}
\bibliographystyle{unsrt}


\clearpage
\newpage



\section{Supplementary Material}


\begin{figure*}
	%
	%
	\centering
	\begin{subfigure}{.32\linewidth}
		\centering
		\includegraphics[width=1\textwidth]{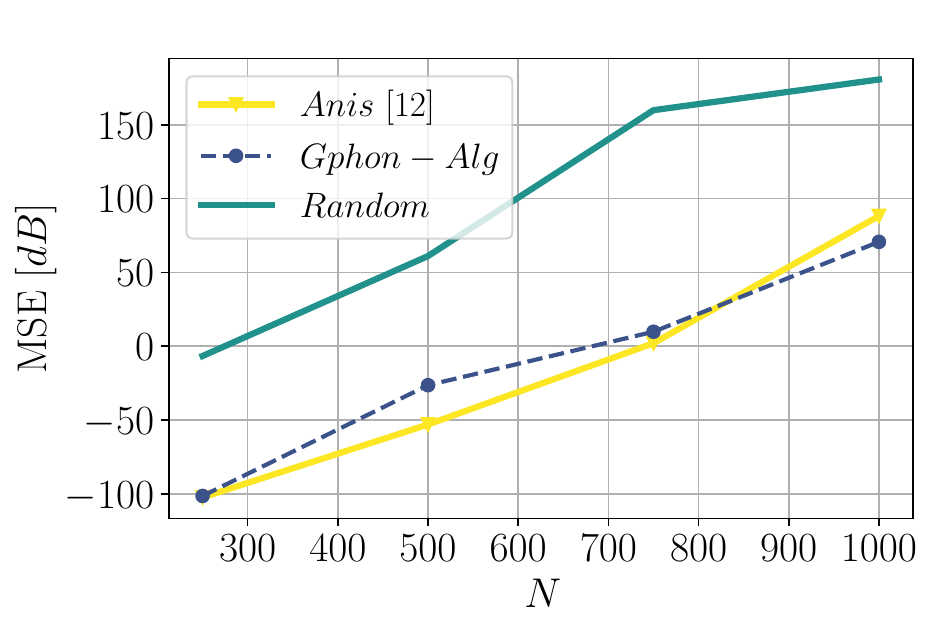} 
	\end{subfigure}
	\begin{subfigure}{.32\linewidth}
		\centering
		\includegraphics[width=1\textwidth]{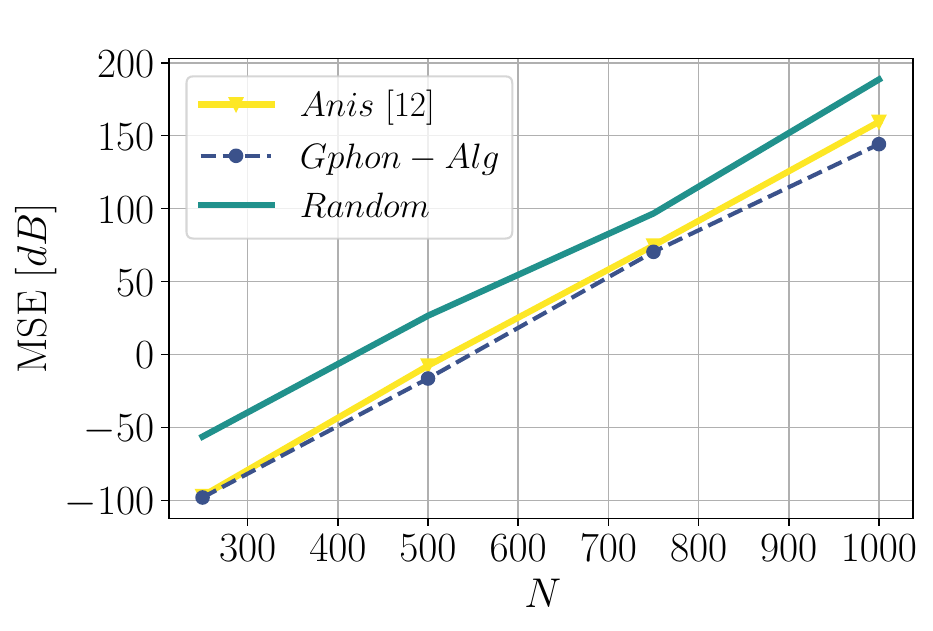} 
	\end{subfigure}
	\begin{subfigure}{.32\linewidth}
		\centering
		\includegraphics[width=1\textwidth]{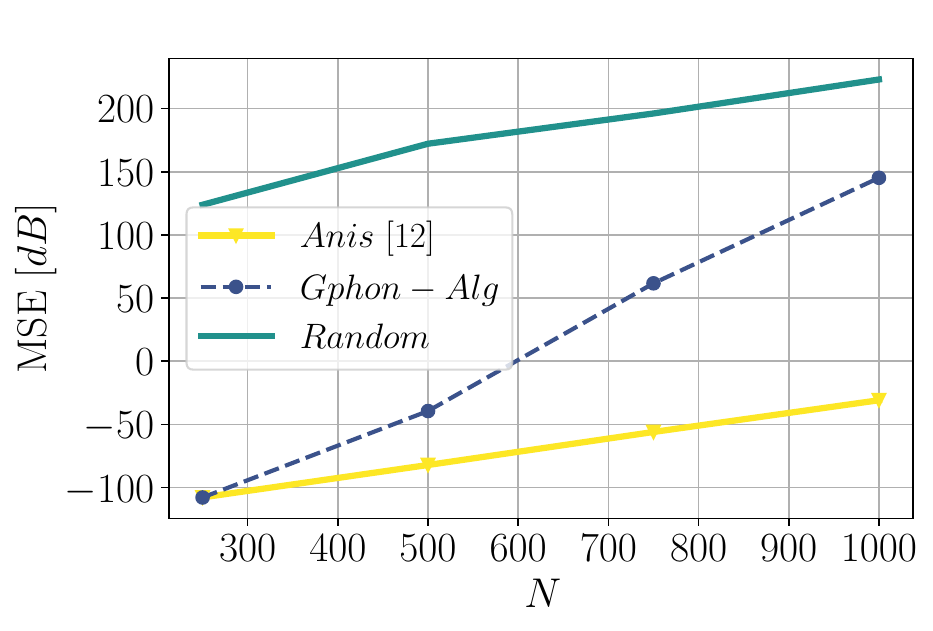} 
	\end{subfigure}
	%
	%
	\centering
	\begin{subfigure}{.32\linewidth}
		\centering
		\includegraphics[width=1\textwidth]{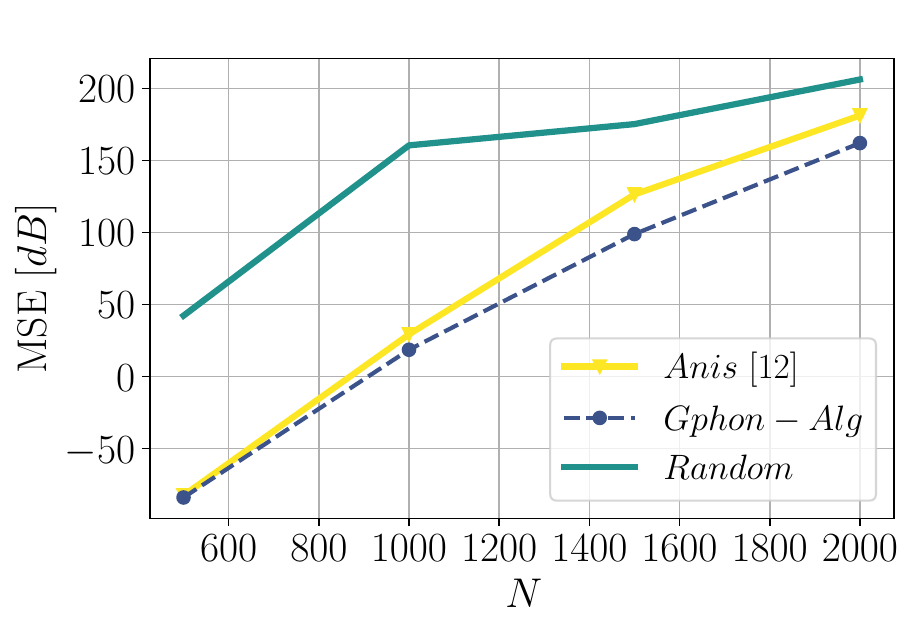} 
		\caption*{$W_{1}(u,v)=\left\vert\sin\left(100uv\right)\right\vert$}
	\end{subfigure}
	\begin{subfigure}{.32\linewidth}
		\centering
		\includegraphics[width=1\textwidth]{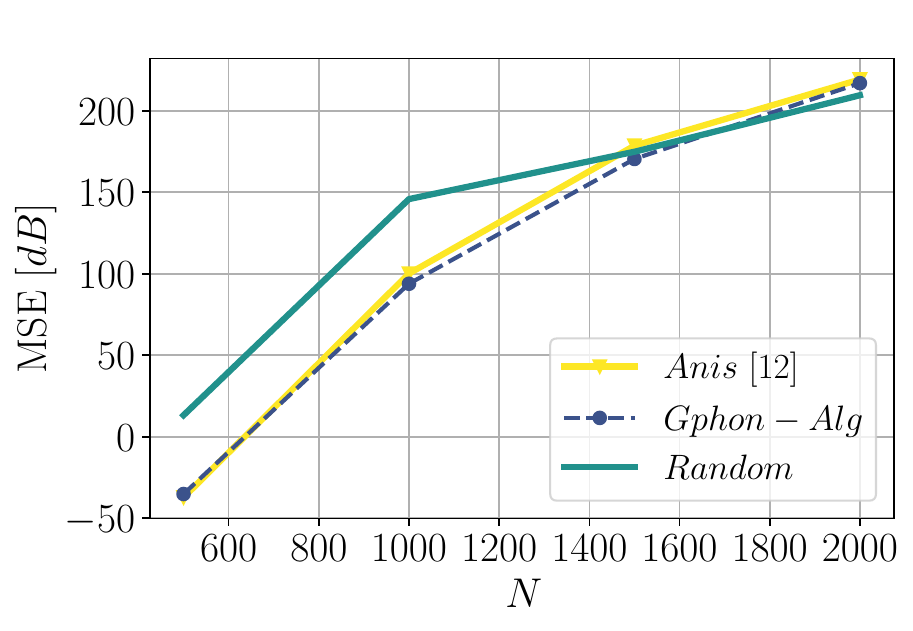} 
		\caption*{$W_{2}(u,v)=\left\vert\sin\left(64uv\right)\right\vert/2+\left\vert\cos\left(64uv\right)\right\vert/2$}
	\end{subfigure}
	\begin{subfigure}{.32\linewidth}
		\centering
		\includegraphics[width=1\textwidth]{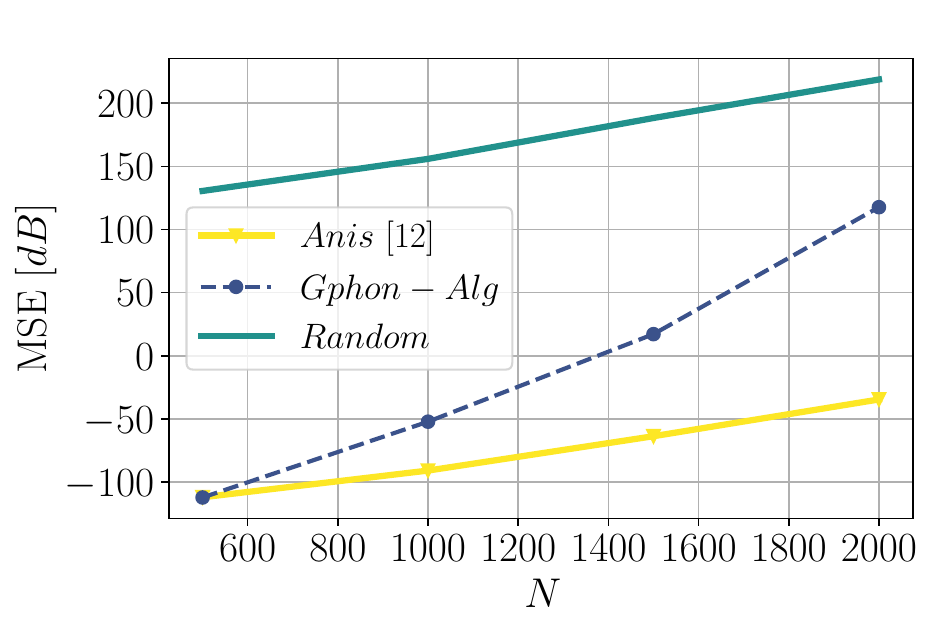} 
		\caption*{$W_{3}(u,v)=\left\vert\sin\left(10uv\right)\right\vert/2+\left\vert\cos\left(10uv\right)\right\vert/2$}
	\end{subfigure}
	\caption{Reconstruction error of sampled bandlimited signals on graphs derived from a graphon. Each column is associated with experiments performed with a particular graphon, $W_1$ for the left column, $W_2$ for the centered column, and $W_3$ for the column on the right. In the first row, the graphs generated have $250, 500, 750$ and $1000$ nodes. In the second row, the graphs generated have $500, 1000, 1500$ and $2000$ nodes. The graphs are obtained from the graphon using the discretization method (GD1). The bandwidth model considered in this figure is~\textbf{BWM1}. The axis $N$ indicates the number of nodes in the graph. The MSE value depicted is averaged over $50$, which is the number of signals used for each reconstruction. The sampling rate is $5\%$.}
	\label{fig_error_rec_exp_4a}
\end{figure*}



\begin{figure*}
	%
	%
	\centering
	\begin{subfigure}{.32\linewidth}
		\centering
		\includegraphics[width=1\textwidth]{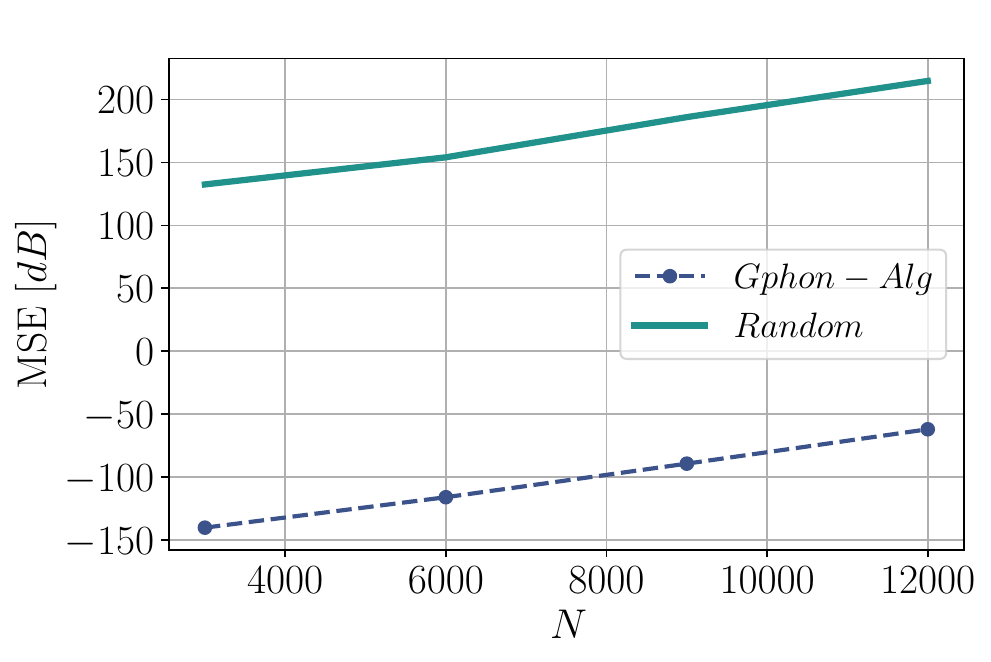}
            \caption*{$W_{1}(u,v)=(u+v)/2$}
	\end{subfigure}
	\begin{subfigure}{.32\linewidth}
		\centering
		\includegraphics[width=1\textwidth]{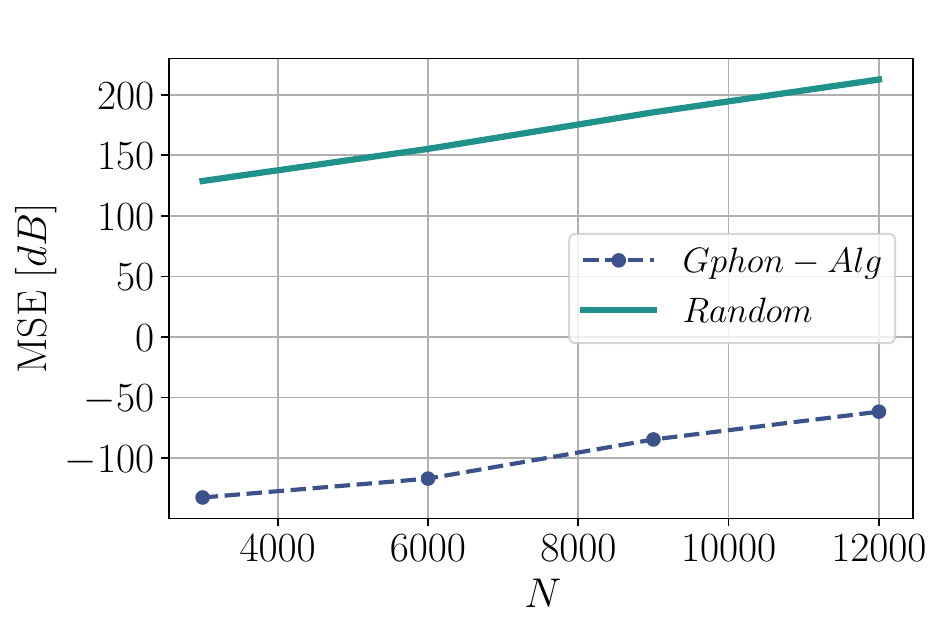} 
            \caption*{$W_{2}(u,v)=\min (u,v)\left( 1-\max(u,v)\right)$}
	\end{subfigure}
	\begin{subfigure}{.32\linewidth}
		\centering
		\includegraphics[width=1\textwidth]{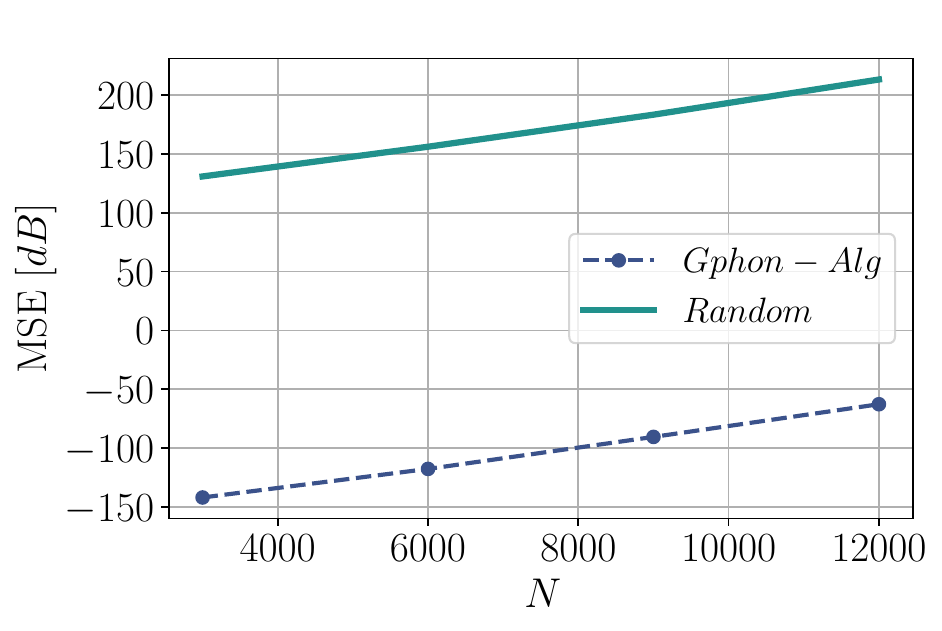} 
            \caption*{$W_{3}(u,v)=\left( u^2 +v^2 \right)/2$}
	\end{subfigure}
	%
	%
	\centering
	\begin{subfigure}{.32\linewidth}
		\centering
		\includegraphics[width=1\textwidth]{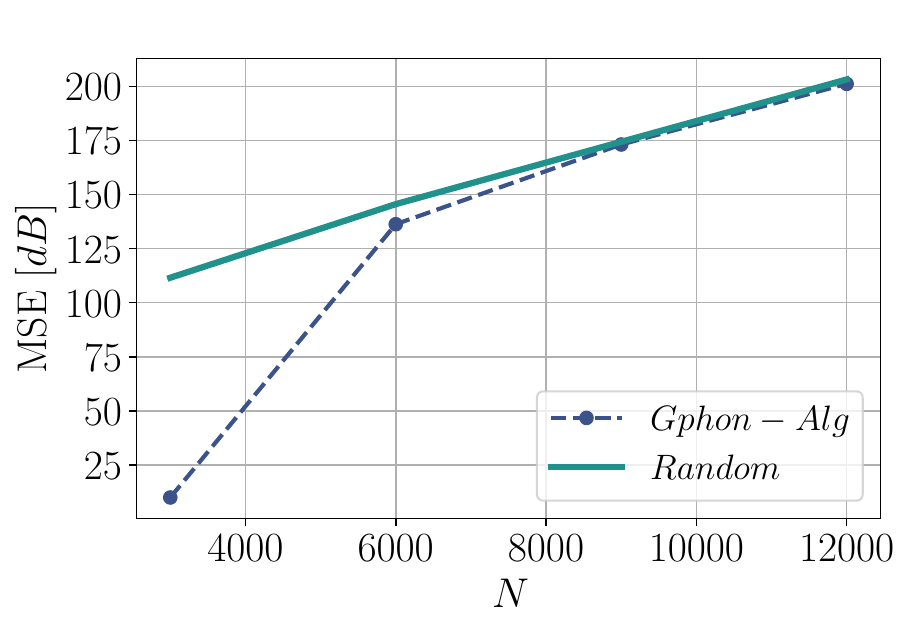} 
		\caption*{$W_{4}(u,v)=\left\vert\sin\left(100uv\right)\right\vert$}
	\end{subfigure}
	\begin{subfigure}{.32\linewidth}
		\centering
		\includegraphics[width=1\textwidth]{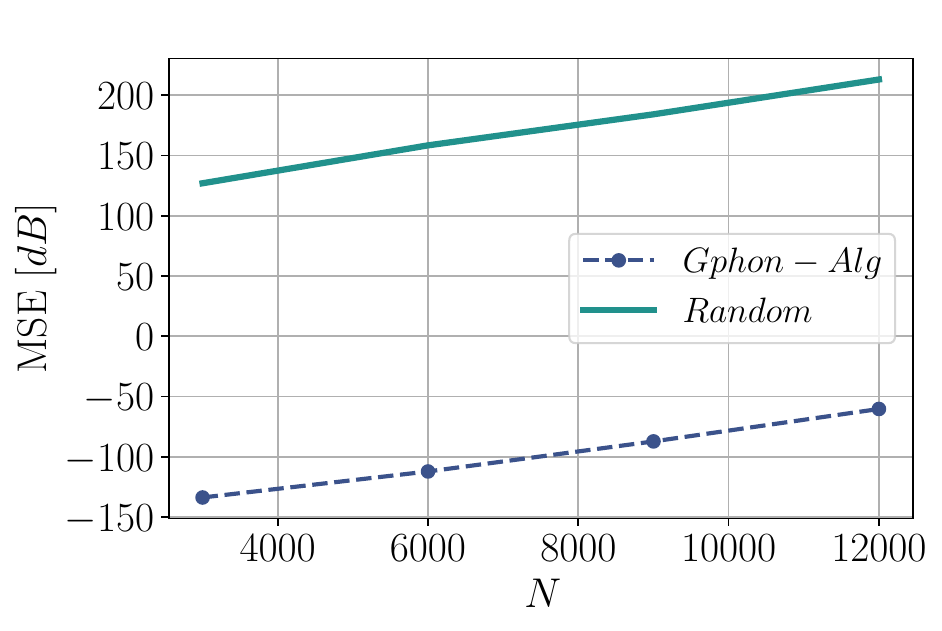} 
		\caption*{$W_{5}(u,v)=1-\max(u,v)$}
	\end{subfigure}
	\begin{subfigure}{.32\linewidth}
		\centering
		\includegraphics[width=1\textwidth]{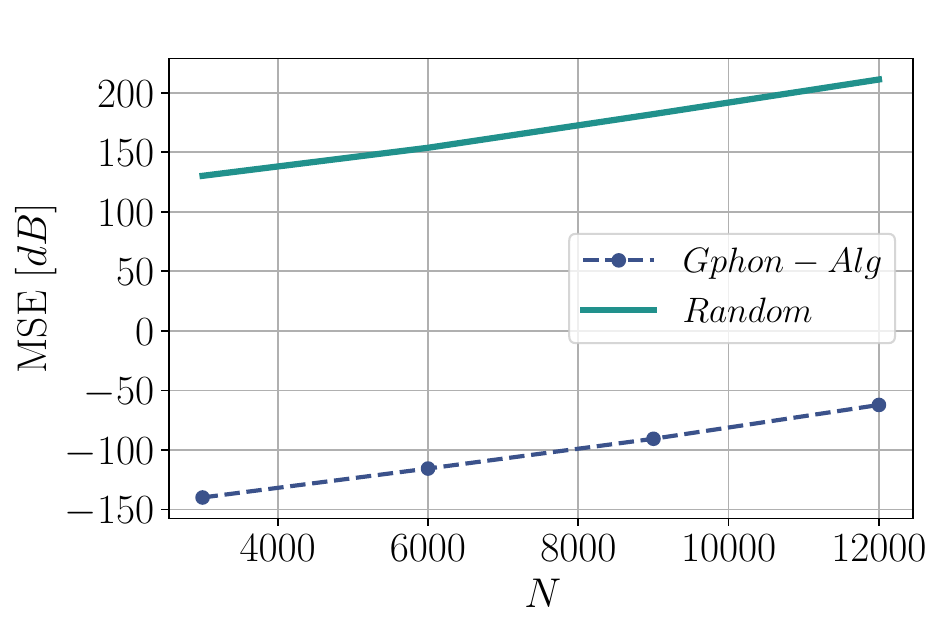} 
		\caption*{$W_{6}(u,v)=\left\vert\sin\left(10uv\right)\right\vert/2+\left\vert\cos\left(10uv\right)\right\vert/2$}
	\end{subfigure}
	\caption{Reconstruction error of sampled bandlimited signals on graphs derived from a graphon. Experiments are performed considering six graphons, generating graphs with $N=3000, N=6000, N=9000$ and $N=12000$ nodes. The graphs are obtained from the graphon using the discretization method (GD1). The bandwidth model considered in this figure is~\textbf{BWM2}. The optimal sampling set is calculated by~\cite{ortega_proxies} on the graph with $N=3000$ nodes, and it is used in Algorithm~\ref{alg_sampt_method} to obtain approximately optimal sampling sets when $N=6000$, $N=9000$, and $N=12000$. The MSE value depicted is averaged over $50$, which is the number of signals used for each reconstruction. The sampling rate is $5\%$.}
	\label{fig_error_rec_exp_bigraphs}
\end{figure*}



\begin{figure*}
	%
	%
	\centering
	\begin{subfigure}{.32\linewidth}
		\centering
		\includegraphics[width=1\textwidth]{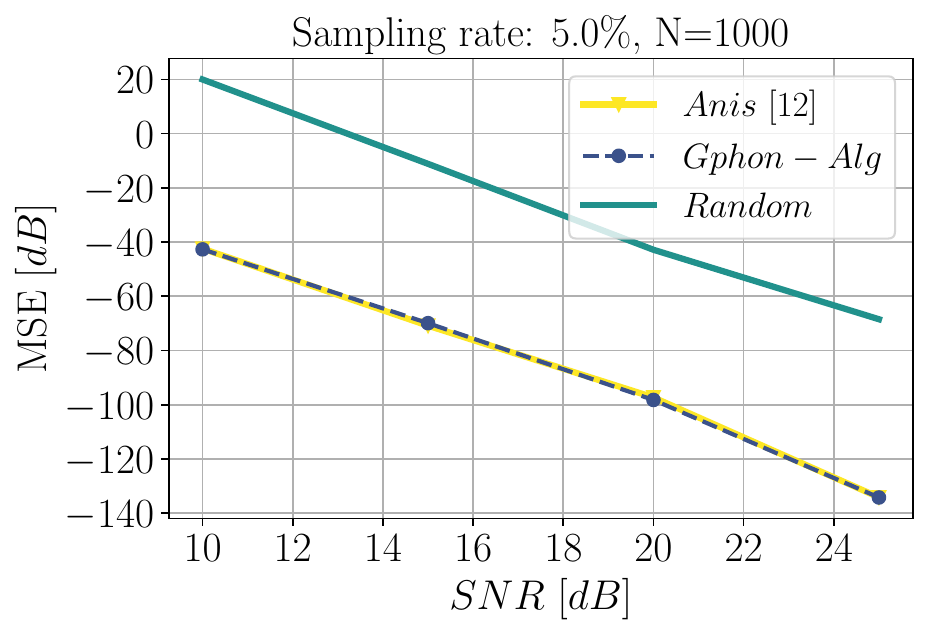} 
	\end{subfigure}
	\begin{subfigure}{.32\linewidth}
		\centering
		\includegraphics[width=1\textwidth]{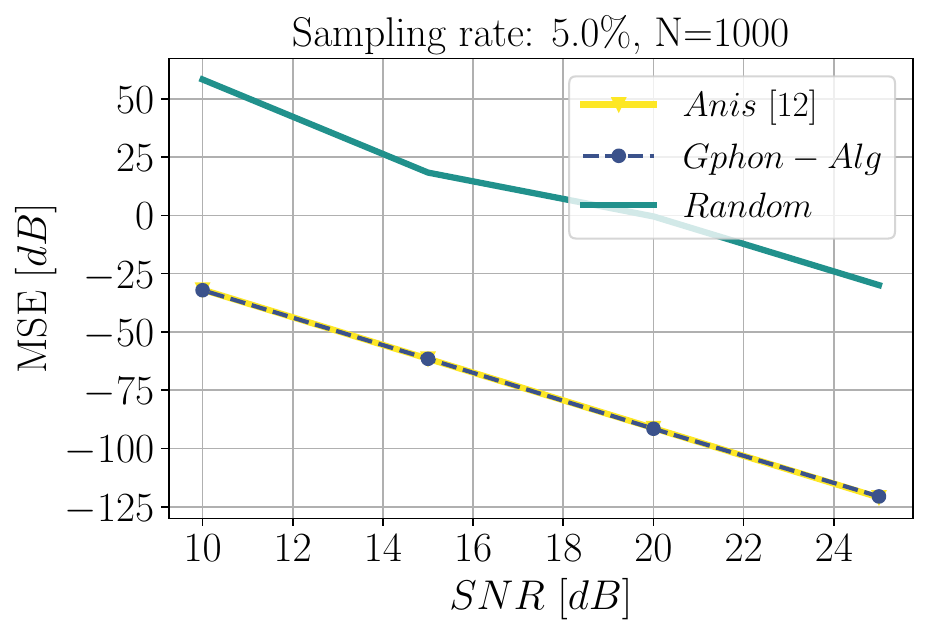} 
	\end{subfigure}
	\begin{subfigure}{.32\linewidth}
		\centering
		\includegraphics[width=1\textwidth]{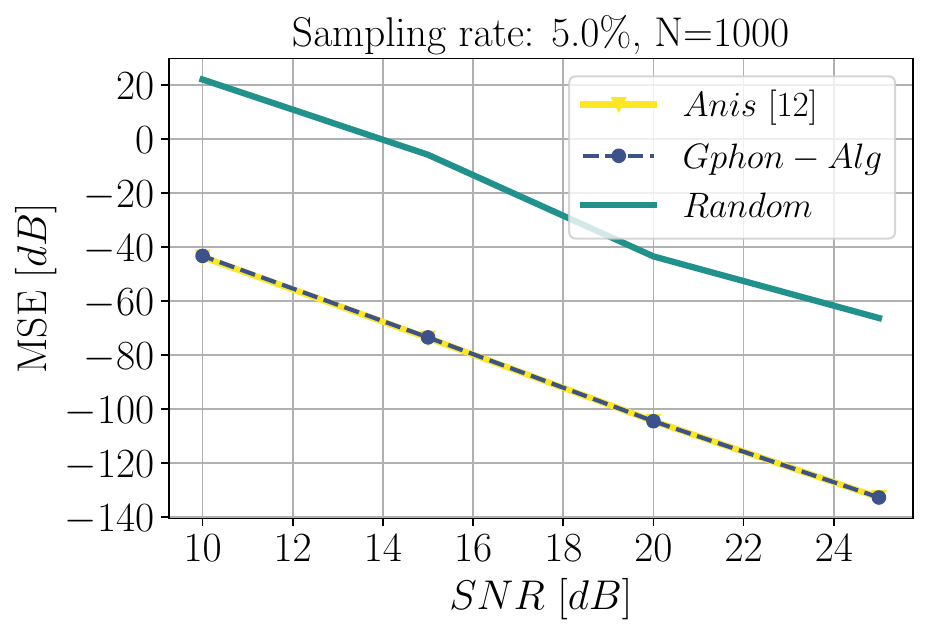} 
	\end{subfigure}
	%
	%
	\centering
	\begin{subfigure}{.32\linewidth}
		\centering
		\includegraphics[width=1\textwidth]{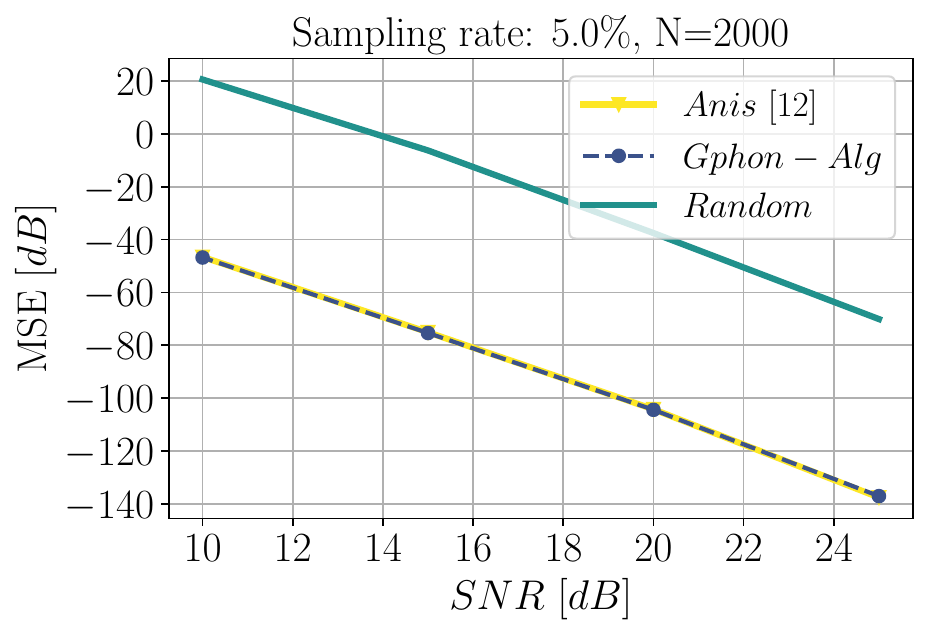} 
	\end{subfigure}
	\begin{subfigure}{.32\linewidth}
		\centering
		\includegraphics[width=1\textwidth]{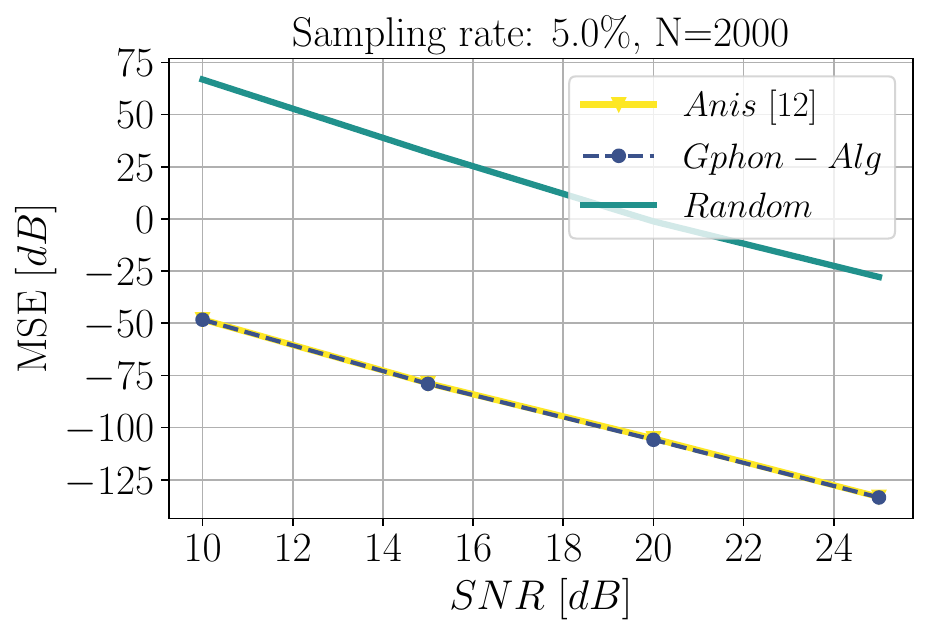} 
	\end{subfigure}
	\begin{subfigure}{.32\linewidth}
		\centering
		\includegraphics[width=1\textwidth]{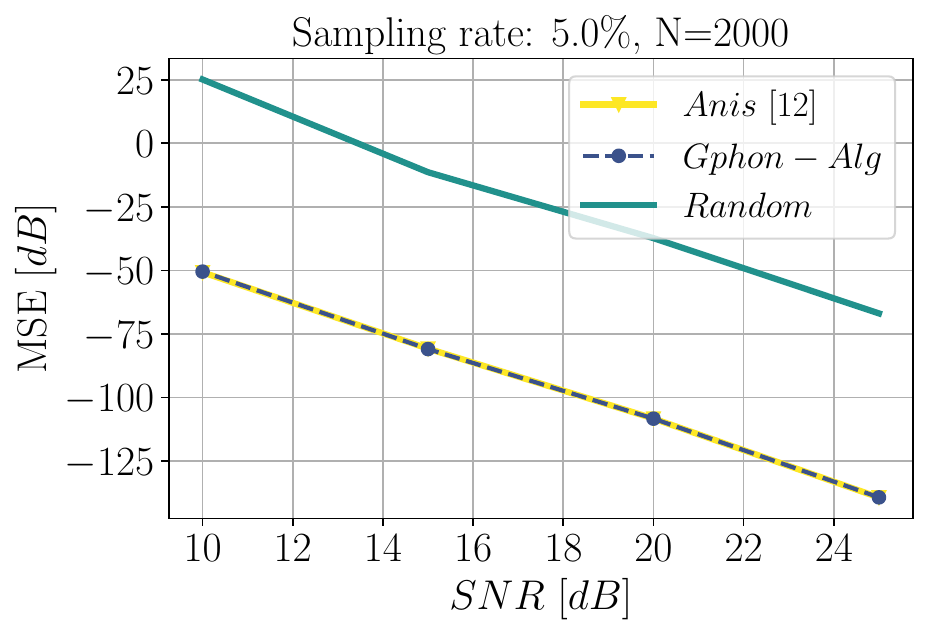} 
	\end{subfigure}
        %
	%
	\centering
	\begin{subfigure}{.32\linewidth}
		\centering
		\includegraphics[width=1\textwidth]{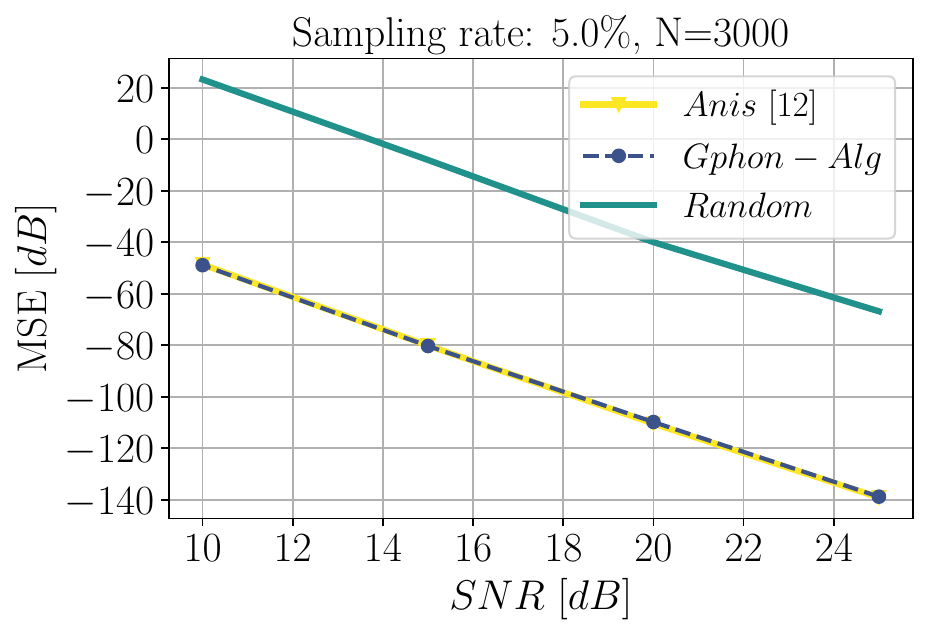} 
		\caption*{$W_{1}(u,v)=\left\vert\sin\left(100uv\right)\right\vert$}
	\end{subfigure}
	\begin{subfigure}{.32\linewidth}
		\centering
		\includegraphics[width=1\textwidth]{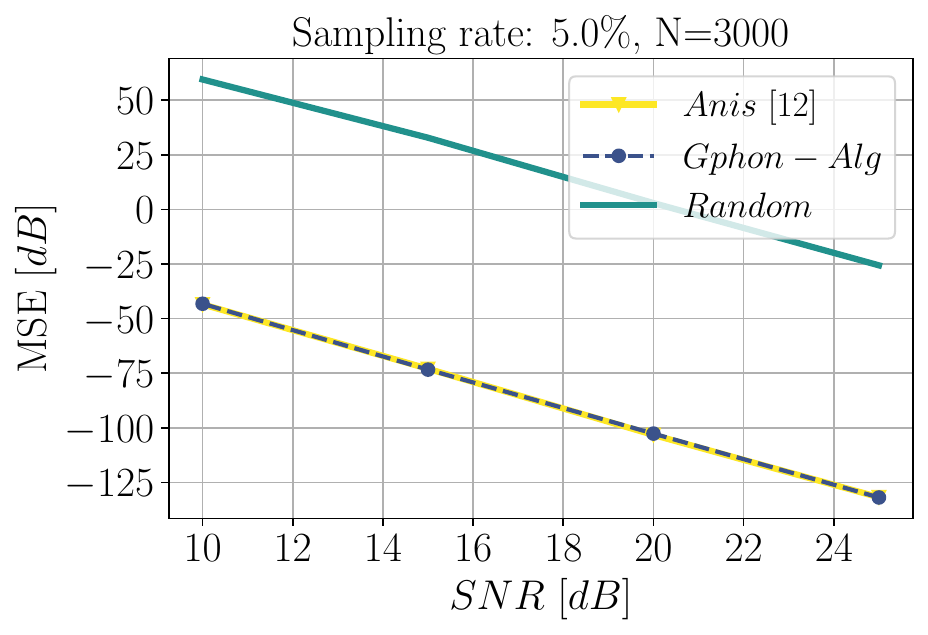} 
		\caption*{$W_{2}(u,v)=\left\vert\sin\left(64uv\right)\right\vert/2+\left\vert\cos\left(64uv\right)\right\vert/2$}
	\end{subfigure}
	\begin{subfigure}{.32\linewidth}
		\centering
		\includegraphics[width=1\textwidth]{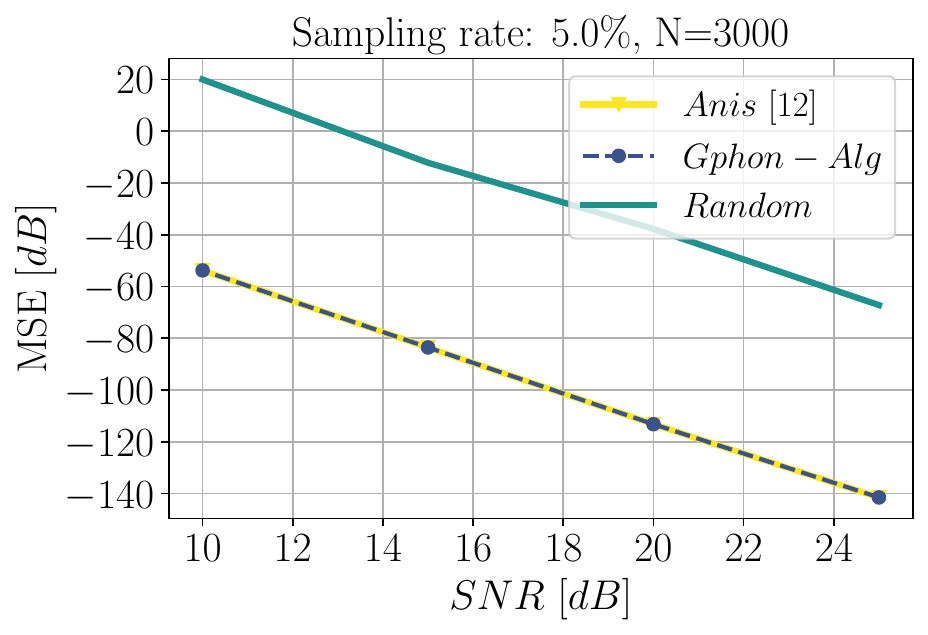} 
		\caption*{$W_{3}(u,v)=\left\vert\sin\left(10uv\right)\right\vert/2+\left\vert\cos\left(10uv\right)\right\vert/2$}
	\end{subfigure}
	\caption{Robustness of the approximately optimal sampling sets from Algorithm~\ref{alg_sampt_method} when the graphs are affected by random perturbations on the edge weights. In each picture, we perform sampling experiments comparing Algorithm~\ref{alg_sampt_method} against random sampling and the method proposed in~\cite{ortega_proxies}. Considering the bandwith model \textbf{BWM2} we generate four graphs, $\left\lbrace G_{i}^{W} \right\rbrace_{i=1}^{4}$, with a number of nodes $N$ and from a given graphon $W(u,v)$. In each $G_{i}^{W}$ we pollute the edges with noise, using the four SNR values given by $SNR=\{ 10, 15, 20, 25 \} [dB]$. To generate the graphs we use the method (GD1). In each column, the pictures presented are associated with the graphon displayed at the bottom and in each row, we display pictures regarding experiments with a given number of nodes. For each sampling experiment on the graph $G_{i}^{W}$ with $N$ nodes, the sampling set is obtained applying Algorithm~\ref{alg_sampt_method} using the optimal sampling set in a graph $G_{0}^{W}$ with $N/2$ nodes generated from $W$.}
    \label{fig_noise_robustness_a}
\end{figure*}



\begin{figure*}
	%
	%
	\centering
	\begin{subfigure}{.32\linewidth}
		\centering
		\includegraphics[width=1\textwidth]{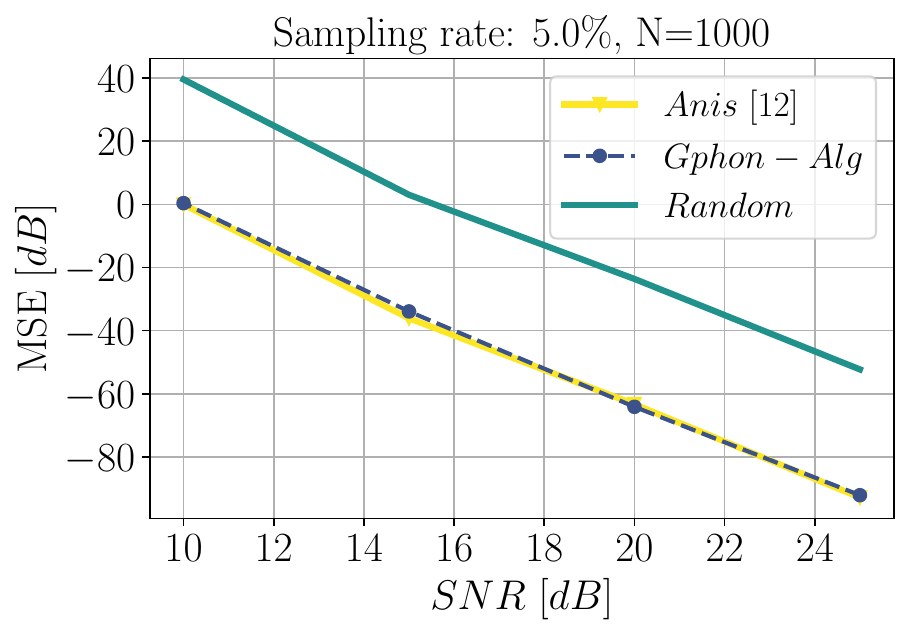} 
	\end{subfigure}
	\begin{subfigure}{.32\linewidth}
		\centering
		\includegraphics[width=1\textwidth]{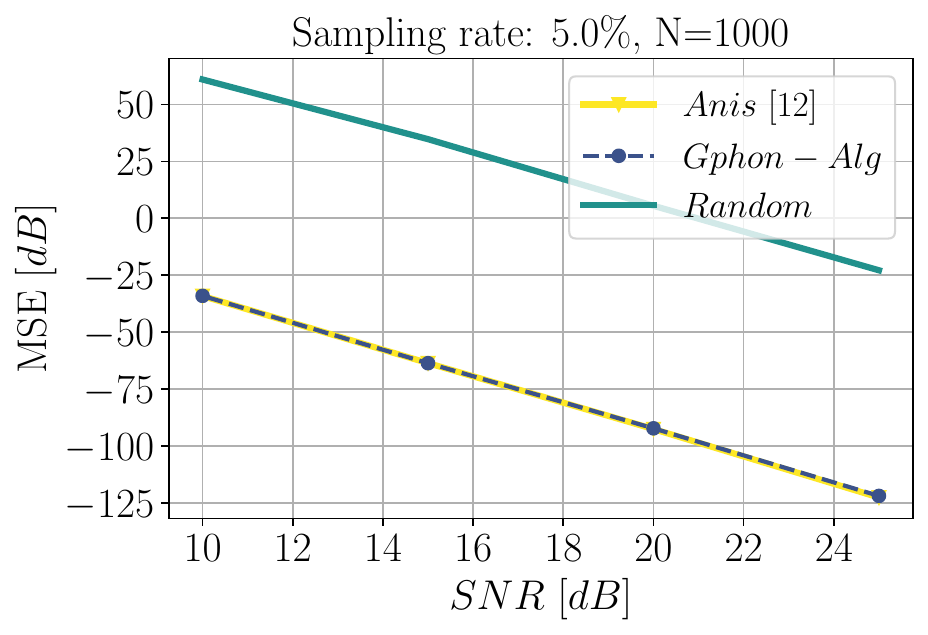} 
	\end{subfigure}
	\begin{subfigure}{.32\linewidth}
		\centering
		\includegraphics[width=1\textwidth]{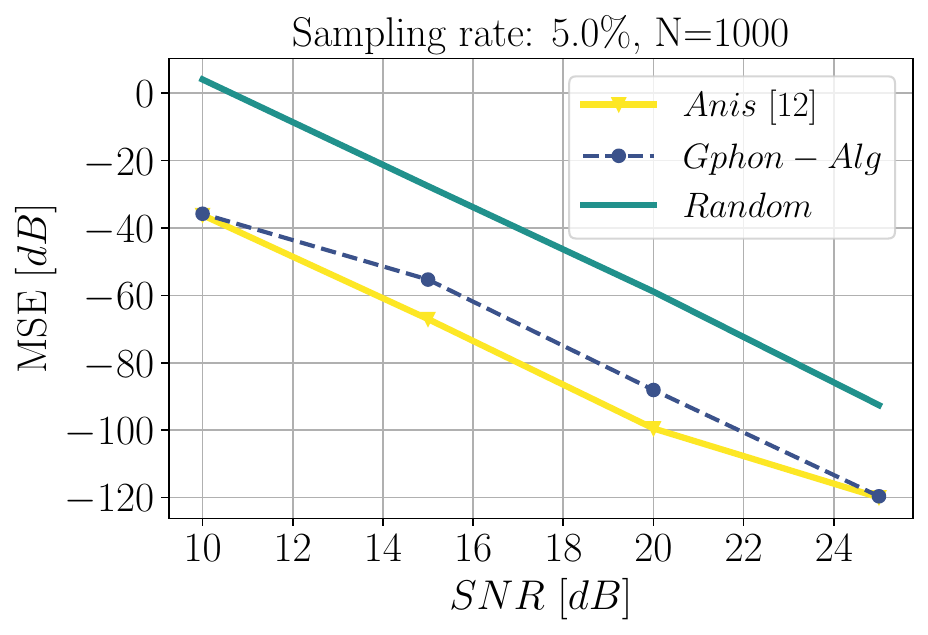} 
	\end{subfigure}
	%
	%
	\centering
	\begin{subfigure}{.32\linewidth}
		\centering
		\includegraphics[width=1\textwidth]{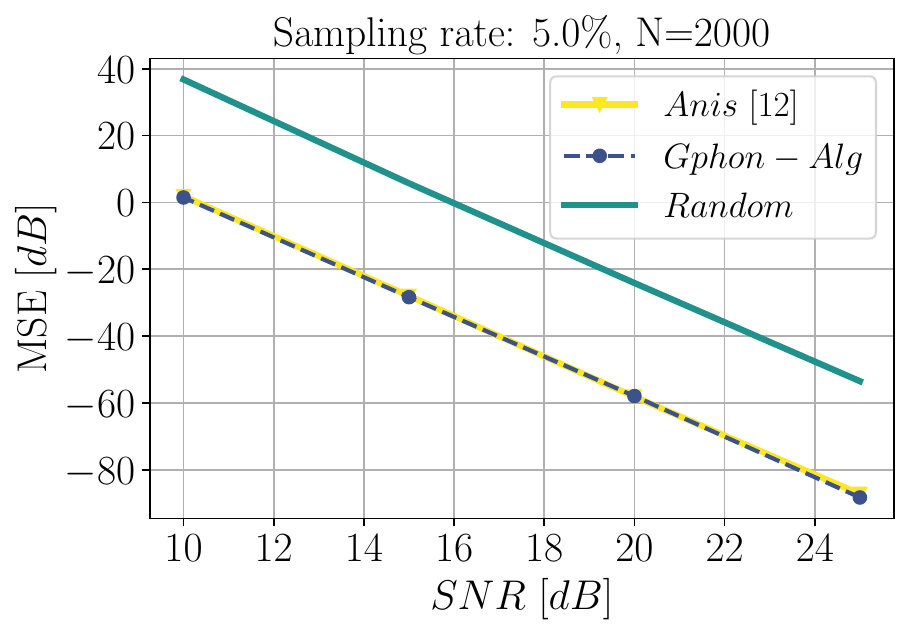} 
	\end{subfigure}
	\begin{subfigure}{.32\linewidth}
		\centering
		\includegraphics[width=1\textwidth]{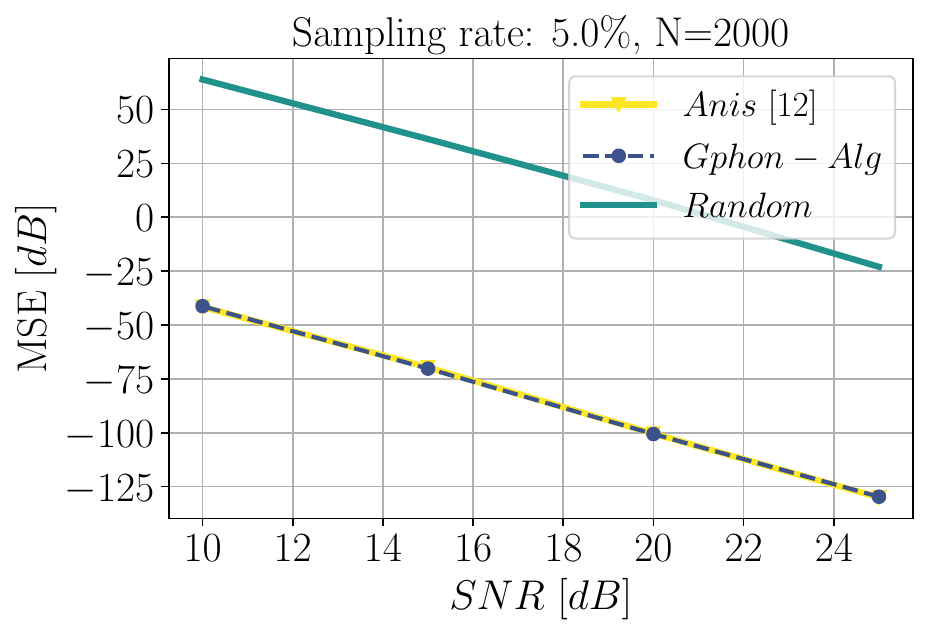} 
	\end{subfigure}
	\begin{subfigure}{.32\linewidth}
		\centering
		\includegraphics[width=1\textwidth]{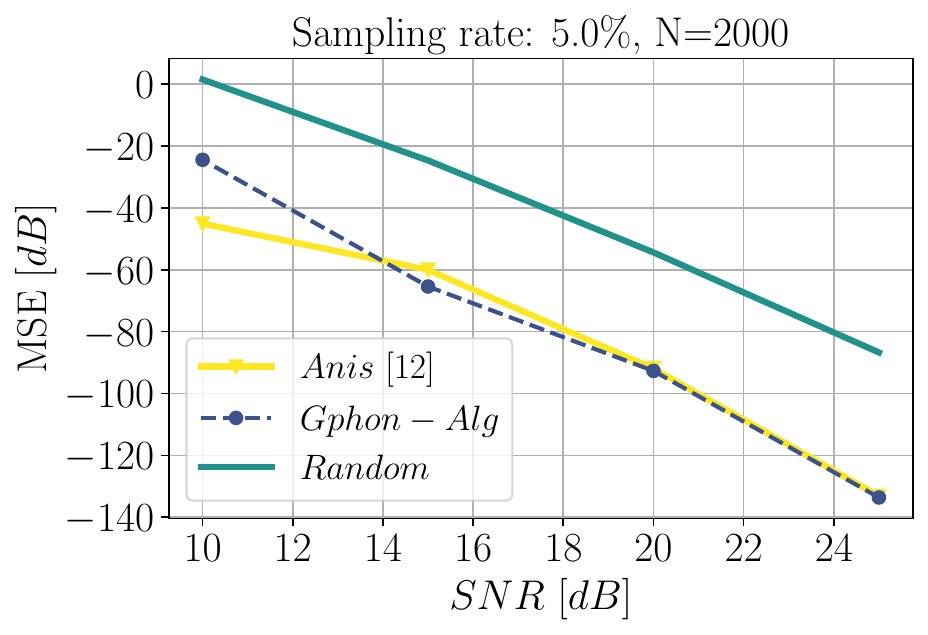} 
	\end{subfigure}
        %
	%
	\centering
	\begin{subfigure}{.32\linewidth}
		\centering
		\includegraphics[width=1\textwidth]{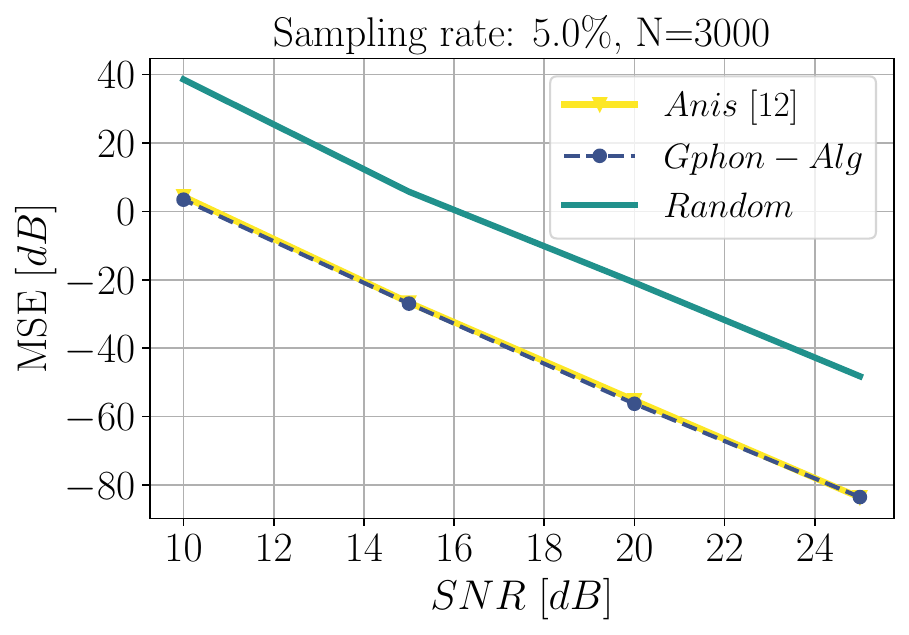} 
		\caption*{$W_{1}(u,v)=\left\vert\sin\left(100uv\right)\right\vert$}
	\end{subfigure}
	\begin{subfigure}{.32\linewidth}
		\centering
		\includegraphics[width=1\textwidth]{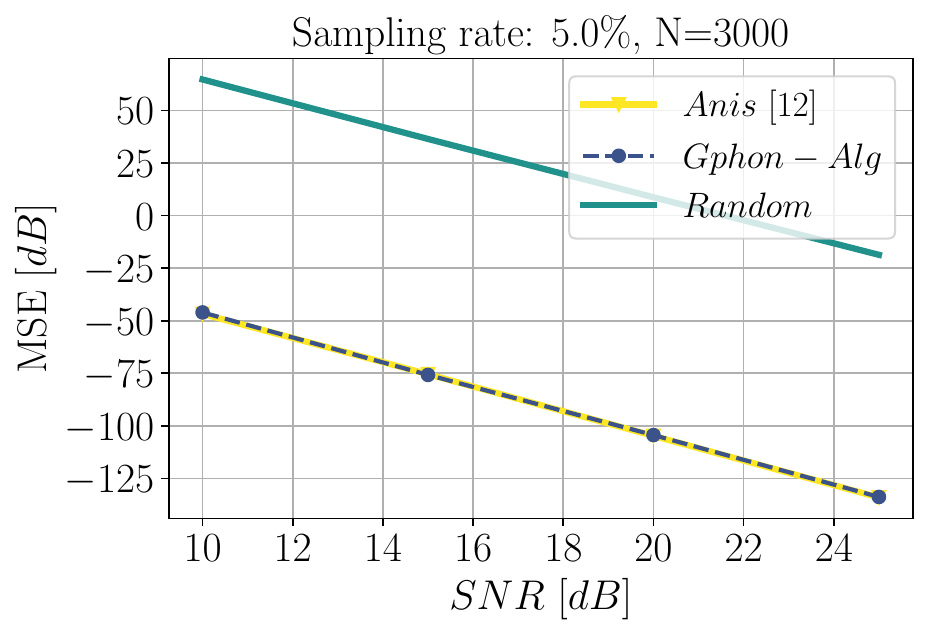} 
		\caption*{$W_{2}(u,v)=\left\vert\sin\left(64uv\right)\right\vert/2+\left\vert\cos\left(64uv\right)\right\vert/2$}
	\end{subfigure}
	\begin{subfigure}{.32\linewidth}
		\centering
		\includegraphics[width=1\textwidth]{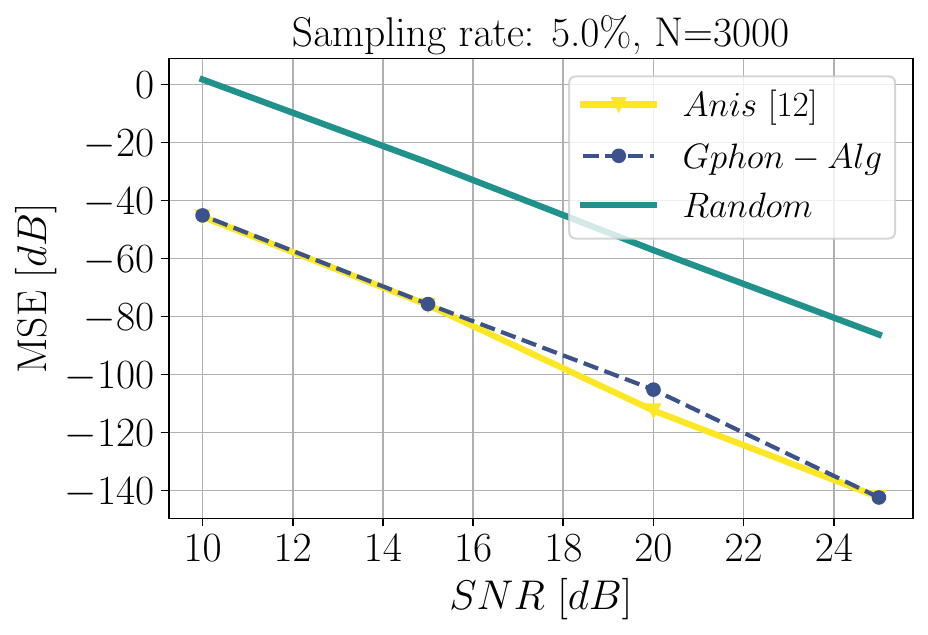} 
		\caption*{$W_{3}(u,v)=\left\vert\sin\left(10uv\right)\right\vert/2+\left\vert\cos\left(10uv\right)\right\vert/2$}
	\end{subfigure}
	\caption{Robustness of the approximately optimal sampling sets from Algorithm~\ref{alg_sampt_method} when the graphs are affected by random perturbations on the edge weights. In each picture, we perform sampling experiments comparing Algorithm~\ref{alg_sampt_method} against random sampling and the method proposed in~\cite{ortega_proxies}. Considering the bandwith model \textbf{BWM2} we generate four graphs, $\left\lbrace G_{i}^{W} \right\rbrace_{i=1}^{4}$, with a number of nodes $N$ and from a given graphon $W(u,v)$. In each $G_{i}^{W}$ we pollute the edges with noise, using the four SNR values given by $SNR=\{ 10, 15, 20, 25 \} [dB]$. To generate the graphs we use the method (GD1). In each column, the pictures presented are associated with the graphon displayed at the bottom and in each row, we display pictures regarding experiments with a given number of nodes. For each sampling experiment on the graph $G_{i}^{W}$ with $N$ nodes, the sampling set is obtained applying Algorithm~\ref{alg_sampt_method} using the optimal sampling set in a graph $G_{0}^{W}$ with $N/2$ nodes generated from $W$.}
	\label{fig_noise_robustness_b}
\end{figure*}


In this supplementary material we add a collection of extra results and experiments not deposited in the main document due to the lack of space. In the following subsections we discuss the details of each experiment.


\subsection{Scenario~\textbf{BMW1} for Sinusoidal Graphons}

In the context of the experiments described in Section~\ref{sec_num_sim}, we present in Fig.~\ref{fig_error_rec_exp_4a}, the results regarding scenario \textbf{BMW1} for the graphons $W(u,v)=\left\vert\sin\left(100uv\right)\right\vert$, $W(u,v)=\left\vert\sin\left(64uv\right)\right\vert/2+\left\vert\cos\left(64uv\right)\right\vert/2$, and $W(u,v)=\left\vert\sin\left(10uv\right)\right\vert/2+\left\vert\cos\left(10uv\right)\right\vert/2$.


\subsection{Experiments with Larger Graphs}

In Fig.~\ref{fig_error_rec_exp_bigraphs} we present numerical experiments along the lines of Section~\ref{sec_num_sim}, but considering significantly larger graphs where applying the approach in~\cite{ortega_proxies} becomes intractable. To show this experiment we consider six different graphons on the scenario \textbf{BWM2} described in Section~\ref{sec_num_sim}. The optimal sampling set is calculated by~\cite{ortega_proxies} on the graph with $N=3000$ nodes, and it is used in Algorithm~\ref{alg_sampt_method} to obtain approximately optimal sampling sets when $N=6000$, $N=9000$, and $N=12000$. Please notice that this experiment is added per request of one of the reviewers and the goal is to show that even when one cannot compare directly with~\cite{ortega_proxies} the goodness of the sampling sets is still preserved by our algorithm. This is, Algorithm~\ref{alg_sampt_method} is a concrete working method for real large graphs.


\subsection{Numerical Robustness against Graph Perturbations}

Figures~\ref{fig_noise_robustness_a} and~\ref{fig_noise_robustness_b} present the results of the experiment we perform to evaluate the robustness of the approximately optimal sampling sets from Algorithm~\ref{alg_sampt_method} when the graphs are affected by perturbations produced by random noise. The experiment proceed according to the following steps:


\begin{enumerate}
    \item We generate four graphs, $\left\lbrace G_{i}^{W} \right\rbrace_{i=1}^{4}$, with a number of nodes $N=\left\vert V\left( G_{j}^{W}\right) \right\vert$, from a given graphon $W(u,v)$. To generate the graphs we use the method (GD1).
    \item All graphs $G_{i}^{W}$ generated from the same graphon, $W$, and with a same number of nodes, $N$, is contaminated with noise considering the SNR values given by $SNR=\{ 10,15,20,25 \} [dB]$. This is, $G_{1}^{W}$ is contaminated with noise such that $SNR=10[dB]$ on each edge, $G_{2}^{W}$ is contaminated with noise such that the $SNR=15[dB]$ on each edge, so on and so forth. 
    \item Given a fixed bandwidth, we obtain approximately optimal sampling sets with Algorithm~\ref{alg_sampt_method} using the optimal sampling set on $G_{0}^{W}$ of size $N/2$ and generated from $W(u,v)$. To obtain the optimal sampling set in $G_{0}^{W}$ we use~\cite{ortega_proxies}.
    \item We proceed to perform the sampling experiment for each $\left\lbrace G_{i}^{W} \right\rbrace_{i=1}^{4}$ along the lines of Section~\ref{sec_num_sim} considering  that the bandwidth, $k_{\omega}$, is $90\%$ of the sampling rate, $m$, which is $5\%$ with respect to the number of nodes $N$, i.e.
\begin{equation}
       	k_\omega = \left\lfloor 0.9m \right\rceil,~\quad m=\left\lfloor 0.05 N \right\rceil.
\end{equation}
    \item We present our results plotting the average reconstruction error vs the SNR describing the perturbed edges of the graphs. 
\end{enumerate}

\ifCLASSOPTIONcaptionsoff
  \newpage
\fi

\end{document}